\documentclass[letterpaper,twocolumn,10pt]{article}
\usepackage{usenix2019_v3}

\usepackage{tikz}
\usepackage{amsmath}

\usepackage{filecontents}

\usepackage{url}
\usepackage{verbatim}
\usepackage{cite}
\usepackage{amsmath,amssymb,amsfonts}
\usepackage{lipsum}
\usepackage{caption}
\usepackage{graphicx}
\usepackage{textcomp}
\usepackage{xcolor}
\usepackage{subfigure}
\usepackage{amsthm}
\usepackage{dsfont}
\usepackage{amsmath}
\usepackage{algorithmic}
\usepackage{algorithm}
\usepackage{multirow}
\usepackage{colortbl}
\usepackage{tabularx}
\usepackage{appendix}
\usepackage{tikz}
\usepackage{diagbox}

\usepackage{filecontents}
\newtheorem{definition}{Definition}
\newtheorem{theorem}{Theorem}

\usepackage[most]{tcolorbox}

\begin{document}

%
\title{ \Large \bf 
Can We Trust the Similarity Measurement in Federated Learning?
}

\author{
{Zhilin Wang$^{\dag}$, Qin Hu$^{\ddag}$, and Xukai Zou$^{\ddag}$}\\
$^{\dag}$ Department of Computer Science, Purdue University, IN, USA\\
$^{\ddag}$ Department of Computer Science, Indiana University Indianapolis, IN, USA\\
Email: wang5327@purdue.edu, qinhu@iu.edu, xzou@iupui.edu
\and
} 

\maketitle

\begin{abstract}
\textit{Is it secure to measure the reliability of local models by similarity in federated learning (FL)?} This paper delves into an unexplored security threat concerning applying similarity metrics, such as the $L_2$ norm, Euclidean distance, and cosine similarity, in protecting FL.  We first uncover the deficiencies of similarity metrics that high-dimensional local models, including benign and poisoned models, may be evaluated to have the same similarity while being significantly different in the parameter values. We then leverage this finding to devise a novel untargeted model poisoning attack, \textit{Faker}, which launches the attack by simultaneously maximizing the evaluated similarity of the poisoned local model and the difference in the parameter values.  Experimental results based on seven datasets and eight defenses show that Faker outperforms the state-of-the-art benchmark attacks by \texttt{1.1-9.0X} in reducing accuracy and \texttt{1.2-8.0X} in saving time cost, which even holds for the case of a single malicious client with limited knowledge about the FL system. Moreover, Faker can degrade the performance of the global model by attacking only once. We also preliminarily explore extending Faker to other attacks, such as backdoor attacks and Sybil attacks. Lastly, we provide a model evaluation strategy, called the \textit{similarity of partial parameters (SPP)}, to defend against Faker. Given that numerous mechanisms in FL utilize similarity metrics to assess local models, this work suggests that we should be vigilant regarding the potential risks of using these metrics.
\end{abstract}

\section{Introduction}\label{intro}
\if()
As a distributed machine learning framework, 
federated learning (FL) has been proposed to address the challenges of isolated data island and privacy-preserving 
machine learning\cite{konevcny2016federated,mcmahan2017communication,kairouz2021advances}. In FL, local devices (i.e., clients) are not required to submit their raw data to the centralized server but collaboratively train a shared model by regularly uploading their local models being trained using their respective datasets. Nowadays, FL has been widely applied in various fields, such as smart city\cite{jiang2020federated,qolomany2020particle,albaseer2020exploiting}, Internet of Things (IoT) \cite{nguyen2021federated,ghimire2022recent,wang2022blockchain}, and edge computing \cite{wang2019adaptive,zhou2022pflf,xu2022psdf}. 
\fi
\if()
Even though FL is proven to be effective in protecting the data privacy of clients \cite{truex2019hybrid,wei2020federated,hao2019efficient}, 
it still faces a number of challenges, such as non-identical independent distributed (non-IID) data\cite{zhao2018federated,sattler2019robust,li2022federated}, stragglers\cite{reisizadeh2022straggler,li2021stragglers,dhakal2019coded}, fairness\cite{yu2020fairness,sultana2022eiffel,fan2022improving}, privacy\cite{truex2019hybrid,wei2020federated,hao2019efficient}, and security\cite{rodriguez2023survey,lyu2020threats,mothukuri2021survey}. This work mainly focuses on the security of FL; more specifically, we study the poisoning attacks on FL. We can classify poisoning attacks into two categories, i.e., targeted attack and untargeted attack. As for the targeted attack, e.g., backdoor attack \cite{bagdasaryan2020backdoor,sun2019can,wang2020attack,rieger2022deepsight} and label-flipping attack\cite{li2021detection,zhang2021robustfl}, its goal is to decrease the test accuracy of the global model on some certain labels, but the overall accuracy is well maintained. As for the untargeted attack, e.g., model poisoning attack (MPA) \cite{zhou2021deep,shejwalkar2022back,wang2022defense} and data poisoning attack\cite{tolpegin2020data,doku2021mitigating,nuding2022data}, is designed to degrade the overall accuracy of the global model on test data.  Those attacks can not only undermine the operation of the FL system but also degrade the performance of the learned global model. 
\fi



\if()
\begin{table}
\centering
\caption{Similarity-based AGMs.}
\arrayrulecolor{black}
\resizebox{\linewidth}{!}{
\begin{tabular}{cc|cc} 
\arrayrulecolor{black}\hline
Metric                    & \multicolumn{1}{c!{\color{black}\vrule}}{AGM} & Metric              & AGM                                                              \\ 
\cline{1-1}\arrayrulecolor{black}\cline{2-3}\arrayrulecolor{black}\cline{4-4}
\multirow{4}{*}{ED}       & Krum   \cite{blanchard2017machine}                                       & \multirow{7}{*}{CS} & Zeno++ \cite{xie2020zeno++}                                                          \\
                          & Multi-Krum  \cite{blanchard2017machine}                                  &                     & ShieldFL \cite{ma2022shieldfl}                                                        \\
                          & Bulyan   \cite{guerraoui2018hidden}                                     &                     & SignGuard \cite{xu2022byzantine}                                                       \\
                          & FLARE \cite{wang2022flare}                                        &                     & CONTRA   \cite{awan2021contra}                                                        \\ 
\arrayrulecolor{black}\cline{1-2}
\multirow{3}{*}{$L_2$ norm} & Norm-clipping \cite{sun2019can}                                &                     & attestedFL\cite{mallah2021untargeted}                                                      \\
                          & Snipher  \cite{cao2019understanding}                                     &                     & FL-Defender \cite{jebreel2023fl}                                                     \\
                          & AFLGuard  \cite{fang2022aflguard}                                    &                     & FoolsGold \cite{fung2018mitigating}                                                       \\ 
\cline{1-3}\arrayrulecolor{black}\cline{4-4}
\multicolumn{2}{c}{Metric}                                                & \multicolumn{2}{c}{AGM}                                                                \\ 
\cline{1-2}\arrayrulecolor{black}\cline{3-4}
\multicolumn{2}{c}{CS \& $L_2$ norm}                          & \multicolumn{2}{c}{\begin{tabular}[c]{@{}c@{}}FLAME\cite{nguyen2022flame}\\DiverseFL\cite{prakash2020mitigating}\\FLTrust \cite{cao2020fltrust}\end{tabular}}  \\
\hline
\end{tabular}}
\label{sim_agms}
\end{table}
\fi
Federated learning (FL) requires multiple local devices (i.e., clients) to train a shared model collaboratively. During this process, clients submit the trained local models to the centralized server for aggregation while the raw training data is maintained locally \cite{mcmahan2017communication}. Due to the distributed nature of FL, it is nearly impossible to ensure that all clients are benign. 

\if()
Thus, security issues have severely hindered the thriving of FL\cite{rodriguez2023survey,lyu2020threats}, 
among which the poisoning attacks are prominent as they can significantly degrade the model performance\cite{shejwalkar2022back}. As one of the most efficient untargeted poisoning attacks, model poisoning attacks have been discussed extensively in the past few years. Model poisoning attacks aim to undermine the overall performance of the global model on test data by modifying the local models before submitting them to the server \cite{zhou2021deep,wang2022defense}. 
\fi

\begin{figure}
\centering
\includegraphics[width=7.5cm]{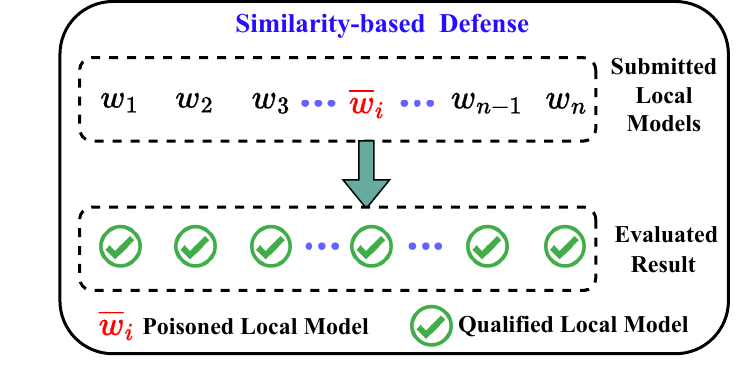}
\caption{Measuring the reliability of local models by similarity is insecure. The attacker can submit well-crafted poisoned local models to pass the detection of similarity metrics.}
\label{faker_overview}
\end{figure}

Such a distributed system is vulnerable to adversarial attacks, including model poisoning attacks \cite{fang2020local}, backdoor attacks \cite{bagdasaryan2020backdoor}, Sybil attacks\cite{fung2020limitations}, etc. These attacks have different targets and initiation methods, but all of them require the manipulation of local model parameters. Take model poisoning attacks as an example,  which undermine the overall accuracy of the global model on test data by modifying local models before submitting them to the server \cite{zhou2021deep,wang2022defense}. The attackers can bypass the detection of the defenses by using an iterative approach to find a shared scalar for all the parameters in the local model to change its direction and/or magnitude \cite{fang2020local, shejwalkar2021manipulating}. Some attacks also utilize generative adversarial networks \cite{zhang2019poisoning} and the data variance \cite{baruch2019little} to generate poisoned local models. However, in this work, we find that most existing attacks are computationally inefficient and incapable of guaranteeing the success of attacks; besides, most of them require the cooperation of multiple malicious clients to launch effective attacks, which is not applicable to distributed machine learning systems in practice.

\if()
The current model poisoning attacks explore the vulnerabilities of defenses in FL, where different approaches are adopted to generate poisoned local models. 
In \cite{fang2020local}, the authors argue that changing the directions of local models can lead to the success of model poisoning attacks, and the authors in \cite{shejwalkar2021manipulating} indicate that maximizing the difference of magnitudes between benign and poisoned local models would be the main objective of launching model poisoning attacks.
The above works employ iterative methods to find a suitable scalar for generating poisoned local models towards fooling different defenses, and other studies utilize generative adversarial networks \cite{zhang2019poisoning} or data variance \cite{baruch2019little} to launch attacks.
These model poisoning attacks are powerful in attacking the global model to some extent, but they are far from computationally efficient for a single local device to launch the attack; besides, their generated poisoned models cannot always bypass the detection of defenses. Here we argue that a successful model poisoning attack should generate poisoned local models quickly while guaranteeing attacking performance and remaining undetected.
\fi

To defend against adversarial attacks, various defensive strategies have also been proposed recently. However, since the server (i.e., the defender) has no access to clients' raw data that are usually non-independent and identically distributed (non-IID) \cite{li2022federated},  
mitigating the impacts of the model poisoning attack on FL is challenging. 
Researchers design defenses based on various metrics to evaluate the local models before aggregation, 
assuming that the poisoned local model is significantly different from the benign one \cite{cao2019understanding,hu2021challenges}. Commonly used metrics include similarity metrics (e.g., $L_2$ norm, Euclidean distance, and cosine similarity as summarized in Table \ref{sim_agms}),
statistical metrics (e.g., mean, median, and trim-mean)\cite{yin2018byzantine}, and sparcification\cite{panda2022sparsefed}. 
However, the security of those metrics is questionable as smart attackers can carefully craft poisoned local models to fool the server once the employed metrics are available to the attackers by any means. A recent work \cite{kasyap2022hidden} illustrates that cosine similarity applied in FLTrust \cite{cao2020fltrust} is not robust. Our investigation in this work finds that most defenses against adversarial attacks do not consider the risk of being attacked due to deficiencies of evaluation metrics, especially the most widely applied similarity metrics. Hence, the question arises: \textit{is it secure to measure the reliability of local models by the evaluated similarity in FL?}


\textbf{Our Work and Contributions:} We find that measuring the reliability of local models by similarity is not secure. We first analyze the vulnerabilities of similarity metrics in evaluating local models \footnote{If not otherwise specified, the similarity metrics refer to $L_2$ norm, Euclidean distance, and cosine similarity.}. It turns out to be that there exist malicious models similar to the benign ones in terms of evaluation results but the values of parameters in the models are significantly different. Such deficiencies are mainly attributed to the fact that the evaluated similarity of high-dimension local models is jointly affected by all parameters. Thus, 
the attacker can carefully craft the parameter values in the poisoned local model to obtain an acceptable evaluation result (see Fig. \ref{faker_overview}). 

 To illustrate the potential threats to FL systems posed by similarity metrics, we design a simple yet effective untargeted model poisoning attack based on their shortcomings, termed \textit{Faker}. The main objective of Faker is to allow the attacker to submit well-crafted poisoned local models to pass the detection of similarity metrics while being malicious. In this way, we transfer the model poisoning attack into an optimization problem that maximizes the similarity of the poisoned local model and the difference in parameter values between the poisoned and benign models. For a single malicious client, launching Faker is as simple as only knowing its own trained local model and the adopted defense, and then solving the formulated optimization problems for a certain defense mechanism. The variables of the optimization problem are the scalars to change each parameter in the local model. 
 In practice, we take measures, such as assigning values to some of them in advance within a specific feasible domain and grouping variables, to reduce the difficulty of the solution. As a result, the worst-case time complexity of Faker can be reduced from $\mathcal{O}(J)$ with $J$ being the number of local model parameters to $\mathcal{O}(T)$ with $T\ll J$ being the number of divided groups.
 
 We conduct extensive experiments on seven datasets using eight benchmark defenses, and the results show that Faker outperforms the most widely discussed benchmark model poisoning attacks (i.e., LA\cite{fang2020local} and MB\cite{shejwalkar2021manipulating}) by \texttt{1.1-9.0X} in reducing accuracy and \texttt{1.2-8.0X} in saving time cost regardless of the data distribution, even with limited knowledge about the FL system. The most significant performance difference from benchmark attacks is that Faker can always maintain a 100\% attack success rate. Besides, Faker can successfully undermine the global model by attacking only once. While our focus in this paper is on untargeted model poisoning attacks, we also explore the potential of expanding Faker in targeted backdoor attacks and Sybil attacks. The preliminary experimental results show that Faker performs well on other adversarial attacks, further demonstrating the threats of similarity metrics' vulnerabilities.

 To defend against Faker, we sketch a strategy, \textit{the similarity of partial parameters (SPP)}. Specifically, SPP requires the server to calculate the similarity of randomly selected partial parameters rather than all the parameters of the local model. Thus, the attackers can not know how local models will be evaluated, increasing the possibility of the poisoned local models being detected. The theoretical analysis and experimental results compared with the benchmark defense show that SPP is effective in resisting Faker.

Overall, the major contribution of our paper is to reveal the unreliability of similarity metrics in FL. Given the extensive applications of similarity metrics in FL systems, ranging from security to other areas such as fairness \cite{divi2021new}, client selection\cite{fraboni2021clustered}, heterogeneity problem \cite{huang2022learn}, and the speedup of convergence \cite{ouyang2021clusterfl}, this finding challenges most existing mechanisms in FL designed based on similarity metrics. We therefore call on researchers and developers to conduct in-depth studies on the security of similarity metrics, as well as to be cautious when using them to design mechanisms.


\if()
\underline{\textit{\textbf{Q1:} How to launch an effective MPA?}}

For different AGMs, the required poisoned local models in MPAs can vary significantly. Therefore, we propose the \textit{importance and bias (I\&B}) framework to analyze the general process of MPAs in a quantitative manner to shed light on the effective poisoned local models regarding a specific AGM. We argue that a successful MPA needs to maximize two objectives: i) the importance (or contribution) of the poisoned local model during aggregation, and ii) the bias (or distance) between the poisoned and the unpoisoned global models. Besides, both the importance and the bias are quantifiable and controllable.

\underline{\textit{\textbf{Q2:} How to generate poisoned local models?}}

To bypass the detection of a similarity-based AGM, the most critical step is to ensure that the generated poisoned local models can meet the similarity requirements of the AGM. To this end, we propose a novel poisoned local model generator, \textit{EvilTwin}, which is based on the vulnerabilities of similarity metrics, i.e., $L_2$ norm, Euclidean distance, and cosine similarity. EvilTwin can generate a fake vector $\overline{X}\neq X$ which has arbitrary similarity $\theta$ to a given vector $Y$, i.e., $S(X,Y)=S(\overline{X},Y)=\theta$ with $S(\cdot,\cdot)$ being the similarity calculation method\footnote{All FL models are reshaped as vectors in this paper.}. EvilTwin calculates the optimal scalars for each parameter in local models based on derived equations and/or inequalities, which is more efficient than the iterative
methods\cite{fang2020local,shejwalkar2021manipulating}.
\fi

\if()
\underline{\textit{\textbf{Q3:} How to undermine similarity-based AGMs?}}

By integrating I\&B and EvilTwin, we design a new MPA, Faker, to undermine different similarity-based AGMs. We theoretically prove that Faker can be launched by a single malicious client using its own local model in any training round and achieve a 100\% success rate without being detected. We further experimentally demonstrate that, with five datasets and seven AGMs, Faker outperforms the most widely discussed benchmark attacks (i.e., LA\cite{fang2020local} and MB\cite{shejwalkar2021manipulating}) by 1.1-9.0x in reducing accuracy and 1.2-8.0x in saving time cost regardless of the data distribution, even with limited knowledge about the FL system. Besides, Faker can successfully undermine the global model by attacking only once. Overall, Faker is a stealthy and adaptive MPA.

\underline{\textit{\textbf{Q4:} How to defend against Faker?}}

We sketch a strategy, \textit{the similarity of partial parameters (SPP)}, to defend against Faker. Specifically, SPP requires the server to calculate the similarity of randomly selected partial parameters rather than all the parameters of the local model. In this way, the attackers do not know how the local models they submit will be evaluated, increasing the probability of the poisoned local models being detected. The theoretical analysis and experimental results show that SPP performs well in resisting Faker.

Beyond the aforementioned four aspects of contributions related to the security of FL, given the importance of similarity in FL, our study on similarity metric security can inspire research on many other subareas of FL, such as fairness \cite{divi2021new}, client selection\cite{fraboni2021clustered}, heterogeneity problem \cite{huang2022learn}, and the speedup of convergence \cite{ouyang2021clusterfl}. Thus, the threats regarding the vulnerabilities of similarity metrics are not limited to the security of FL. We also discuss some promising research directions in Appendix \ref{d_f}, including the potential of extending Faker to backdoor attacks \cite{bagdasaryan2020backdoor} and the security of other evaluation metrics, as well as a preliminary exploration of robust and lightweight evaluation for large-scale models in industrial FL systems. Overall, this paper calls on researchers and developers to pay attention to the evaluation process of local models when designing FL-related mechanisms, especially to the security of evaluation metrics.
\fi


\textbf{Outline:} We provide preliminaries and related work in Section \ref{rl}, and the threat model is detailed in Section \ref{threat model}. We analyze the vulnerabilities of similarity metrics in Section \ref{sec_EvilTwin} and introduce  Faker in Section \ref{faker}. The experimental evaluation of Faker is presented in Section \ref{ev}. The defense SPP is shown in Section \ref{spp_fa}, and we conclude this paper in Section \ref{conclusion}. Interested readers can also check the discussions of future research directions in Appendix \ref{f_d}.

\begin{table}[t]
\centering
\caption{Key Notations.}
\resizebox{\linewidth}{!}{
\begin{tabular}{c|c}
\hline
\rowcolor[HTML]{FFCCC9} 
Notations & Meanings \\ \hline
\hline
$w_i$       &   The benign local model of client $i$.       \\
$w_{i,j}$       &   The $j$th parameter of $w_i$.       \\
$w_{i,-j}$       &   Any parameter except $w_{i,j}$ in $w_i$.       \\
 $\overline{w}_i$       &   The poisoned local model of client $i$.       \\ 
 $\overline{w}_{i,j}$       &   The $j$th parameter of $\overline{w}_i$.       \\
 $w_r$       &   The reference model of the defender.       \\ 
 $w_g$       &     The global model of FL.     \\ 
$n$       &     The number of total clients in FL.     \\ 
$m$          &  The number of total malicious clients in FL.        \\ 
$J$       &     The number of parameters in FL's model.     \\ 
$\overline{\delta}_i$       &   The difference between poisoned and benign models.       \\ 
$s_l$ & The lower bound of similarity requirement.\\
$s_u$ & The upper bound of similarity requirement.\\
$\overline{s}_i$       &   The similarity of poisoned  local model $\overline{w}_i$ .       \\ 
$T$     &   The number of divided groups.       \\    
$\alpha_{i}$     & The vector of scalars for $w_i$.         \\
$\alpha_{i,j}$     &The $j$th element in $\alpha_i$. \\
$\alpha_{i,-j}$     &Any scalar except $\alpha_{i,j}$ in $\alpha_i$. \\
\hline
\end{tabular}}
\label{notation}
\end{table}

\section{Background and Related Work}\label{rl}
In this section, we introduce some preliminaries and related works, including 
adversarial attacks, three widely applied similarity metrics in FL, and similarity-based defenses. For convenience, we provide the notation explanations below.

\textbf{Notations:} Assume there are $n$ clients in total to train a global model $w_g$ collaboratively in the FL system, among which $m$ clients are malicious. Given the original local model $w_i=(w_{i,1},\cdots,w_{i,j},\cdots,w_{i,J})$\footnote{All the FL models are treated as vectors in this work.} of client $i$ (i.e., the attacker) in this system, 
with $J$ being the total number of parameters, we denote the poisoned local model submitted by client $i$ as $\overline{w}_i=(\overline{w}_{i,1},\cdots,\overline{w}_{i,j},\cdots,\overline{w}_{i,J})$. Please refer to Table \ref{notation} for more details of the key notations in this work.


\subsection{Adversarial Attacks} We introduce some of the most popular attacks against FL, with more details about model poisoning attacks and a brief discussion about other kinds of attacks.

\textbf{Model Poisoning Attacks.}  Malicious clients may tamper with the local models before submission, resulting in the aggregated global model failing to perform as expected, which is termed model poisoning attacks. We then detail two of the most widely discussed model poisoning attacks, which are also the benchmark attacks in our experiments. In \cite{fang2020local}, Cao \textit{et al.} propose a local attack (LA) that aims to change the local models' directions as large as possible to poison the global model. Specifically, LA first generates a random vector consisting of 1 and -1 to determine whether to change the direction of each parameter in the local model or not, followed by an iterative method to derive a shared scalar for modifying the magnitudes of parameters. Similarly, the authors in \cite{shejwalkar2021manipulating} propose a model poisoning attack called manipulating the Byzantine rules (MB), which also adopts an iterative method to search for a shared scalar for all parameters. In general, most existing attacks are not computationally efficient and cannot guarantee the success of attacks, as well as require the collaboration of multiple attackers.

\textbf{Other Adversarial Attacks.} 
1) Backdoor attacks. In \cite{bagdasaryan2020backdoor}, the authors propose a constraint-and-scale (C\&S) method to generate poisoned local models to undermine the accuracy of certain classes while maintaining the overall accuracy.
2) Sybil Attacks. The attackers undermine the global model by multiple Sybil clients with carefully crafted poisoned local models\cite{fung2018mitigating}.   
Interested readers can refer to \cite{mothukuri2021survey} for more details.

Overall, all of the above-mentioned malicious attacks involve the manipulation of the local models. In addition, 
the existing attacks do not rely on rigorous quantitative methods in generating poisoned models, and thus the success of the attack cannot be guaranteed.


\subsection{Defenses Against Adversarial Attacks}
We first present the widely applied similarity metrics in FL and then introduce the similarity-based defenses against model poisoning attacks and other adversarial attacks.

\textbf{Similarity Metrics.}
Taking the similarity calculation between the local model $w_i$ and the poisoned local model $\overline{w}_i$ as the example, we introduce three widely applied similarity metrics in FL. The $L_2$ norms of $\overline{w}_i$ and $w_i$ are calculated as  $L(\overline{w}_i)=[\sum_{j=1}^J \overline{w}_{i,j}^2]^\frac{1}{2}$ and $L(w_i)=[\sum_{j=1}^J w_{i,j}^2]^\frac{1}{2}$, respectively. $\overline{w}_i$ and $w_i$  are regarded as more similar if $L(\overline{w}_i)/{L(w_i)}\to 1$.  The Euclidean distance between $\overline{w}_i$ and $w_i$ is calculated as $E(\overline{w}_i,w_i)=[\sum_{j=1}^J(\overline{w}_{i,j}-w_{i,j})^2]^\frac{1}{2}.$ When $E(\overline{w}_i,w_i)$ is smaller, i.e., $E(\overline{w}_i,w_i) \to 0$, the similarity between $\overline{w}_i$ and $w_i$ is higher. The cosine similarity between $\overline{w}_i$ and $w_i$ can be calculated as $C(\overline{w}_i, w_i)=\frac{\sum_{j=1}^J \overline{w}_{i,j} w_{i,j}}{L(\overline{w}_i)L(w_i)}$
$\in [-1,1]$, and if $C(\overline{w}_i, w_i)\to 1$, $\overline{w}_i$ and $w_i$ are more similar. 
For simplicity, we use $S(\overline{w}_i,w_i)$ to express any of the above similarity metrics.

\textbf{Defenses Against Model Poisoning Attacks.} 
In this part, we introduce the benchmark similarity-based defenses against model poisoning attacks. Table \ref{sim_agms} summarizes the existing similarity-based defenses, and six of them are detailed below. 
FLTrust\cite{cao2020fltrust} allows the server to collect a clean dataset at the beginning and train a clean model each round. Based on this model, the server evaluates the received local models using cosine similarity and $L_2$ norm. 
FLTrust uses cosine similarity with ReLU to filter out local models with opposite directions to the clean model and utilizes $L_2$ norm to decrease the influence of magnitude change so as to recover the modified parameters. 
Krum \cite{blanchard2017machine} selects only one local model as the global model among $n-m-1$ received local models based on Euclidean distance, where $m$ is the number of malicious clients. Specifically, Krum calculates the Euclidean distance of each local model from all other local models. Then, Krum sorts the calculated distances for each local model, sums up the top $n-m-1$ distances, and selects the local model with the smallest sum of distances as the global model. Note that Krum assumes the server knows $m$ in each round. Norm-clipping \cite{sun2019can} sets an upper bound for the value of $L_2$ norm for each local model. If $L(w_i)$ is larger than the upper bound, then $w_i$ will be discarded before aggregation; otherwise, $w_i$ will be included during aggregation. This approach reduces the risk of attacks by eliminating local models scaled up, where the upper bound is known to the server. FLAME\cite{nguyen2022flame} calculates the cosine similarity among local models, and the server can get a matrix of cosine similarity, and then a clustering method is applied to filter out the malicious local models. $L_2$ norm is adopted to prune the remaining local models, and some carefully generated noise will be added to the local models to improve the performance. In DiverseFL\cite{prakash2020mitigating}, the server trains a clean model based on collected clean data and then calculates the cosine similarity and the ratio of $L_2$ norm between the local model and the clean one. The local model with non-positive cosine similarity and the abnormal ratio of $L_2$ norm will be rejected. ShieldFL\cite{ma2022shieldfl} is a privacy-preserving robust defense, but in this work, we do not use its mentioned privacy-preserving mechanism but directly use its mechanism for detecting malicious models. Specifically, the cosine similarity of the local models is computed several times to identify poisoned models.


\begin{table}[]
\caption{Similarity-based defenses in FL. $L_2$, ED, and CS refer to $L_2$ norm, Euclidean distance, and cosine similarity, respectively.}
\resizebox{\linewidth}{!}{
\begin{tabular}{c|ccc|c|c|ccc}
\hline
\rowcolor[HTML]{FFCCC9} 
Defenses                                                                                                          & $L_2$                                & ED                                   & CS                                   &                     & Defenses                                                                           & $L_2$                                & ED                                   & CS                                   \\  
\hline
\hline
Krum \cite{blanchard2017machine}                                                             &                                      & $\checkmark$                         &                                      &                     & CONTRA \cite{awan2021contra}                                   &                                      &                                      & $\checkmark$                         \\  
\cellcolor[HTML]{C0C0C0}Buyukates \textit{et al.} \cite{buyukates2023proof} & \cellcolor[HTML]{C0C0C0}$\checkmark$ & \cellcolor[HTML]{C0C0C0}             & \cellcolor[HTML]{C0C0C0}$\checkmark$ &                     & \cellcolor[HTML]{C0C0C0}FLTrust \cite{cao2020fltrust}          & \cellcolor[HTML]{C0C0C0}$\checkmark$ & \cellcolor[HTML]{C0C0C0}             & \cellcolor[HTML]{C0C0C0}$\checkmark$ \\ 
Bulyan \cite{guerraoui2018hidden}                                                            &                                      & $\checkmark$                         &                                      &                     & FL-Defender \cite{jebreel2023fl}                               &                                      &                                      & $\checkmark$                         \\
$G^2$uardFL \cite{yu2023g}                                                                   &                                      &                                      & $\checkmark$                         &                     & FLARE \cite{wang2022flare}                                     &                                      & $\checkmark$                         &                                      \\ 
\cellcolor[HTML]{C0C0C0}MESAS \cite{krauss2023avoid}                                         & \cellcolor[HTML]{C0C0C0}             & \cellcolor[HTML]{C0C0C0}$\checkmark$ & \cellcolor[HTML]{C0C0C0}$\checkmark$ &                     & SignGuard \cite{xu2022byzantine}                               &                                      &                                      & $\checkmark$                         \\ 
Multi-Krum \cite{blanchard2017machine}                                                       &                                      & $\checkmark$                         &                                      &                     & FoolsGold \cite{fung2018mitigating}                            &                                      &                                      & $\checkmark$                         \\ 
\cellcolor[HTML]{C0C0C0}FLAME \cite{nguyen2022flame}                                         & \cellcolor[HTML]{C0C0C0}$\checkmark$ & \cellcolor[HTML]{C0C0C0}             & \cellcolor[HTML]{C0C0C0}$\checkmark$ &                     & attestedFL \cite{mallah2021untargeted}                         &                                      &                                      & $\checkmark$                         \\  
AFLGuard \cite{fang2022aflguard}                                                             &                                      & $\checkmark$                         &                                      &                     & Norm-clipping \cite{sun2019can}                                & $\checkmark$                         &                                      &                                      \\  
Sniper \cite{cao2019understanding}                                                           & $\checkmark$                         &                                      &                                      &                     & ShieldFL \cite{ma2022shieldfl}                                 &                                      &                                      & $\checkmark$                         \\  
Zeno++ \cite{xie2020zeno++}                                                                  &                                      &                                      & $\checkmark$                         & \multirow{-11}{*}{} & \cellcolor[HTML]{C0C0C0}DiverseFL \cite{prakash2020mitigating} & \cellcolor[HTML]{C0C0C0}             & \cellcolor[HTML]{C0C0C0}$\checkmark$ & \cellcolor[HTML]{C0C0C0}$\checkmark$ \\   
APFed \cite{chen2023apfed}                                                                  &                                      &                                      & $\checkmark$                 &     &   Hu \textit{et al.} \cite{hu2023efficient}                                                                  &                                      &                                      & $\checkmark$ \\ 
\cellcolor[HTML]{C0C0C0}Han \textit{et al.} \cite{han2023kick}                                         & \cellcolor[HTML]{C0C0C0}$\checkmark$ & \cellcolor[HTML]{C0C0C0}             & \cellcolor[HTML]{C0C0C0}$\checkmark$ &                     &          FPD \cite{wan2023four}              &                                      &                                      &    $\checkmark$   \\
\hline
\end{tabular}}
\label{sim_agms}
\end{table}

\textbf{Defenses Against Other Attacks.} 
The above-mentioned norm-clipping, FLAME, and DiverseFL can also be applied to defend against backdoor attacks since the poisoned local models are also different from the benign ones in terms of similarity. FoolsGold \cite{fung2018mitigating}  is designed to mitigate the impacts of Sybil attacks, which distinguishes between malicious and benign models based on the cosine similarity.

Generally, these defenses can evaluate local models in terms of magnitude and direction, but there is usually a strong assumption that the adopted evaluation metrics are robust enough, which may not hold in reality. And these defenses could totally fail if adversaries exploit the vulnerabilities of these metrics to launch attacks.
Based on this finding, we design an efficient model poisoning attack in this work to attack similarity-based defenses.

\if()
Some studies apply similarity metrics to evaluate the submitted local models, such as FLTrust, Krum, and norm-clipping (see Section \ref{sm_agm}). Some studies use statistical methods to mitigate the impacts of MPAs. For example, Yin \textit{et al.} propose Median to defend against AGM by first ranking the parameters among all the local models and then choosing the median to get the global model \cite{yin2018byzantine}. 
Other sophisticated defenses are also designed. In \cite{sun2021fl}, the authors use certified robustness to defend against the MPA and provide a guarantee of convergence for FedAvg. In \cite{panda2022sparsefed}, SparseFed,  an efficient method using global top-k sparsification and gradient clipping on local devices, is designed to resist MPAs. Cao \textit{et al.} design FedRecover to recover the global model with high accuracy by empowering the server to estimate the model updates of clients based on their historical data 
\cite{cao2022fedrecover}. Flare, a defense method against model poisoning attacks, relies on model parameters and penultimate layer representations of local models to demonstrate the adversarial impacts on local updates \cite{wang2022flare}. Zhang \textit{et al.} design FLDetector by detecting the consistency of the local model updates and removing the detected malicious updates\cite{zhang2022fldetector}. 
\fi

\section{Threat Model}\label{threat model}
In this paper, Faker is mainly designed to launch model poisoning attacks, which is also extended to other attacks such as backdoor attacks and Sybil attacks in experimental evaluation. Therefore, we only present the generally adopted threat model for model poisoning attacks in below but leave the threat models for other attacks in later sections.

\textbf{Attackers' Objectives.} 
 We consider the untargeted poisoning attacks in this work, where attackers aim to degrade the overall performance of FL as much as possible by submitting carefully crafted poisoned local models. The attackers are malicious clients of FL or outsiders who control several clients. In the following, we use attackers and malicious clients interchangeably. Besides, attackers would like to conduct attacks stealthily with malicious behaviors not being detected during the aggregation stage. 
 
\textbf{Attackers' Knowledge and Capabilities.} 
The attackers launch the model poisoning attack in each communication round before local model submission.
Similar to other existing work \cite{fang2020local,shejwalkar2021manipulating}, the attackers are assumed to know the applied defenses but cannot control or collude with the server or benign clients. 
The attackers can only train local models based on their own local data and are not allowed to know any models or data of benign clients, nor can they get additional clean data from the server. Besides, we assume that the number of malicious clients is no more than 50\% of the total number of clients, with a focus on the case of only one malicious client.


\textbf{Defender's Objectives.} Typically, the defender is deployed on the server side, and we use the defender and the server interchangeably in this work. The main goal of the defender is twofold: one is to accurately identify malicious models, and the second is to reduce the impact of malicious models on the global model. In addition, the defender also expects that defending against an attack does not cost excessive computational resources.

\textbf{Defender's Knowledge and Capabilities.} The defender has the computational capacity and necessary data required by certain defense schemes. For example, FLTrust requires the server to collect some clean data to train a model for comparison with the local models. Besides, the defender does not know in advance the number of attackers or their identities. 

\if()
\section{I\&B: Quantitative Analysis of MPAs
}\label{i_b}
To launch an effective MPA, 
the attackers need to know what kind of poisoned local models should be generated to degrade the global model performance.
To that end, we propose a novel analytical framework, called  \textit{importance and bias (I\&B)}, to assist in understanding the MPA by quantifying the impact of the poisoned local models on the global model. We first detail the goals of the MPA and then introduce the general model of I\&B.
\if()
\begin{figure}[h!t]
\centering
\includegraphics[width=240]{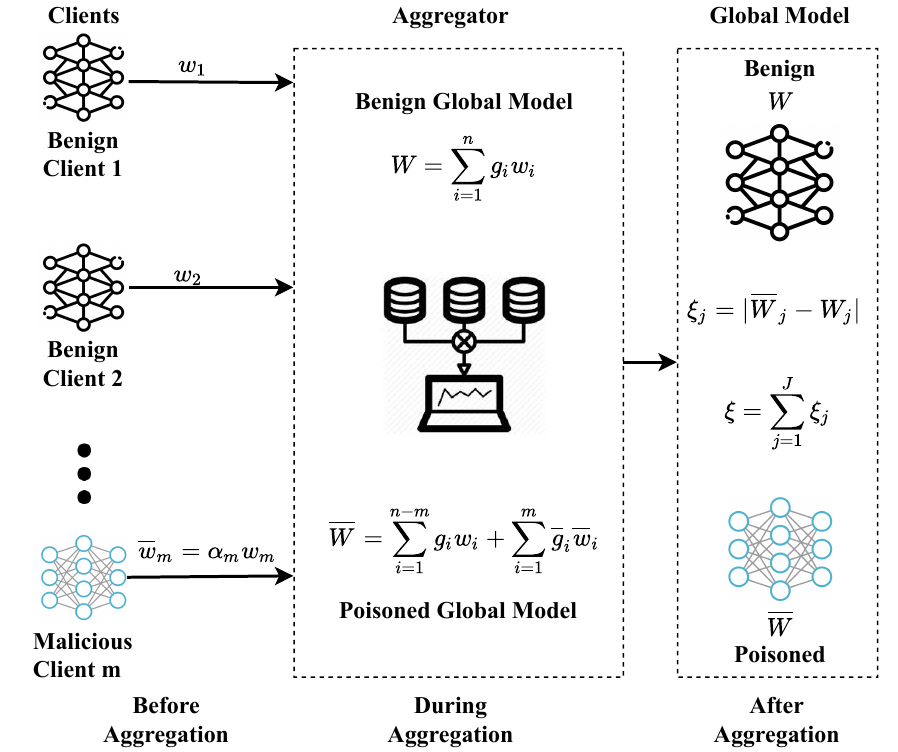}
\caption{Three phases of model poisoning attack. $w_i$ and $\overline{w}_i$ are benign and poisoned local models, respectively; $m$ is the number of malicious clients; $g_i$ and $\overline{g}_i$ are importance; $W_j$ and $\overline{W}_j$ are the parameters of global models; $\xi_j$ and $\xi$ are biases.  }
\label{evil}
\end{figure}
\fi
We divide the process of MPA into three phases, i.e., before, during, and after aggregation (see Fig. \ref{evil_flow}). The first phase includes model modification and submission, 
the second phase is aggregating the collected local models by the aggregator, and the third phase is obtaining the new global model after aggregation. 
The local model submitted in the first phase is determined by the next two phases, so we detail these two phases in this section to help the attacker determine which poisoned model should be generated in the first phase, which will be elaborated in Sections \ref{sec_EvilTwin} and \ref{faker}.


\if()
\begin{figure}[h!t]
\centering
\includegraphics[width=240]{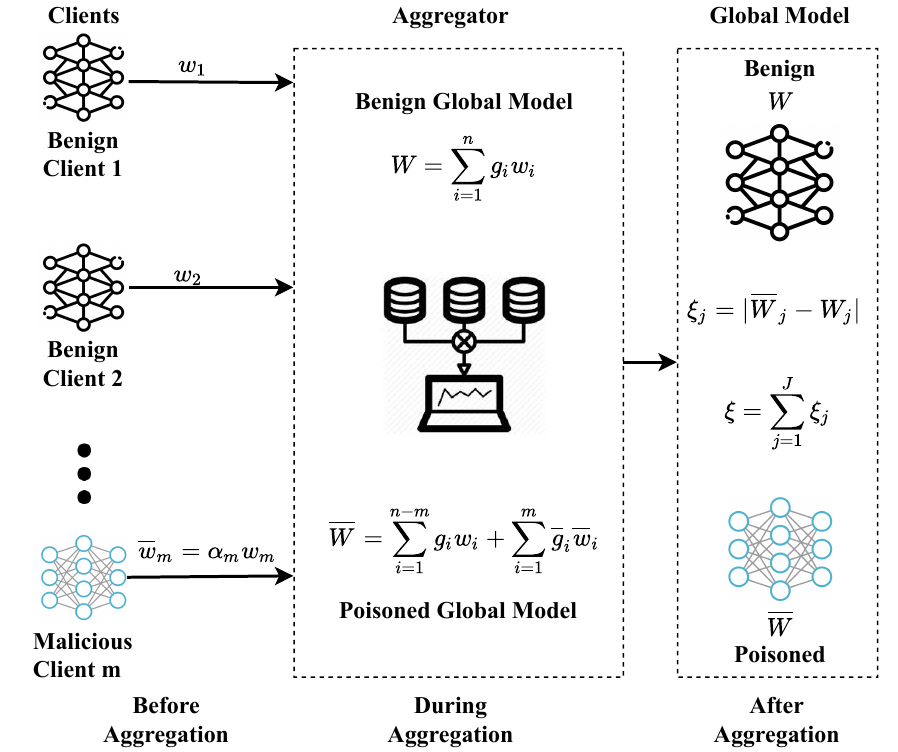}
\caption{Three phases of model poisoning attack. $w_i$ and $\overline{w}_i$ are benign and poisoned local models; $m$ is the number of malicious clients; $g_i$ and $\overline{g}_i$ are importance; $W_j$ and $\overline{W}_j$ are the parameters of global models; $\xi_j$ and $\xi$ are biases. The objective of I\&B is to maximize both $\overline{g}_i$ and $\xi$.}
\label{evil_flow}
\end{figure}
\fi
\subsection{General Goal of MPA}

The general goal of MPA is to degrade the performance of the global model $W$, including preventing the convergence of the loss function and decreasing the test accuracy.
Achieving such a goal is challenging in MPA. Fang \textit{et al.} \cite{fang2020local} argue that the objective can be achieved by changing the direction of the global model maximally, and Shejwalkar \textit{et al.} \cite{shejwalkar2021manipulating} emphasize that the attackers should maximize the difference between the magnitudes of the benign and poisoned global models. Even though these two opinions are reasonable to some extent,  they cannot capture the exact effects of MPAs on the performance of the global model in a comprehensive way. In most of the existing defenses, there are some predefined requirements for the directions and magnitudes of the submissions, so changing directions or magnitudes solely is not always effective in practice. 
Different from these two attack strategies, in our work, we argue that the general objective of MPA is to bypass the detection of the AGMs and enlarge the bias between the benign and poisoned global models in both directions and magnitudes. We detail this argument by introducing the following two goals.

\if()
\textbf{Goal of MPA Before Aggregation}:  
the existing studies have rarely discussed the objectives of MPA before model aggregation. Although such an objective is not the focus of our paper, we present it for a better understanding of MPA. Intuitively, the attackers must ensure that their poisoned local models can be submitted to the aggregator in time and avoid their malicious behaviors being detected. 
\fi
\subsection{Goal of MPA During Aggregation}\label{ge_IB}
The state-of-the-arts usually ignore how the poisoned local models affect the process of global aggregation; however, this must be addressed because it determines whether the poisoned local models would be accepted by the AGM. In the following, we first define the \textit{importance} of the local model to assist in demonstrating the process of global aggregation.

\begin{definition}
(Importance of Local Model $g_i$) The importance of the local model $w_i$  evaluates the contribution of a local model in the global aggregation, which is termed $g_i$.
\end{definition}

We can consider model aggregation as the process of assigning an importance value to each local model and then summing the weighted local models. For example, FedAvg \cite{mcmahan2017communication} uses the proportion of data size of local training data as the importance; and Krum\cite{blanchard2017machine}  assigns the importance value 1 to the local model with the shortest distance to other models and 0 otherwise. It can be seen that when $g_i$ is larger, the contribution of $w_i$  to the global model aggregation is more significant, and it is more likely that $w_i$ will affect the global model. Then, we can have the following definition for the goal of MPA during model aggregation. 

\begin{definition} (Goal of MPA During Aggregation) The goal of MPA during aggregation is to maximize the importance of the poisoned local model $\overline{w}_i$, i.e., $\overline{g}_i$.
\end{definition}

We can roughly classify the AGMs into two categories, i.e., weight-based AGMs and statistics-based AGMs. Weight-based AGMs assign each local model a weight and sum up the weighted local models as the global model, such as FLAME and FLTrust, where the importance of the local model is exactly the assigned weight during aggregation. As for statistics-based AGMs, usually, only one local model will be chosen by the aggregator as the global model, such as Krum; thus the importance of the local model is either 1 or 0. Please note that even though there are also mixed AGMs based on both weight and statistics, such as multi-Krum\cite{fang2020local}, we argue that $g_i$ and $\overline{g}_i$ can be calculated since global aggregation in FL is essentially obtaining a global model based on multiple local models.

\subsection{Goal of MPA After Aggregation}
Some existing MPAs try to maximize the overall distance between the poisoned and the benign aggregated models, however, 
there is no formal mathematical expression of such difference. We first define the \textit{bias} of the single parameter to describe its changes after aggregation as below.
\begin{definition} (Parameter $j$'s Bias $\xi_j$) The bias of the parameter $j$ in the global model is the distance between the parameter of the benign global model ($W_j$) and that of the poisoned global model ($\overline{W}_j$), denoted as $\xi_j=|W_j-\overline{W}_j|$.
\label{xi_df}
\end{definition}
We use the absolute value to measure the bias considering both the direction and magnitude changes.
Then we provide the calculation methods of $\xi_j$ for two classical types of AGMs mentioned above.

\begin{theorem} Given the scalar of parameter $j$ for malicious client $i$ (i.e., $\alpha_{i,j}$), $g_i$ and $\overline{g}_i$ for benign and poisoned local models, respectively,  $\xi_j$ is approximately calculated as:
\begin{align}
\xi_j\approx\begin{cases}
|g_i-\overline{g}_i\alpha_{i,j}|,&\text{weight-based AGM},\\
|1-\alpha_{i,j}|,& \text{statistics-based AGM},
\end{cases}
\nonumber
\end{align}
where $\alpha_{i,j}$ is the variable.
\label{xi_j}
\end{theorem}

Please see Appendix \ref{proof_xi_j} for proofs. MPAs essentially find a suitable scalar for each parameter of the local model, i.e., $\alpha_{i,j}$. Thus, we can control the bias $\xi_j$ by modifying the scalars. 
These two functions are all piecewise functions, so it is necessary to consider the domain of variables in using them to make attack decisions. We argue that MPAs should enlarge the bias of each parameter as much as possible to increase the overall bias of the global model.
Then, we have the following definition for the goal after aggregation.

\begin{definition}(Goal of MPA After Aggregation) The goal of MPA after global model aggregation is to maximize the bias of global model $\xi=\sum_{j=1}^J \xi_j$.  
\label{xi_d}
\end{definition}

\subsection{General Model of I\&B}
We argue that the MPA should maximize both the importance of the poisoned local model and the bias of the global model to enlarge the poisoning effects.
Thus, we conclude the novel quantitative analysis paradigm of MPAs as below:
\begin{align}
\textbf{I\&B:} \quad &\mathop{\max}_{\alpha_{i,j}}\quad(\overline{g}_i, \xi),\notag
\end{align}
with certain constraints according to applied AGMs.
This is clearly a multi-objective optimization problem with multiple variables, which is hard to be solved. In practice, we can significantly decrease the difficulty of solving this problem by transforming the optimization objectives and reducing the variables as presented in Section \ref{faker}. 

 We argue that I\&B is different from the existing frameworks \cite{fang2020local,shejwalkar2021manipulating,bagdasaryan2020backdoor}. First, I\&B needs to maximize both importance and bias, rather than a balance of these two goals; and importance is not only related to the acceptance of the models but also determines the impact of the local models on the global model. Besides, I\&B is the first framework to demonstrate the effects of MPAs quantitatively, achieving controllable attacks. 

With the assistance of I\&B, the attackers can know what kind of poisoned models should be submitted in order to make the attack more effective. However, since local models are usually high-dimensional and similarity-based AGMs usually set some similarity requirements to filter out malicious submissions,
efficiently generating poisoned models that can satisfy the similarity requirements is hard.
Therefore, a poisoned local model generator ensures that the submissions can remain undetected and also maintain attacking effectiveness is required.
\fi

\section{The ``Curse'' of Similarity}\label{sec_EvilTwin} 
In this section, we reveal an unexplored security threat that measuring the reliability of local models with the calculated similarity is insecure. 
We mainly discuss the vulnerabilities of three representative similarity metrics, i.e., $L_2$ norm, Euclidean distance, and cosine similarity, by both intuitive explanation and theoretical analysis, followed by discussions about manipulating their vulnerabilities to undermine FL systems. 
\if()
Similar to Hash functions, the functions of similarity metrics 
are vulnerable to collision attack. In other words, there could be multiple different inputs that have the same output value.  Considering that these similarity metrics are quadratic functions involving calculating the square sum of the parameters in local models, we can carefully modify the parameters' values to enable the calculation results to satisfy the given requirements. 
\fi
\subsection{Intuitive Explanation}
\begin{figure}[htpt]
\centering
\includegraphics[width=0.5\textwidth]{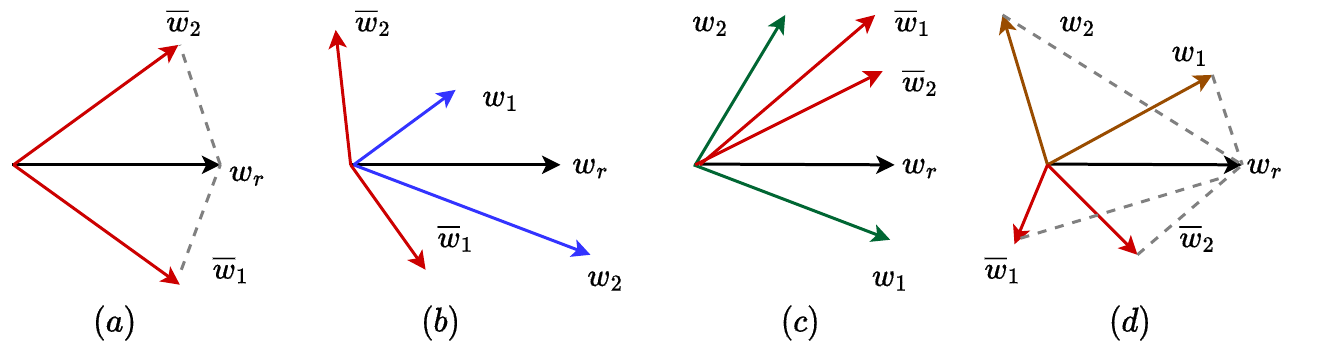}\caption{The vulnerabilities of similarity metrics in the two-dimensional plane. (a) shows the general similarity metrics' vulnerabilities; (b)-(d) illustrate the vulnerabilities of $L_2$ norm, cosine similarity, and Euclidean distance, respectively. 
}
\label{eviltwin_v}
\end{figure}

\if()
\begin{figure*}
\centering
\includegraphics[width=0.98\textwidth]{eviltwin_flow (13).pdf}
\caption{The illustration of EvilTwin with $J_t=2,~t\in \{1,\cdots,T\}$. Steps 1 and 2 illustrate the partitions of reference vector $Y$ and poisoned vector $\overline{X}$; step 3 presents the processes of scalar calculation and obtaining elements in $\overline{X}$. Specifically, $\overline{v}_t=\alpha_t v_t$ with $\alpha_t$ being the unknown shared scalar in each group, and then we can solve $S(\overline{V}, V)$ to derive the shared scalars which will be multiplied with the corresponding elements in $Y$ to obtain the poisoned elements in $\overline{X}$.}
\label{eviltwin_to}
\end{figure*}
\fi

Assume there are two benign local models $w_1$ and $w_2$ matching the lower and upper bounds of the similarity, respectively, two poisoned local models $\overline{w}_1$ and $\overline{w}_2$, and a reference model $w_r$ on the serve. The reference model is a hypothetical model that the defender uses to assist in model evaluation. 
In a more intuitive way, we demonstrate the vulnerabilities of similarity metrics in Figure \ref{eviltwin_v}.  
 In Figure \ref{eviltwin_v}(a), we allow $\overline{w}_1$ and $\overline{w}_2$ share the same value of $L_2$ norm and have the same cosine similarity and Euclidean distance with $w_r$;  however, they are in different directions. This means that local models that have the same similarity can be different. In Figure \ref{eviltwin_v}(b), poisoned local models $\overline{w}_1$ and $\overline{w}_2$ are evaluated as qualified but they are in different directions, and the same phenomena can be spotted in both in Figure \ref{eviltwin_v}(c) and Figure \ref{eviltwin_v}(d), revealing that multiple local models, which are significantly different, could be treated as qualified local models since their evaluated similarities are within the allowed range. In this way, the attackers can submit a well-crafted poisoned local model that satisfies the similarity requirements to undermine the global model.
 \if()

We believe that this approach is unsafe: firstly, the upper and lower bounds are difficult to determine; moreover, according to what is shown in Figure \ref{eviltwin_v}(a), we can find multiple vectors that are different but have the same evaluated similarity. Thus, the robustness of similarity metrics can be undermined. Please see Appendix \ref{vul_sim} for theoretical analysis. 
\fi
\subsection{Theoretical Analysis}
Though Figure \ref{eviltwin_v} illustrates the vulnerabilities of similarity metrics in FL, the mathematical principles behind the phenomenon are still unclear, which will be elaborated on next. 

In general, when there is no attack, submissions from clients are supposed to have similar parameter values; however, when the FL system is under attack, the effective poisoned local models should have significantly different parameter values compared with the benign ones. Thus, we can summarize the vulnerabilities of similarity metrics below.
 
\begin{definition}
(The ``Curse'' of Similarity)
    Given the benign local model $w_i$, the reference model $w_r$, and the similarity lower and upper bounds $s_l$ and $s_u$, the attacker can carefully craft a poisoned local model $\overline{w}_i$, so that $S(\overline{w}_i,w_r)\in[s_l,s_u]$ while the parameter values in $w_i$ and $\overline{w}_i$ are significantly different.
\end{definition}

To generate such a stealthy poisoned local model, a reasonable model for reference is required. Although $w_r$ is the optimal one, it is not accessible to the attacker. Recall the discussions about the capabilities and knowledge of attackers, they only know their own trained local models instantly and the global model in the previous round. Compared to the to-be-updated global model, the current local model $w_i$ is much closer to the reference model. So the attacker can use $w_i$ to approximate $w_r$ to generate the poisoned local model, i.e., $\overline{w}_i=\alpha_i \otimes w_i$ where $\otimes$ implies that $\overline{w}_{i,j}=\alpha_{i,j} w_{i,j}$ and $\alpha_i=(\alpha_{i,1},\cdots,\alpha_{i,j},\cdots,\alpha_{i,J})$ is the vector of scalars. However, how to ensure that there exists such an effective vector of scalars $\alpha_i$ remains another challenge. We provide the following theorem regarding this problem.
\begin{theorem}
There exists at least one combination of scalars in $\alpha_i$ for local model $w_i$, $\forall \alpha_{i,j}>0, \exists \alpha_{i,j} \neq 1$, so that the attacker can generate a stealthy poisoned local model $\overline{w}_i$ based on $w_i$, i.e., $\overline{w}_i=\alpha_i \otimes w_i$.     
\end{theorem}


The detailed proof is shown in Appendix \ref{the_1}. With the above theorem, the attacker only needs to find such a vector to launch the attack. As for the method of obtaining $\alpha_i$, it will be detailed in Section \ref{faker}.

From the above analysis, we know the deficiencies of the similarity metrics and the evidence for the existence of such deficiencies from a general perspective. However, the reasons for the prevalence of these defects in FL have not been explained. The main reason is that local models in FL usually have high dimensionality, and the similarity calculation considers the model as a whole but does not measure the value of each parameter, leaving room for attackers to design suitable scalars for different parameters to satisfy the similarity requirements. Besides, a given similarity requirement is usually not effective in detecting all aspects of the model. In particular, it is difficult for a single metric to measure both direction and magnitude. Although some defenses adopt multiple similarity metrics to evaluate the models, such as FLTrust and FLAME, we can still devise optimal attack strategies by transferring the attacks into optimization problems (see Section \ref{faker_fl}). 

\subsection{Manipulating Similarity in FL}
Since we use the local model $w_i$ to approximate the reference model $w_r$, we can let $S(\overline{w}_i,w_i)$ as the approximate similarity of the poisoned local model $S(\overline{w}_i,w_r)$ during evaluation, which is denoted as $\overline{s}_i$. 
As an attacker, once the similarity metrics used by the defender are known, a poisoned local model can be generated based on $w_i$, which needs to satisfy the similarity requirements, i.e., the evaluated similarity $\overline{s}_i$ needs to be in the range of $[s_l,s_u]$. Please note that such a range is only for the convenience of expression, it does not mean that all the defenses have strict upper and lower bounds of similarity, and the specific similarity requirements should be determined by analyzing different defenses.  In general, the closer $\overline{s}_i$ is to the upper limit of its theoretical value, the more similar $\overline{w}_i$ is to $w_r$ and the less likely to eliminate $\overline{w}_i$. Thus, we can allow the attacker to maximize $\overline{s}_i$ in practice. 

\textbf{Note:} Though the attacker can generate a stealthy poisoned local model, the attacking effectiveness cannot be guaranteed. Therefore, we have to explore maximizing the attack performance of such a poisoned local model generated based on the similarity metrics' vulnerabilities, which will be introduced in the next section.


\if()
\subsection{General Model of EvilTwin} \label{ge_ev}
[] and [] use an iterative approach, allowing the same scalar to be applied to each parameter in $w_i$ to obtain such a poisoned local model $\overline{w}_i$; however, this method has a low success rate and can be computationally expensive. In addition, one may employ some heuristic algorithms to search for a suitable scalar for each parameter, but the models tend to have high dimensionality which makes this method impractical. We propose a novel and efficient method, EvilTwin, to forge vectors based on the vulnerabilities of similarity metrics. 

The workflow of EvilTwin is summarized in \textbf{Algorithm} \ref{al_ev} with detailed explanations below. Besides, an intuitive use case that explains \textbf{Algorithm} \ref{al_ev} line by line is provided in Appendix \ref{case_evil}.

\if()
\begin{algorithm}
\caption{EvilTwin}
\label{al_1}
\begin{algorithmic}[1]
\REQUIRE $X$, $Y$, $J$, $S(\cdot, \cdot)$, $T$
\ENSURE $\overline{X}$
\STATE $\theta \leftarrow S(X,Y)$
\STATE Partition $Y$ into $T$ groups
\FOR{$t \in \left\{ 1,2,\cdots, T\right\} $}
\STATE $v_t$ $\leftarrow$ $\sum_{y_{t,j}\in Y_t} y_{t,j}^2$ 
\STATE $\overline{v}_t\leftarrow\alpha_{t} v_t$ $\textcolor{blue}{\rhd}$ $\alpha_t$ is the unknown shared scalar
\STATE Append $v_t$ and $\overline{v}_t$ into $V$ and $\overline{V}$, respectively
\ENDFOR
\STATE  $(\alpha_1,\cdots, \alpha_t,\cdots, \alpha_T)$ 
$\leftarrow$ Solve $S(\overline{V},V)=\theta$
\STATE $\overline{x}_j=\alpha_t y_{t,j}, \forall 1\leq t \leq T, 1\leq j \leq J_t$
\RETURN $\overline{X}=(\overline{x}_1,\cdots,\overline{x}_j,\cdots,\overline{x}_J)$
\end{algorithmic}
\label{al_ev}
\end{algorithm}
\fi

\begin{algorithm}
\caption{EvilTwin}
\label{al_1}
\begin{algorithmic}[1]
\REQUIRE $w_i$, $w_i$, $J$, $S(\cdot, \cdot)$, $T$
\ENSURE $\overline{w}_i$
\STATE $\theta \leftarrow S(w_i,w_i)$
\STATE $w_{r,1},\cdots,w_{r,t},\cdots,w_{r,T} \leftarrow$ Partition $w_i$ into $T$ groups
\STATE Assign the parameters to the shared scalar $\alpha_{i,t}$ in the same group $\textcolor{blue}{\rhd}$ $\alpha_{i,t}\in (\alpha_{i,1},\cdots,\alpha_{i,T})$ is the variable
\STATE $\overline{w}_i\leftarrow(\overline{w}_{i,1},\cdots, \overline{w}_{i,J})$ where $\overline{w}_{i,j} = \alpha_{i,t} w_{i,j}$ if $j\in w_{r,t}$
\STATE  $(\alpha_{i,1},\cdots, \alpha_{i,t},\cdots, \alpha_{i,T})$ 
$\leftarrow$ Solve $S(\overline{w}_i,w_i)=\theta$
\STATE Obtain $\overline{w}_i$ with the calculated scalars by following Line 4
\RETURN $\overline{w}_i$
\end{algorithmic}
\label{al_ev}
\end{algorithm}

First, we need to determine the similarity requirement $\theta$, which can be calculated by $S(w_i,w_i)$ directly (\textbf{Line} 1). To overcome the ``curse" of dimensionality and calculate the scalars efficiently, we partition $J=|w_i|$ elements in $w_i$ into $1\leq T\leq J$ groups $w_{r,1},\cdots,w_{r,t},\cdots,w_{r,T}$ (\textbf{Line} 2). Please note that the partitioned groups are disjoint and $\cup_{t=1}^T w_{r,t} = w_i$. Accordingly, we define an unknown shared scalar $\alpha_{i,t} \in(\alpha_{i,1},\alpha_{i,2},\cdots,\alpha_{i,T})$ for each group (\textbf{Line} 3). In this way, instead of calculating $J$ scalars separately, we only need to derive $T$ scalars, thus significantly reducing the computational complexity when $T\ll J$. Then, we can express the $j$th parameter in group $t$ of $\overline{w}_i$ as $\overline{w}_{i,j} = \alpha_{i,t} w_{i,j}$ so that to get the rewrite poisoned local model $\overline{w}_i$  (\textbf{Line} 4).  Taking the scalar calculation for Euclidean distance as an example,  $E(\overline{w}_i,w_i)=\theta$ can be expressed as $[\sum_{t=1}^J (\overline{w}_{i,j}-w_{i,j})^2]^\frac{1}{2}=[\sum_{t=1}^T ({\alpha_{i,t}}-1)^2 v_t]^\frac{1}{2}=\theta$ where $v_{i,t}=\sum_{j=1}^{|w_{r,t}|} w_{i,j}^2$ is the parameter sum in group $t$. 
Next, we only have to solve $S(\overline{w}_i,w_i)=\theta$ to obtain the shared scalars for different groups (\textbf{Line} 5).
For different similarity metrics, the calculation methods are different, and we detail the methods of \textbf{Line} 5 in Section \ref{ev_sim}. Finally, by following the expression method in \textbf{Line} 4, we can get the poisoned local model $\overline{w}_i$ directly (\textbf{Lines} 6-7). The computational complexity of EvilTwin is $\mathcal{O}(T)$.

\if()
Given vectors $X$ and $Y$ with $J$ elements, our goal is to generate another vector $\overline{X}\neq X$ with the same length to satisfy $ S(X,Y)=S(\overline{X},Y)=\theta$. The similarity requirement $\theta$ is measured by $S(X,Y)$ directly (\textbf{Line} 1). Let $\overline{x}_j=\sigma_jy_j$ where $\sigma_j$ is the scalar for each element in $Y$, and the essential task of this algorithm is to calculate all the scalars,  ensuring  $S(\overline{X}, Y)=\theta$. 
However, as the number of dimensions in the reference vector $Y$ increases, it is difficult to search qualified scalars in a complicated space. To overcome the ``curse" of dimensionality and calculate the scalars efficiently, we first partition $J$ elements in $Y$ into $1\leq T\leq J$ groups $Y_1,\cdots,Y_t,\cdots,Y_T$ with the sizes of $J_1,\cdots,J_t,\cdots,J_T$,  i.e., $Y_t=(y_{t,1},\cdots,y_{t,j},\cdots,y_{{t,J_t}})$  (\textbf{Line} 2). Please note that the partitioned groups are disjoint and $\cup_{t=1}^T Y_t = Y$. Accordingly, we define an unknown shared scalar $\alpha_t \in(\alpha_1,\alpha_2,\cdots,\alpha_T)$ for each group; thus, we have $\sigma_j=\alpha_t$ for all $\sigma_j$'s in group $t$. In this way, instead of calculating $J$ scalars separately, we only need to derive $T$ scalars, thus significantly reducing the computational complexity when $T\ll J$. Define $v_t=\sum_{y_{t,j}\in Y_t} y_{t,j}^2$  as the square sum of elements in each group and let $V=(v_1,\cdots,v_t,\cdots,v_T)$ be the vector that contains all square sums.  Similarly, we define $\overline{v}_t=\alpha_t v_t$ and denote $\overline{V}=(\overline{v}_1,\cdots,\overline{v}_t,\cdots,\overline{v}_T)$ (\textbf{Lines} 3-7). Then we can resolve $S(\overline{V},V)=\theta$ to derive the reduced-size scalar vector with $T$ elements. Via multiplying the shared scalars with the corresponding groups of elements in $Y$, i.e., $\overline{x}_j=\alpha_t y_{t,j}$ for any $1\leq t\leq T$ and $1\leq j\leq J_t$, we can return the fake vector $\overline{X}$ with the required similarity (\textbf{Lines} 8-10). As for the methods of solving $S(\overline{V}, V)=\theta$ in \textbf{Line} 8, we will introduce them in Section \ref{ev_sim} for specific similarity metrics. The computational complexity of EvilTwin is $\mathcal{O}(T)$.
\fi

\subsection{EvilTwin for Different Similarity Metrics}\label{ev_sim}
To solve $S(\overline{w}_i,w_i)=\theta$ (\textbf{Line} 5 in \textbf{Algorithm} \ref{al_ev}), one way is to get the numerical expression of scalars, but there is not enough information for solving such an equation with $T$ variables. Our approach is as below. Our approach is based on the intuitive idea that by first assigning values to $T-1$ variables, the value of the remaining one can be derived by submitting the $T-1$ variables that have already been assigned values. The challenge is how to assign values to $T-1$ variables in advance. We can assume that these variables are known, and then derive their domain $\mathcal{F}$. We denote $\alpha_{i,-t}\in \mathcal{F}$ as any scalar except $\alpha_{i,t}$, i.e., $\alpha_{i,-t} \in (\alpha_{i,1}, \cdots, \alpha_{i,t-1}, \alpha_{i,t+1},\cdots, \alpha_{i,T})$. Similarly, we let $v_{r,-t}$ be any parameter sum except $v_{r,t}$, i.e., $v_{r,-t} \in (v_{r,1},\cdots, v_{r,t-1}, v_{r,t+1},\cdots, v_{r,T})$. Thus, we can allow $\alpha_{i,-t}\in\mathcal{F}$ and remain $\alpha_{i,t}$ as the variable. This has the advantage of giving a mathematical expression without multiple iterations, which can significantly reduce the computational cost. For simplicity, we directly present the following three theorems to get one variable $\alpha_{i,t}$ and $T-1$ scalars $\alpha_{i,-t}$ in $\mathcal{F}$. Please refer to Appendix \ref{proofs} for detailed proofs.


\begin{theorem} (EvilTwin for $L_2$ Norm)
By applying EvilTwin to $L_2$ norm with the similarity requirement ${\theta}\geq 0$, we have $\theta=(\sum_{t=1}^{T} \alpha^2_{i,t} v_{r,t})^\frac{1}{2}$;
let $\alpha_{i,-t}\in \mathcal{F}=(-[\frac{\theta^2}{(T-1)\max(v_{r,-t})}]^\frac{1}{2},[\frac{\theta^2}{(T-1)\max(v_r,{-t})}]^\frac{1}{2})$, and we can get $\alpha_{i,t}=\pm (\frac{1}{v_{r,t}}[\sum_{-t=1}^{T-1} (-\alpha^2_{i,-t} v_{r,-t})+{\theta}^2])^\frac{1}{2}$.
\if()
\begin{align}
\alpha_{i,t}=\pm (\frac{1}{v_{r,t}}[\sum_{-t=1}^{T-1} (-\alpha^2_{i,-t} v_{r,-t})+{\theta}^2])^\frac{1}{2}.\notag
\end{align}
\fi
\label{th_l}
\end{theorem}

By observing the equation $\theta=(\sum_{t=1}^{T} \alpha^2_{i,t} v_{r,t})^\frac{1}{2}$, we can see that when $\alpha_t$ is smaller (i.e., $\alpha_{i,t} \to 0$), $\theta$ will be smaller. Thus, in practice, we can adjust the value of scalars to meet the similarity requirements.
\begin{theorem} (EvilTwin for Euclidean Distance)
By applying EvilTwin to Euclidean distance with the similarity requirement $\theta\geq 0$,  we can get $\theta=[\sum_{t=1}^T (\alpha^2_{i,t}-2\alpha_{i,t}+1) v_{r,t}]^\frac{1}{2}$; let $\alpha_{i,-t} \in \mathcal{F}= [1-\frac{\theta}{[(T-1)\max(v_{r,-t})]^\frac{1}{2}}, 1+\frac{\theta}{[(T-1)\max(v_{r,-t})]^\frac{1}{2}}]$, and we can have $\alpha_{i,t}= \frac{1}{v_{r,t}}(v_{r,t}\pm [v_{r,t}\big[\sum_{-t=1}^{T-1}[(2\alpha_{i,-t}\notag-\alpha^2_{i,-t}-1)v_{r,-t}]+{\theta}^2\big]^\frac{1}{2}).$
\if()
\begin{align}
    \alpha_{i,t}&= \frac{1}{v_{r,t}}(v_{r,t}\pm [v_{r,t}\big[\sum_{-t=1}^{T-1}[(2\alpha_{i,-t}\notag-\alpha^2_{i,-t}-1)v_{r,-t}]+{\theta}^2\big]^\frac{1}{2}).\notag
\end{align}
\fi
\label{t_cs}
\end{theorem}

According to $\theta=[\sum_{t=1}^T (\alpha^2_{i,t}-2\alpha_{i,t}+1) v_t]^\frac{1}{2}$, if the values of the scalars are closer to 1, the distance is smaller and thus the similarity is higher; but if the values of the scalars are far from 1, $\theta$  would be larger, indicating a lower similarity.

\begin{theorem} (EvilTwin for Cosine Similarity)
By applying EvilTwin to cosine similarity with the similarity requirement $\theta\in[-1,1]$, we have $\theta=\frac{\sum_{t=1}^T\alpha_{i,t} v_{r,t}}{(\sum_{t=1}^T v_{r,t})^\frac{1}{2} (\sum_{t=1}^T \alpha^2_{i,t} v_{r,t})^\frac{1}{2}}$;
let $\alpha_{i,-t}\in \mathcal{F}$ where $\mathcal{F}$ is rational but does not include $0$, and we can have
\begin{align}
    &\alpha_{i,t}=\frac{1}{v_{r,t}({\theta}^2 \sum_{t=1}^{T}v_{r,t} -v_{r,t})}\bigg (v_{r,t}\sum_{t=1}^{T-1}\alpha_{i,-t} v_{r,-t}\notag \\
    &\pm \theta\bigg[-v_{r,t}\sum_{t=1}^{T} v_{r,t} \big[ \sum_{-t=1}^{T-1}\alpha^2_{i,-t} v_{r,-t}^2({\theta}^2-1)\notag\\ 
    &+\sum_{-t=1}^{T-1}\alpha^2_{i,-t}{\theta}^2\zeta_{r,t}-v_{r,t}\sum_{-t=1}^{T-1} \alpha^2_{i,-t}v_{r,-t}-2\pi
    \big]\bigg]^\frac{1}{2}
    \bigg),
    \label{e_cs}
\end{align}\label{th_3}
where $\zeta_{r,t}=\sum_{-t=1}^{T}v_{r,t} v_{r,-t}$ and $\pi=\sum \alpha_{i,-t}v_{r,-t}* \alpha_{i,-t'}v_{r,-t'}$ with $-t,-t'\neq t$ being any two different indexes.
\end{theorem}

We can see that the computational cost of $\alpha_{i,t}$ is higher if $T$ is larger because the calculation time of $\pi$ will increase a lot. Please note that we have no requirements for the domain of $\alpha_{i,-t}$ since it is flexible. Besides, if the variance of $\alpha_{i,t}$ is smaller, the similarity requirement $\theta$ will be larger.

According to the above theorems, we can calculate the scalars for different similarity metrics and then obtain the parameters in poisoned local model $\overline{w}_i$. Please see Appendix \ref{dev} for more experiments about EvilTwin.
Though EvilTwin can generate a fake model with a given similarity to the reference model, the effectiveness of the attack cannot be guaranteed. In addition, some defenses use more than just one similarity metric, so we still need to explore how EvilTwin can be utilized.
\if()
\subsection{Effects of Model Poisoning Attack}

Unlike the existing studies, we analyze the MPA from the perspective of the global model rather than the local model. 

In general, there are two methods to modify the global model: i) the most straightforward way is to change its direction, which can be done by changing the directions of the parameters from positive to negative or vice versa; and ii) another way is to modify the magnitude, which can be realized by scaling the parameters up or down in the same direction. In practice, these two methods can work together to degrade the performance of the global model. We design the following experiments to explore how the performance can be influenced by modifying the global model. We use a scalar to modify all parameters of the global model obtained in each round.
The detailed experimental settings are described in Section \ref{exp_se}, and the results are shown in Table \ref{ifluence_gm}. When the absolute value of the scalar is large,  we find that all AGMs cannot converge well, i.e., the error rate is larger when the absolute value of the scalar is larger. However, counter-intuitively, when applying FedAvg, norm-clipping, and FLTrust with a scalar equal to -1.00, i.e., changing the direction of the global model completely without changing its magnitude, the performance is comparable to that without this change (i.e., when the scalar is 1.00). Therefore, we draw the conclusion that a successful MPA should consider not only changing the local model's direction and magnitude but also how the global model is affected.

\begin{table}[H]
\centering
\caption{Influence of modifying global model, which is measured by test error rates with 100 clients in total, and each client is assigned five labels of training data. The experiments are based on MNIST.}
\begin{tabular}{|c|c|c|c|c|}
\hline
Scalars & FedAvg & Krum & Norm-clipping & FLTrust \\ \hline
-10.00  & 0.48   & 0.90 & 0.48          & 0.83    \\ \hline
-1.00   & \textbf{0.10}   & 0.90 & \textbf{0.10}          & \textbf{0.10}    \\ \hline
-0.50   & 0.25   & 0.90 & 0.25          & 0.36    \\ \hline
0.50    & 0.21   & 0.90 & 0.21          & 0.37    \\ \hline
1.00    & 0.11   & 0.48 & 0.11          & 0.09    \\ \hline
1.50    & 0.22   & 0.90 & 0.22          & 0.26    \\ \hline
10.00   & 0.44   & 0.90 & 0.44          & 0.46    \\ \hline
\end{tabular}
\label{ifluence_gm}
\end{table}

Note that the above experiments are conducted on the global model, but a local attacker cannot modify the global model directly. Therefore, it is challenging to design an MPA that can significantly impact the global model.

\fi
\fi

\section{Faker: ``Similar'' but Harmful}\label{faker}
\begin{figure*}
\centering
\includegraphics[width=0.80\textwidth]{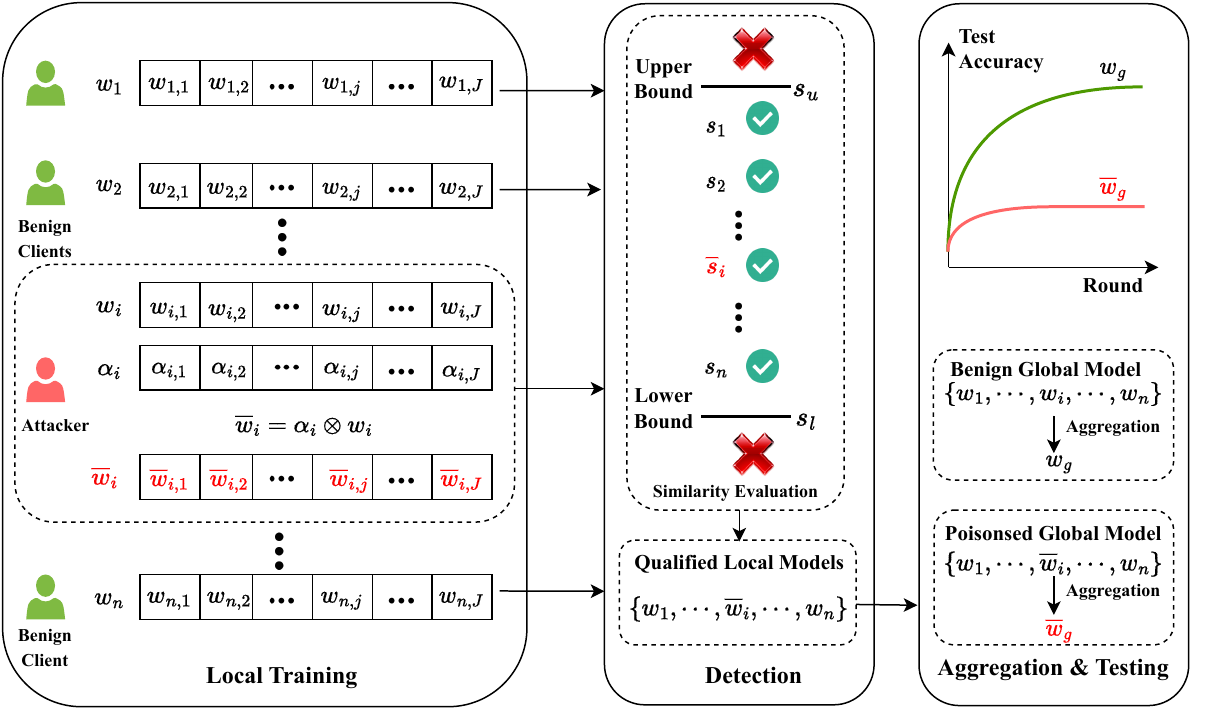}
\caption{The illustration of Faker for the single attacker. After training, the attacker obtains local model $w_i$, and it launches the attack by calculating the scalars and generating poisoned local model $\overline{w}_i$ according to \textbf{Algorithm \ref{al_faker}}; when the submissions are evaluated by similarity-based defenses, the poisoned local model generated by Faker can avoid being detected and the poisoned global model will be significantly different from the benign one, degrading the test accuracy. }
\label{ill_faker}
\end{figure*}

We present a novel and effective model poisoning attack coined \textit{Faker} by exploiting the deficiencies of similarity metrics to undermine FL. The core idea of Faker is to find an effective vector of scalars, i.e., $\alpha_i$, for the attacker to generate the poisoned local model $\overline{w}_i$ based on $w_i$. 
Specifically, Faker enables $\overline{w}_i$ to pass the detection of similarity-based defenses while maximizing its negative impacts on the global model. To generate such a poisoned local model, the attacker needs to ensure that the model will be recognized as ``similar'' but significantly different from the benign model in parameter values. Next, we introduce how Faker achieves both goals and formulate the general model of Faker. The illustration of Faker for a single device is in Figure \ref{ill_faker}. Note that Faker is designed for a single malicious client, meaning that cooperation among multiple attackers is not required, while we also provide strategies for the cooperation case in Section \ref{attack_mode}.

\subsection{Model Formulation}

As we have discussed in Section \ref{sec_EvilTwin}, the attacker uses the benign local model $w_i$ as an approximation to the unknown reference model $w_r$, and in order to avoid the bias caused by such approximation, the attacker can maximize the similarity between the generated poisoned local model and the benign local model, i.e., $\overline{s}_i$. If there is only one similarity metric being applied, we only need to calculate the similarity of $\overline{w}_i$ directly by $\overline{s}_i=S(\overline{w}_i, w_i)$; otherwise, we can multiply the calculated multiple similarities. For example, the similarity of $\overline{w}_i$ is $\overline{s}_i=C(\overline{w}_i,w_i)\frac{L(\overline{w}_i)}{L(w_i)}$ when FLTrust is applied and $\overline{s}_i=E(\overline{w}_i,w_i)$ when Krum is adopted. The reason for multiplying these similarities is that multiplied result is more convenient as an optimization objective than maximizing them separately. In this way, the first goal of Faker, i.e., generating ``similar'' poisoned local models, can be satisfied by maximizing $\overline{s}_i$. Since $\overline{w}_{i}=\alpha_{i}\otimes w_{i}$, we know that $\overline{s}_i=S(\overline{w}_i,w_i)$ is a function of the scalars in $\alpha_i$. 

Intuitively, if two models are not the same, their parameters are different; and if we want to make the two models even more distinct, we can enlarge the difference in their parameters. We use the absolute value of the difference of the parameter values to represent the difference of the parameters, i.e., $\overline{\delta}_{i,j}=|\overline{w}_{i,j} - w_{i,j}|=|w_{i,j}(\alpha_{i,j}-1)|$. Thus, the difference between $\overline{w}_i$ and $w_i$ can be written as $\overline{\delta}_i=\sum_{j=1}^{J}{\overline{\delta}_{i,j}}=\sum_{j=1}^J|w_{i,j}(\alpha_{i,j}-1)|$.  To achieve another goal of the attacker, which is to have significant differences between the submitted poisoned model and the benign one, we need to maximize $\overline{\delta}_i$. Since $w_{i,j}$ is a constant and $\alpha_{i,j}$ is positive, we allow $\overline{\delta}_i\approx \sum_{j=1}^J \alpha_{i,j}$ to serve as another maximization objective for the attacker to simplify the computation. 

According to the above analysis, the attacker has to maximize both $\overline{s}_i$ and $\overline{\delta}_i$ by optimizing the scalars in $\alpha_{i}$ simultaneously. To further simplify the optimization problem, we let $f(\alpha_{i})=\overline{s}_i \overline{\delta}_i$ represent the overall optimization objective. Thus, we can propose a general model of Faker for a single attacker $i$ as follows:
\begin{align}
\textbf{P0:} \quad &\mathop{\arg\max}_{\alpha_{i}} \quad f(\alpha
_{i}) \notag \\
s.t.\quad
&C1:~\overline{s}_i\in [s_l,s_u]\notag,\\
&C2: ~ \forall \alpha_{i,j}>0, \exists \alpha_{i,j}\neq 1, \notag
\end{align}
where $f(\alpha_{i})=\overline{s}_i \overline{\delta}_i$ is the optimization objective function; 
$\overline{s}_i$ and $\overline{\delta}_i$ are the similarity and difference of poisoned local model $\overline{w}_i$, respectively; the scalars $\alpha_{i,1},\cdots,\alpha_{i,j},\cdots,\alpha_{i,J}$ are the optimization variables, and there are $J$ variables in total; $C1$ is the similarity requirement of the adopted defense, which means that the poisoned local model should satisfy the requirement to avoid being detected; and $C2$ is the feasible domain of $\alpha_{i,j}$. Since different defenses adopt different similarity metrics and have different rules, the attacker has to adapt $\textbf{P0}$ to specific defenses. The two main steps for the attacker to launch Faker are to formulate $f(\alpha_{i})$ and determine the bounds of $\overline{s}_i$, which will be detailed when we design Faker against different defenses in Section \ref{f_a}.

$\textbf{P0}$ is a non-linear optimization problem with inequity constraints and $J$ variables in total, and thus solving it is nontrivial. To enable a quick decision for the attacker, we intend to avoid using 
computationally intensive solutions, such as heuristic algorithms, distributed optimization algorithms, and reinforcement learning based algorithms. We propose a simple yet effective strategy to reduce the complexity of solving $\textbf{P0}$ by choosing only one scalar, e.g., $\alpha_{i,j}$, as a variable, while any scalar except $\alpha_{i,j}$ in $\alpha_i$ is set as a constant from a well designed feasible domain. Note that we need to set values for $J-1$ scalars except $\alpha_{i,j}$ in $\alpha_i$, and for convenience of expression, we use $\alpha_{i,-j}$ and $w_{i,-j}$ to denote any such scalar and the corresponding parameter in $w_{i}$, respectively.

\if()
In this part, we introduce the process of launching Faker on a single local device. 
Denote $\mathcal{N}$ as the set of all the local clients with the total number $n$ and $\mathcal{M}$ as the set of malicious local clients with the total number $m$. In the following, client $i$ is from $\mathcal{M}$ if there is no other specification. 
We first have to transfer launching the MPA into a model faking problem according to EvilTwin. We first term the reference model as $w_i$. Typically, the reference model is used on the server side to assist in evaluating local models, which is not available to the local devices and has to be approximated by the attacker.  
Our method is to approximately represent $w_i$ using the original local model of each attacker, i.e., $w_i \approx w_i$. This allows a single attacker to launch MPA on its own without knowing the local models of other attackers. 
To enhance the effectiveness of this approximation in the non-IID scenario in practice, we propose an adjustment method in Section \ref{apro_theta}.

The pseudocode of Faker is summarized in \textbf{Algorithm} \ref{al_faker}.
Based on the analysis in Section \ref{i_b}, we know that the goals during and after aggregation will determine the impacts of the MPA on the global model, and Faker should enlarge the importance of the poisoned local model (i.e., $\overline{g}_i$) and the bias between poisoned and benign global models (i.e., $\xi$). So as the first step, the attacker can calculate $\overline{g}_i$ and $\xi$ according to I\&B and the corresponding AGM (\textbf{Line} 1), which will be detailed when designing specific attacks against similarity-based AGMs. 
Since $w_i\approx w_i$, the attacker can divide its local model $w_i$ into $T$ groups directly according to EvilTwin, and the square sum of all the parameters in each group is $\psi_{i,t}\in (\psi_{i,1}, \psi_{i,2},\cdots,\psi_{i,T})$ (\textbf{Line} 2). Faker has to find a shared scalar $\alpha_{i,t}\in(\alpha_{i,1},\alpha_{i,2},\cdots,\alpha_{i,T})$ for parameters in the same group. We  use $\alpha_{i,-t}$ to represent any scalar except $\alpha_{i,t}$ in $\alpha_i$, i.e., $\alpha_{i,-t} \in (\alpha_{i,1},\cdots, \alpha_{i,t-1},\alpha_{i,t+1},\cdots, \alpha_{i,T})$; similarly, we let $\psi_{i,-t}$ represent any square sum except $\psi_{i,t}$, i.e., $\psi_{i,-t} \in (\psi_{i,1},\cdots, \psi_{i,t-1}, \psi_{i,t+1},\cdots, \psi_{i,T})$. Then, the attacker has to rewrite $\overline{g}_i$ and $\xi$ with $\alpha_{i,t}$ and $\psi_{i,t}$. As for the modified expression of $\overline{g}_i$, we will present the details when designing Faker for each AGM. As for $\xi$, since the parameters in each group share the same scalar, we randomly choose one parameter in each group (i.e., $w_{t,j}$) and calculate the bias $\xi_{t,j}$ to approximately represent the bias of each group for simplicity, and we define $\xi=\sum_{t=1}^{T}\xi_{t,j}$ as the bias of the global model (\textbf{Line} 3). We do not define $\xi$ as the sum of the biases of all the parameters to reduce the computing complexity.  
Based on the above analysis, we can conclude the general model of Faker as below (\textbf{Line} 4).
\begin{align}
\textbf{P0:} \quad &\mathop{\arg\max}_{\alpha_{i,t}} \quad (\overline{g}_i,  \xi)\notag \\
\textbf{s.t.}\quad&C1:~\overline{w}_i=EvilTwin(w_i,w_i,J,S(\cdot,\cdot),T)\notag, i \in \mathcal{M},\\
&C2:~\overline{W}=AGM(w_1, \cdots, \overline{w}_i,\cdots,w_n), i \in \mathcal{N}\notag,\\
&C3:~|\alpha_{i,t}|>0, i \in \mathcal{M},\notag
\end{align}
where the objective of Faker is to maximize both $\overline{g}_i$ and $\xi$ with only one variable $\alpha_{i,t}$;  $C1$ shows that Faker generates poisoned local models based on EvilTwin and the similarity requirement $\theta$ is usually given; $C2$ is the aggregation process that indicates that Faker should meet the requirements of the applied AGM\footnote{The local models from benign clients are not required in Faker.}; $C3$ is the domain of $\alpha_{i,t}$ which requires the scalar should be non-zero since the parameters of local models are usually non-zero. 
By solving \textbf{P0}, we can get the vector that contains all shared scalars  $\alpha_i$, thus we can generate the poisoned local model by multiplying the shared scalars with the parameters in $w_i$ (\textbf{Lines}  5-7). The methods of solving $\textbf{P0}$ in \textbf{Line} 5 will be detailed in the next subsection.
\fi


\begin{algorithm}[ht]
\caption{Faker}
\label{al_1}
\begin{algorithmic}[1]
\REQUIRE The local model $w_i=(w_{i,1},\cdots, w_{i,j},\cdots, w_{i,J})$ and the adopted defense
\ENSURE The generated poisoned local model $\overline{w}_i$\\
\textcolor{blue}{The malicious client $i$ executes:}
\STATE $\alpha_i=(\alpha_{i,1},\cdots, \alpha_{i,j},\cdots, \alpha_{i,J}), \forall \alpha_{i,j}>0, \exists \alpha_{i,j}\neq 1\leftarrow$ Initiates $J$ scalars as the unknown variables  
\STATE $\overline{s}_i$, $\overline{\delta}_i$ $\leftarrow$ Expresses the similarity and difference of the poisoned local model $\overline{w}_i$ by $\alpha_{i,j}$ and $w_{i,j}$
\STATE $s_l, s_u\leftarrow$ Obtains the lower and upper bounds of similarity requirement by analyzing the adopted defense
\STATE Maximizes the objective function $f(\alpha_i)=\overline{s}_i \overline{\delta}_i$ that subjects to $\overline{s}_i \in [s_l,s_u]$ by optimizing $\alpha_{i,j}$

\STATE  $\alpha_i\leftarrow$ Solves the above optimization problem
\RETURN  $\overline{w}_i=\alpha_i \otimes w_i =(\overline{w}_{i,1},\cdots,\overline{w}_{i,j},\cdots,\overline{w}_{i,J})$
\end{algorithmic}
\label{al_faker}
\end{algorithm}

We summarize Faker as \textbf{Algorithm} \ref{al_faker}. Specifically, when launching Faker, an attacker only needs its own local model $w_i$ and the defense adopted by the defender. Sometimes some other information is needed, such as the global model $w_g$ from the previous round when attacking Krum. The goal of the attacker is to find a $J$-dimensional vector $\alpha_i$ to generate a poisoned local model $\overline{w}_i$. All scalars in $\alpha_i$ must be positive and cannot both have value 1 (\textbf{Line 1}).  Next, the attacker needs to express the similarity $\overline{s}_i$ and difference $\overline{\delta}_i$ of $\overline{w}_i$ using $\alpha_{i,j}$ and $w_{i,j}$ respectively (\textbf{Line 2}). The lower and upper bounds of the similarity requirement, i.e., $s_l$ and $s_u$, can be deduced from the adopted defense (\textbf{Line 3}). Then, the attacker formulates an optimization problem with maximizing the objective function $f(\alpha_i)=\overline{s}_i\overline{\delta}_i$ that subjects to $\overline{s}_i \in [s_l, s_u]$ by optimizing all the scalars in $\alpha_i$. As for the solutions to the formulated optimization problem, we will detail them in Section \ref{f_a} according to different defenses (\textbf{Line 5}). In the end, the attacker can generate $\overline{w}_i$ by $\overline{w}_i=\alpha_i \otimes 
 w_i$ and submit it to the server (\textbf{Line 6}). The complexity analysis is presented in Section \ref{complexity}.
\subsection{Faker Against Similarity-based Defenses}\label{f_a}
Due to limited space, we cannot present detailed designs for attacking all similarity-based defenses. We choose three representative defenses, i.e., FLTrust, Krum, and norm-clipping, to illustrate how to adjust $\textbf{P0}$ for different defenses. Since these three defenses cover the three similarity metrics mentioned earlier, adapting Faker to attack other defenses can also refer to the following designs. 
Appendix \ref{faker_other} provides the detailed designs of Faker against three other benchmark defenses, i.e.,  FLAME, DiverseFL, and ShieldFL.


\subsubsection{Faker Against FLTrust}\label{faker_fl}
FLTrust applies both cosine similarity and $L_2$ norm to protect FL, and we can calculate the poisoned local model's similarity as $\overline{s}_i=C(\overline{w}_i,w_i)\frac{L(\overline{w}_i)}{L(w_i)}$, and we can express it with $\alpha_{i,j}$ and $w_{i,j}$ as $\overline{s}_i=\frac{\sum_{j=1}^J w^2_{i,j} \alpha_{i,j}}{\sum_{j=1}^J w^2_{i,j}\alpha_{i,j}^2}$. In this way, the objective function can be written as:
\begin{align}
    f(\alpha_{i})=\frac{\sum_{j=1}^J w^2_{i,j} \alpha_{i,j} \sum_{j=1}^J \alpha_{i,j} }{\sum_{j=1}^J w^2_{i,j}\alpha_{i,j}^2}. \notag
\end{align}

\if()
\begin{align}
    \overline{s}_i 
    &=\frac{\sum_{j=1}^J w^2_{i,j} \alpha_{i,j}}{\sum_{j=1}^J w^2_{i,j}\alpha_{i,j}^2}.\notag
\end{align}

Based on  (\ref{fltrust}), we know that the importance $\overline{g}_i$ is determined by $ReLU(C(\overline{w}_i,w_i))\frac{L(w_i)}{L(\overline{w}_i)}$ while $\frac{1}{\sum_{i=1}^{n} ReLU(C(\overline{w}_i,w_i))}$ is shared by every local model. Thus we can define  $\overline{g}_i=ReLU(C(\overline{w}_i,w_i))\frac{L(w_i)}{L(\overline{w}_i)}$ for simplicity. Since $ReLU(\cdot)$ sets $C(\overline{w}_i,w_i)$ as 0 if $C(\overline{w}_i,w_i)\leq 0$, which means the corresponding $\overline{w}_i$ will be discarded by FLTrust during aggregation, so we need to ensure $C(\overline{w}_i,w_i)>0$. In this way, $\overline{g}_i$ can be expressed as $\overline{g}_i=C(\overline{w}_i,w_i)\frac{L(w_i)}{L(\overline{w}_i)}$.
By applying EvilTwin to calculate $\overline{g}_i$, for $i\in \mathcal{M}$, we have
\begin{align}
    \overline{g}_i&=\frac{\sum_{t=1}^T \psi_{i,t} \alpha_{i,t}}{(\sum_{t=1}^T\psi_{i,t})^\frac{1}{2}(\sum_{t=1}^T \psi_{i,t}\alpha_{i,t}^2)^\frac{1}{2}}\frac{(\sum_{t=1}^T\psi_{i,t})^\frac{1}{2}}{(\sum_{t=1}^T \psi_{i,t}\alpha_{i,t}^2)^\frac{1}{2}}\notag\\
    &=\frac{\sum_{t=1}^T \psi_{i,t} \alpha_{i,t}}{\sum_{t=1}^T \psi_{i,t}\alpha_{i,t}^2}.\notag
\end{align}

Since FLTrust is a weight-based method, we can get $\xi_{t,j}$ according to \textbf{Theorem \ref{xi_j}} as below:
\begin{align}
    \xi_{t,j}&\approx abs(g_i-\overline{g}_i\alpha_{i,t})=abs\big(g_i-\frac{\sum_{t=1}^T \psi_{i,t} \alpha_{i,t}}{\sum_{t=1}^T \psi_{i,t}\alpha_{i,t}^2}\alpha_{i,t}\big).\notag
\end{align}

Then, we can express $\xi$ for FLTrust as:
\begin{align}
    \xi&=\sum_{t=1}^T\xi_{t,j}\approx\sum_{t=1}^T\bigg|g_i-\frac{\sum_{t=1}^T \psi_{i,t} \alpha_{i,t}}{\sum_{t=1}^T \psi_{i,t}\alpha_{i,t}^2}\alpha_{i,t}\bigg|\notag\\
    &=\bigg|g_i-\frac{\sum_{t=1}^T \psi_{i,t} \alpha_{i,t} \sum_{t=1}^T\alpha_{i,t}}{\sum_{t=1}^T \psi_{i,t}\alpha_{i,t}^2}\bigg|.\notag
\end{align}
\fi

Then, we can transform  Faker against FLTrust into the following optimization problem: 
\begin{align}
    \textbf{P1}: &\mathop{\arg\max}_{\alpha_{i}}\quad f(\alpha_{i}),\notag \\
    \textit{s.t.}
     \quad~&\textit{C1}: C(\overline{w}_i,w_i)>0, \notag\\
     \quad~&\textit{C2}: ~\forall \alpha_{i,j}>0, \exists \alpha_{i,j}\neq 1,\notag
    \nonumber
\end{align}
where $C1$ ensures the corrupted local model will not be discarded by FLTrust. By solving \textbf{P1}, we have the following theorem. The detailed proofs are in Appendix \ref{the_2}.

\begin{theorem} (Faker Against FLTrust) Setting $\alpha_{i,-j}$ with random positive values,  we can get the approximate optimal value of $\alpha_{i,j}$ by
\begin{align}
    \alpha_{i,j}^*&= \frac{w^2_{i,j}(\beta-\lambda \gamma) + [w^2_{i,j}(\lambda^2+w^2_{i,j}\beta)(w^2_{i,j}\notag \gamma^2+\beta)]^{\frac{1}{2}}}{w^2_{i,j}(\lambda+w_{i,j}^2\gamma)}\notag
,\
\label{faker_fltrust}
\end{align}
where $\lambda=\sum_{-j=1}^{J-1}\alpha_{i,-j}w^2_{i,-j}$, $\beta=\sum_{-j=1}^{J-1}\alpha_{i,-j}^2 w^2_{i,-j}$, and $\gamma=\sum_{-j=1}^{J-1}\alpha_{i,-j}$.
\if()
\begin{align}
    \alpha_{i,j}^*&= \frac{1}{\psi_{i,t}\sum_{-t=1}^{T-1}\alpha_{i,-t}\psi_{i,-t}+\psi_{i,t}^2\sum_{-t=1}^{T-1}\alpha_{i,-t}}\notag\\
    &\bigg(\psi_{i,t}\sum_{-t=1}^{T-1}\alpha_{i,-t}^2\psi_{i,-t}-\psi_{i,t}\sum_{-t=1}^{T-1}\alpha_{i,-t}\psi_{i,-t}\sum_{-t=1}^{T-1}\alpha_{i,-t}\notag\\
    &\pm\bigg[\psi_{i,t}[(\sum_{-t=1}^{T-1}\alpha_{i,-t}\psi_{i,-t})^2+\psi_{i,t}\sum_{-t=1}^{T-1}\alpha_{i,-t}^2\psi_{i,-t}][\psi_{i,t}\notag\\
    &(\sum_{-t=1}^{T-1}\alpha_{i,-t})^2+\sum_{-t=1}^{T-1}\alpha_{i,-t}^2\psi_{i,-t}]\bigg]^{\frac{1}{2}}\bigg).\
\label{faker_fltrust}
\end{align}
\fi
\end{theorem}


\subsubsection{Faker Against Krum}
In Krum, only one local model will be selected as the global model in each round. Thus, to attack Krum, we must ensure that $\overline{w}_i$ will be selected.  The similarity of $\overline{w}_i$ is $\overline{s}_i=E(\overline{w}_{i},w_{i})$, and the objective function is:
\begin{align}
    f(\alpha_{i})=[\sum^J_{j=1}(\alpha_{i,j}w_{i,j}-w_{i,j})^2]^{\frac{1}{2}}\sum_{j=1}^J  \alpha_{i,j}.\notag
\end{align}
To ensure that $\overline{w}_{i}$ can be selected, we need to enable it to have the minimum sum of Euclidean distances from the other models. We can satisfy this requirement by finding the upper bound of the distance of $\overline{w}_i$ between any other benign model. Such a distance can be approximately represented by the distance of $\overline{w}_i$ and $w_i$, i.e., $E(\overline{w}_i, w_i)$. As a single attacker, it only knows the global model $w_g$ besides its trained local model $w_i$ in this round. In this way, we can use the distance between the global model $w_g$ and the local model $w_i$, i.e., $E(w_g,w_i)$, as the approximated upper bound of $E(\overline{w}_i,w_i)$. Thus, we have $E(\overline{w}_i,w_i)<E(w_g,w_i)$.
\if()
Define $\overline{d}_i$ as the Euclidean distance between $\overline{w}_i$ and any other submitted local model (we can use the reference model $w_i$ as an approximation) and $\overline{D}_{i}=\sum_{i=1}^{n-m-1}\overline{d}_i\approx (n-m-1) E(\overline{w}_i,w_i)$ as its sum of distance. 
We have to determine the upper bound of $\overline{D}_i$, i.e., $\overline{D}'_i$, 
by calculating the upper bound of $E(\overline{w}_i, w_i)$. As a single attacker, it only knows the global model $w_g$ besides its trained local model $w_i$ in this round. In this way, we can use the distance between the global model $w_g$ and the local model $w_i$, i.e., $E(w_g,w_i)$, as the approximated upper bound of $E(\overline{w}_i,w_i)$. Thus, we have $E(\overline{w}_i,w_i)<E(w_g,w_i)$.
\fi
\if()
In \cite{fang2020local,shejwalkar2021manipulating}, the poisoned model is supposed to be selected by submitting the same or similar local models to the aggregator. In Faker, instead of submitting identical local models, EvilTwin generates highly similar but completely different models to fool Krum. Specifically, in $\mathcal{M}$, we let the distance between $\overline{w}_i$ and other malicious models be 0 by sending $\overline{w}_i$ to the other malicious clients and 
We define $d_{-i}^b$ as the Euclidean distance between $\overline{W}$ (i.e., $\overline{w}_i$) and benign client's local model and $d_{-i}^p$ as the Euclidean distance between $\overline{W}$ and malicious client's local model. Krum first sorts the distances, then selects the smallest $n-m-1$ distances, and sums them as the total distance, i.e., $D_i=\sum_{i=1}^{m} d_{-i}^p +  \sum_{i=1}^{n-2m-1}d_{-i}^b$. Similarly, for arbitrary local model $w_{-i}$ except $w_i$, it also has a total distance from the others, defined as $D_{-i}$. The basic idea of attacking Krum is to ensure $D_i<D_{-i}$. 
, i.e., $d_{-i}^p=0$. Thus, the total distance can be rewritten as:
\begin{align}
    D_i&=\sum_{-i=1}^{m} d_{-i}^p +  \sum_{-i=1}^{n-2m-1}d_{-i}^b=\sum_{-i=1}^{n-2m-1}d_{-i}^b\notag\\
    &\approx d_{-i}^b (n-2m-1). \notag
\end{align}
Since we assume the local models of benign clients are unknown, we can not get $d_{-i}^b$ directly. We use $d_i=E(w_i,\overline{w}_i)$ to approximately represent $d_{-i}^b$, so we can have $D_i\approx d_i (n-2m-1)$. Similarly, we can get $D_{-i}=d_{-i}^b(n-m-1)\approx \min(d_{i}^b)(n-m-1),$
where $d_{i}^b$ is the distance between $w_i$ and another unpoisoned local model of malicious client. Please note that the application of $\min(d_{i}^b)$ is to ensure that $D_i<D_{-i}$. Thus, we have $d_i (n-2m-1) < \min(d_{i}^b)(n-m-1)$. Since $n$, $m$, and $\min(d_i^b)$ are constants, we can get $d_i < \frac{min(d_{i}^b)(n-m-1)}{n-2m-1}.$
\if()
\begin{align}
   d_i < \frac{min(d_{i}^b)(n-m-1)}{n-2m-1}.\notag
\end{align}
\fi
According to EvilTwin, we know $d_i=E(w_i,\overline{w}_i)=E(w_i, EvilTwin(w_i,w_i,J,S(\cdot,\cdot),T))=[\sum_{t=1}^2 (\alpha_{i,t}^2-2\alpha_{i,t}+1) \psi_{i,t}]^\frac{1}{2}$ and $\theta=\frac{min(d_{i}^b)(n-m-1)}{n-2m-1}$. 
\fi
Then, we formulate Faker against Krum as below:
\begin{align}
    \textbf{P2}: &\mathop{\arg\max}_{\alpha_{i}}\quad f(\alpha_{i})\notag\\
    \textit{s.t.}:
    \quad~&\textit{C1}: E(\overline{w}_i,w_i) < E(w_g,w_i),\notag\\
    \quad~&\textit{C2}:~\forall \alpha_{i,j}>0, \exists \alpha_{i,j}\neq 1,\notag
    \nonumber
\end{align}
where $C1$ ensures that $\overline{w}_i$ can be selected; and $C2$ is the domain of $\alpha_{i,j}$. By solving this optimization problem, we have the following theorem.
\begin{theorem} (Faker Against Krum)
Setting $\alpha_{i,-j}$ with any value that satisfies $0<\alpha_{i,-j}\leq 1+\frac{E(w_g,w_i)}{[(J-1)\max(w_{i,-j})]^\frac{1}{2}}$, then the approximate value of $\alpha_{i,j}$ is calculated by
\if()
\begin{align}
    \alpha_{i,t}>\frac{1}{\psi_{i,t}}\bigg[\psi_{i,t}&- [\psi_{i,t}(\sum_{t=1}^{T-1}((2\alpha_{i,-t}\notag-\alpha_{i,-t}^2\notag\\
    &-1)\psi_{i,-t})+{\theta}^2)]^\frac{1}{2}\bigg],
    \label{low_bound}
\end{align}
or \begin{align}
    \alpha_{i,t}<\frac{1}{\psi_{i,t}}\bigg[\psi_{i,t}&+ [\psi_{i,t}(\sum_{-t=1}^{T-1}((2\alpha_{i,-t}\notag-\alpha_{i,-t}^2\notag\\
    &-1)\psi_{i,-t})+{\theta}^2)]^\frac{1}{2}\bigg].
    \label{upper_bound}
\end{align}
\fi
\begin{align}
0<\alpha_{i,j}<\frac{1}{w_{i,t}}(w_{i,j}+\Omega),\notag
\end{align}
where $\Omega=[w_{i,j}(\sum_{-j=1}^{J-1}((2\alpha_{i,-j}\notag-\alpha_{i,-j}^2-1)w_{i,-j})+{E(w_g,w_i)}^2)]^\frac{1}{2}$.
\end{theorem}
Please see Appendix \ref{the_3} for proofs. Note that the inequality above only provides a lower bound and an upper bound of $\alpha_{i,j}$, and since $f(\alpha_i)$ is a monotonically increasing function when $\alpha_{i,j}>0$, we can take a value slightly smaller than upper bound as the optimal value $\alpha_{i,j}^*$. 


\subsubsection{Faker Against Norm-clipping}
Since norm-clipping adopts $L_2$ norm to measure the reliability of local models, the optimization objective function becomes:
\begin{align}
f(\alpha_{i})=[\sum_{j=1}^J \alpha_{i,j}^2w_{i,j}^2]^\frac{1}{2}\sum_{j=1}^J \alpha_{i,j}.\notag
\end{align}

Norm-clipping requires that the $L_2$ norms of local models are less than the upper bound and treats local models failing the requirement as malicious models for removal. We can use the $L_2$ norm of the local model $w_i$ as the upper bound. In this way, we have $L(\overline{w}_i)\leq L(w_i)$.

\if()
Since $\overline{g}_i$ is not correlated to the similarity metrics during aggregation in norm-clipping, we can treat $\overline{g}_i$ as a constant for simplicity.
Then, we compute $\xi$. Similar to FLTrust, norm-clipping is a weight-based approach. So we can let $\xi_{t,j}\approx|g_i-\overline{g}_i\alpha_{i,t}|$ and $\xi=\sum_{t=1}^T\xi_{t,j}$.  In $\mathcal{M}$, the $L_2$ norm of $\overline{w}_i=EvilTwin(w_i,w_i,J,S(\cdot,\cdot),T)$ is $L(\overline{w}_i)$; to attack norm-clipping, we need to make sure that $L(\overline{w}_i)\leq \theta$. 
\fi
Then, we can formulate Faker against norm-clipping as:
\begin{align*}
    \textbf{P3}: &\mathop{\arg\max}_{\alpha_{i}}\quad f(\alpha_{i}) \notag\\
    \textit{s.t.}:
    \quad~&\textit{C1}:L(\overline{w}_i)\leq L(w_i),\notag\\
    \quad~&\textit{C2}: ~\forall \alpha_{i,j}>0, \exists \alpha_{i,j}\neq 1,
    \nonumber
\end{align*}
where $C1$ ensures that the poisoned local models will not be discarded by norm-clipping and $C2$ is the domain. We can solve \textbf{P3} and summarize it into the following theorem. Please refer to Appendix \ref{the_4} for proofs.
\begin{theorem} (Faker Against Norm-clipping)
Let $\alpha_{i,-j}$ be any number which satisfies  $0<\alpha_{i,-j}\leq [\frac{L(w_i)^2}{(J-1)\max(w_{i,-j})}]^\frac{1}{2}$, then the approximate optimal value of $\alpha_{i,j}$ is calculated by
\begin{align}
\alpha_{i,j}^*=\big(\frac{1}{w_{i,j}}\big[\sum_{-j=1}^{J-1} (-\alpha^2_{i,-j} w_{i,-j})+{L(w_i)}^2\big]\big)^\frac{1}{2}.\nonumber
\end{align}
\end{theorem}
\textbf{Note:} Once the malicious client $i$ calculates all the scalars in $\alpha_i$, it can directly generate the poisoned local model by multiplying scalars and parameters, i.e., $\overline{w}_i=\alpha_i \otimes w_i$, and then submit the poisoned local model to the server. 

\subsection{Complexity Reduction}\label{complexity}
Following our previous analysis, Faker needs to determine the values of the $J$ scalars in $\alpha_i$. Even though we adopt the method that sets values for $J-1$ in a well-defined domain and remains only one variable to reduce the complexity, the computational cost is still significant when $J$ is large with the worst-case time complexity $\mathcal{O}(J)$. To further simplify the computation of $\alpha_i$, we can divide the scalars into $T\ll J$ groups and allow the scalars in the same group to share the same value, thus the time complexity can be reduced to $\mathcal{O}(T)$. 
As for the dividing method, one straightforward way is to treat the parameters in the same layer of deep learning models as in the same group. 

\subsection{Attacking Mode}\label{attack_mode}
In most cases, it is difficult to attack an FL system by controlling multiple local clients at the same time. Unlike most existing attacks, we advocate that a single attacker launches an attack independently without requiring cooperation among multiple clients. In this way, Faker only needs to generate poisoned local models based on the attacker's own local model. In addition, we also provide strategies for Faker in the case of cooperation among attackers. Specifically, attackers are allowed to communicate with each other the local models used to get an intermediate global model $w_g$ to approximate the reference model $w_r$ on the defender. Besides, for attacking Krum, we let attacker $i$ send its obtained poisoned local model $\overline{w}_i$ to the other $m-1$ attackers, who also submit $\overline{w}_i$ to the server; in this way, the upper bound of similarity should be adjusted as $E(\overline{w}_i,w_i) <\frac{n-m-1}{n-2m-1} E(w_g,w_i)$. This paper only provides a preliminary exploration of the attacking mode and more in-depth research is needed in the future.

\if()
\begin{figure}
\centering
\includegraphics[width=0.52\textwidth]{t_faker (3).pdf}
\caption{Dividing $\alpha_i$ into $T$ groups to reduce complexity. The scalars in the same group share the same value.}
\label{ill_t}
\end{figure}
\fi

\if()
\subsection{Basic Faker Attack}\label{basic_faker}

When $T=2$, we call the attack the basic Faker Attack.
\begin{itemize}
    \item Basic Faker Attack Against FLTrust and DiverseFL:
    \begin{align}
        \alpha_{i,t}^*=\pm \alpha_{i,-t}\bigg[\frac{\psi_{i,-t}}{\psi_{i,t}}\bigg]^\frac{1}{2}.
        \notag
    \end{align}
    \item Basic Faker Attack Against Krum:

\begin{align}
    \frac{1}{\psi_{i,t}}(\psi_{i,t}-\Omega)<\alpha_{i,t}<\frac{1}{\psi_{i,t}}(\psi_{i,t}+\Omega)\notag,
\end{align}
where $\Omega=[\psi_{i,t}((2\alpha_{i,-t}\notag-\alpha_{i,-t}^2-1)\psi_{i,-t}+{\theta}^2)]^\frac{1}{2}$.
    
    \item Basic Faker Attack Against Norm-clipping and FLAME:
    \begin{align}
        \alpha_{i,t}^*=\pm \bigg[\frac{1}{\psi_{i,t}}(\theta^2-\alpha_{i,-t}^2\psi_{i,-t})\bigg]^\frac{1}{2}.\notag
    \end{align}
\end{itemize}
\fi

\if()
\subsection{Adjustment of $\theta$}\label{apro_theta}
In Faker, the attacker's local model is utilized as an approximate reference model, i.e., $w_i\approx w_i$, which can introduce an approximation bias in the non-IID scenario.  To this end, we apply a penalty factor $\rho\in(0,1]$ on $\theta$ to ensure that Faker can work effectively in practice. Note that not all attacks need to set $\rho<1$, even with extremely non-IID data. In fact, in the seven benchmarks of AGMs used below, the adjustment of $\theta$ is only required for attacking Krum.  The attacker can evaluate the effectiveness of the attack by checking  the differences between the returned global model and the local one or checking the accuracy during training, enabling them to make  decisions regarding the adjustment of $\theta$.  Overall, the utilization of the local model as an approximation of the reference model in non-IID scenarios is practical.
\fi


\if()
\textbf{Communication Complexity:}
we only consider the communication activities during the attack in each round.
Since there is no data transmission among malicious attackers in  Faker against FLTrust and norm-clipping, we argue that the communication cost is 0. As for Faker against Krum, we require the malicious clients to share their trained local models. Specifically, if client $i$ is decided to send the targeted local model to the aggregator, then the other malicious clients will send their unpoisoned local models to client $i$. After client $i$ generates poisoned local model $\overline{w}_i$, it will send $\overline{w}_i$ to the other malicious clients. Thus, the communication cost of Faker against Krum depends on the number of malicious clients and the data size of the local model. In LA and MB, the communication cost is similar to Faker when attacking FLTrust and norm-clipping. While LA and MB do not require the other malicious clients to send their local models to client $i$, client $i$ needs to send the poisoned local model to the other malicious clients.
\fi


\section{Evaluation of Faker}\label{ev}
In this section, we verify the threat of Faker to FL performance through extensive experiments. We describe the experimental settings in detail and then present results with corresponding discussions. The experiments are conducted using Python 3.10, TensorFlow 2.8, and PyTorch 2.0 running on a desktop with an NVIDIA GeForce RTX 3080 GPU.
\subsection{Experimental Settings}\label{exp_se}
 


\textbf{Datasets.} 
Experimental results presented below are mainly based on MNIST \cite{lecun-mnisthandwrittendigit-2010}, Fashion MNIST (FMNIST)  \cite{xiao2017fashion}, and CIFAR-10 \cite{krizhevsky2009learning}.
MNIST is a handwritten digit database that contains 70000 images for 0 to 9 with the size of 28 $\times$ 28. We split the dataset into training and test data with 60000 and 10000 images, respectively.
 FMNIST is an article image dataset that contains 60000 images as training data and 10000 images as test data. Each sample in FMNIST is a 28 $\times$ 28 grayscale image with a label from 0 to 9.
 CIFAR-10 is a 32 $\times$ 32 color image dataset with 6000 images in each class 
 and 10 classes in total, including 50000 images as training images and 10000 test images.

\textbf{Data Distribution.}
To study the effect of data distribution on adversarial attacks,  
we obtain the IID and Non-IID datasets by the following method: since MNIST, FMNIST, and CIFAR-10 have 10 classes,
we can divide the training data by controlling the number of classes $c\in[1,10]$ that each client can obtain. We consider such a division as IID when $c=10$, and we obtain the non-IID data by setting $c<10$. 

\textbf{Deep Learning Models.} 
We use an MLP model to train MNIST data, which contains a Flatten layer, two Dense layers, and one Dropout layer. As for FMNIST, a CNN model with two Convolution2D layers, one MaxPooling2D layer, two Dropout layers, two Dense layers, and one Flatten layer is designed. Besides, we design a CNN model with three Convolution2D layers, two Maxpooling2D layers, two Dense layers, and one Flatten layer for CIFAR-10.  At the beginning of local training, the parameters are randomly generated. During training, we adopt Adam with the default hyperparameter settings in PyTorch as the optimizer of the deep learning models. To maintain consistency, in our experiments, local models are converted to vectors when evaluated by defenses.

\if()
\begin{table}[H]
\caption{Deep Learning Models.}
\centering
\begin{tabular}{|c|c|l|c|}
\hline
Model & Datasets & Layers & Parameters \\ \hline
MLP   & MNIST    &        &            \\ \hline
CNN & FMNIST   &        &            \\ \hline
CNN & CIFAR-10  &        &            \\ \hline
\end{tabular}
\end{table}
\fi
\textbf{Benchmark Defenses.} 
In experiments, seven aggregation methods, i.e., FedAvg (FA), Krum (KM), norm-clipping (NC), FLTrust (FT), FLAME (FM), DiverseFL (DF), and ShieldFL (SF), are applied to defend against model poisoning attacks; and we use NC, FM, and DF to defend against backdoor attack; and FoolsGod is implemented to detect Sybil attacks. As for FLTrust, DiverseFL, and ShieldFL, the clean data set containing 100 examples is randomly selected from testing data. 
As for norm-clipping, besides its upper bound, we also provide a lower bound, which is four-fifths of the upper bound to prevent the attackers from submitting poisoned local models with very small parameters. FedAvg cannot tolerate adversarial attacks, so we use its results in the case of non-attack (N/A) to compare with the results of other defenses.

\textbf{Benchmark Attacks.} 
We apply LA and MB as the benchmark model poisoning attacks. We follow the basic ideas of LA and MB and design corresponding attacks against the benchmark defenses. We set  $10^{-5}$ as the threshold for both LA and MB. For Faker, we divide the vector of scalars $\alpha_i$ into two parts (i.e., $T=2$) to reduce the time consumption of generating poisoned local models. As for the dividing methods, we allow Faker to choose one layer (e.g., the output layer) as the first group and the rest of the layers as the second group, and the shared scalar for the first group is variable. 

\textbf{Federated Learning Framework.} 
We consider one FL system with 100 clients. In each round, we assume that the server will select all local clients to participate in the training and that each client has sufficient computational, communication, and storage resources to submit a local model in time. 

\textbf{Performance Measurements.} We use the error rate (ER), success rate (SR), and time consumption (TC) of launching the attack to measure the attacking performance of Faker on the model poisoning and Sybil attacks. As for the backdoor attack, we adopt the main task accuracy (MA), targeted task accuracy (TA), and TC as the evaluation metrics. The above measurements will be detailed during the evaluation.

\textbf{Attacking Mode.} For Faker, if not specifically stated, the experiments are conducted with a non-cooperative attack strategy, i.e., the attackers would not be aware of each other. As for other benchmark attacks, attackers are allowed to collude with each other; however, when measuring the time cost it will be the same as Faker, only measuring the time consumption to launch the attack by a single device to ensure a fair comparison.

\subsection{Experimental Results}
We present partial experimental results in this section, and readers can refer to Appendix \ref{ex_faker} for extra results.

\if()
\begin{table}[H]
\centering
\caption{Error rates of IID ($c=10$) vs. Non-IID ($c=2$) data based on MNIST, FMNIST, and CIFAR-10. NC refers to norm-clipping.}
\resizebox{\linewidth}{!}{
\begin{tabular}{|c|c|c|c|c|c|}
\hline
                                                  c                  & Datasets                 & FedAvg                   & Attacks   & Krum   & NC   & FLTrust   \\ \hline
\multirow{12}{*}{10}                                               & \multirow{4}{*}{MNIST}   & \multirow{4}{*}{0.06} & No Attack & 0.14 & 0.06 & 0.07 \\ \cline{4-6} 
                                                                    &                          &                       & LA       & 0.16 & 0.07 & 0.04 \\ \cline{4-7} 
                                                                    &                          &                       & MB        & 0.16 & 0.07 & 0.04 \\ \cline{4-7} 
                                                                    &                          &                       & Faker     & \textbf{0.18} & \textbf{0.13} & \textbf{0.12} \\ \cline{2-6} 
                                                                    & \multirow{4}{*}{FMNIST}  & \multirow{4}{*}{0.17}  & No Attack & 0.22 & 0.17 & 0.15 \\ \cline{4-7} 
                                                                    &                          &                       & LA       & 0.22 & 0.18 & 0.18 \\ \cline{4-7} 
                                                                    &                          &                       & MB        & 0.21 & 0.19 & 0.17 \\ \cline{4-7} 
                                                                    &                          &                       & Faker     & \textbf{0.23} & \textbf{0.23} & \textbf{0.23} \\ \cline{2-6} 
                                                                    & \multirow{4}{*}{CIFAR-10} & \multirow{4}{*}{0.40}  & No Attack & 0.49 &   0.40   & 0.41 \\ \cline{4-7} 
                                                                    &                          &                       & LA       & 0.52 &  0.43    & 0.75 \\ \cline{4-7} 
                                                                    &                          &                       & MA        & 0.51 &   0.41   & 0.74 \\ \cline{4-7} 
                                                                    &                          &                       & Faker     & \textbf{0.64} & \textbf{0.55}      & \textbf{0.77} \\ \hline
\multirow{8}{*}{\begin{tabular}[c]{@{}c@{}}2\end{tabular}} & \multirow{4}{*}{MNIST}   & \multirow{4}{*}{0.17}  & No Attack & 0.89 & 0.17 & 0.12 \\ \cline{4-6} 
                                                                    &                          &                       & LA       & 0.91 & 0.33 & 0.30 \\ \cline{4-7} 
                                                                    &                          &                       & MB        & 0.91 & 0.19 & 0.16 \\ \cline{4-7} 
                                                                    &                          &                       & Faker     & 0.91 & \textbf{0.80} & \textbf{0.34} \\ \cline{2-6} 
                                                                    & \multirow{4}{*}{FMNIST}  & \multirow{4}{*}{0.20}  & No Attack & 0.81 & 0.20 & 0.18 \\ \cline{4-7} 
                                                                    &                          &                       & LA       & 0.85 & 0.20 & 0.72 \\ \cline{4-7} 
                                                                    &                          &                       & MB        & 0.86 & 0.20 & 0.80 \\ \cline{4-7} 
                                                                    &                          &                       & Faker     & \textbf{0.90} & \textbf{0.23} & \textbf{0.82} \\ \hline
\end{tabular}}
\label{noniid}
\end{table}
\fi

\begin{table}[ht]
\centering
\caption{Error rates of IID data ($c=10$) with $n=100$ and $m=20$. When there is no attack, the error rates of FA for MNIST, FMNIST, and CIFAR-10 are 0.06, 0.17, and 0.40. }
\arrayrulecolor{black}
\resizebox{\linewidth}{!}{
\begin{tabular}{cccccccc} 
\hline
\rowcolor[HTML]{FFCCC9} 
Dataset                 & Attack & KM & NC & FT & FM & DF & SF\\
\hline
\hline
\multirow{4}{*}{MNIST}   
& N/A    & 0.14   &  0.06  & 0.07   &  0.07  &  0.06  &  0.07\\
& LA    &  0.16  &  0.07  &  0.04  &  0.06  & 0.06   &  0.06  \\
& MB      & 0.16   &  0.07  &   0.04 &  0.06  & 0.06   &  0.06 \\
& Faker   &  \textbf{0.18}  &  \textbf{0.17}  &  \textbf{0.12}  & \textbf{0.22}   & \textbf{0.56}  &  \textbf{0.90}  \\ 
\arrayrulecolor{black}\cline{1-1}\arrayrulecolor{black}\cline{2-8}
\multirow{4}{*}{FMNIST}  
& N/A    & 0.22   & 0.17   & 0.15   &   0.19 &  0.18  & 0.15  \\
& LA    & 0.22   & 0.18   & 0.18   & 0.18   & 0.20   & 0.17  \\
& MB      & 0.21   & 0.19   & 0.17   &  0.16  & 0.20   &  0.17 \\
& Faker   & \textbf{0.23}   & \textbf{0.23}   & \textbf{0.23}   &  \textbf{0.77}  &  \textbf{0.40}  &   \textbf{0.90}  \\ 
\arrayrulecolor{black}\cline{1-1}\arrayrulecolor{black}\cline{2-8}
\multirow{4}{*}{CIFAR-10} & N/A    &  0.49  &  0.40  &  0.41  & 0.46   &  0.48  &  0.48 \\
                         & LA    &  0.52  & 0.43   &  0.75  &  0.46  &  0.48  & 0.49   \\
                         & MB      &  0.51  & 0.41   &  0.74  &  0.46  &  0.50  & 0.49  \\
                         & Faker   &  \textbf{0.64} &  \textbf{0.55}  &  \textbf{0.77}  & \textbf{0.48}    & \textbf{0.65}   &  \textbf{0.68} \\
\arrayrulecolor{black}\cline{1-1}\arrayrulecolor{black}\cline{2-8}
\end{tabular}}
\arrayrulecolor{black}
\label{iid}
\end{table}


\begin{table}[ht]
\centering
\caption{Error rates of non-IID  data ($c=2$) with $n=100$ and $m=20$. The error rates of FA for the two datasets are 0.17 and 0.20 when there is no attack.}
\arrayrulecolor{black}
\resizebox{\linewidth}{!}{
\begin{tabular}{cccccccc} 
\hline
\rowcolor[HTML]{FFCCC9} 
Dataset                 & Attack & KM & NC & FT & FM & DF & SF \\
\hline
\hline
\multirow{4}{*}{MNIST}   & N/A    &  0.89  &  0.17  &  0.12  &  0.15  &  0.20  &  0.16  \\
                         & LA    &  0.90  & 0.33   &  0.30  & 0.15   &  0.20  &  0.90 \\
                         & MB      &  0.90  & 0.19   &  0.16  &  0.15  & 0.17   &  0.90\\
                         & Faker   &   0.90 &  \textbf{0.80}  &  \textbf{0.34}  &  \textbf{0.50}  & \textbf{0.78}   &  0.90 \\
\arrayrulecolor{black}\cline{1-1}\arrayrulecolor{black}\cline{2-8}
\multirow{4}{*}{FMNIST}  & N/A    &  0.81  & 0.20   & 0.18   & 0.19  & 0.20   &  0.90 \\
                         & LA    & 0.85   & 0.20   &  0.72  &  0.19  &   0.20 &  0.90 \\
                         & MB      & 0.86   & 0.20   &  0.80  & 0.19   & 0.20   & 0.90   \\ 
                         & Faker   & \textbf{0.90}   & \textbf{0.73}  & \textbf{0.82}   &  \textbf{0.90}  &  \textbf{0.48}  & 0.90 \\ 
\arrayrulecolor{black}\cline{1-1}\arrayrulecolor{black}\cline{2-8}

\end{tabular}}
\arrayrulecolor{black}
\label{noniids}
\end{table}

\textbf{Impacts of Data Distribution on Faker.} By setting $c=10$ and $c=2$ to get IID and non-IID data,  we evaluate Faker with different data distributions. The evaluation of IID data is based on MNIST, FMNIST, and CIFAR-10, and the evaluation of non-IID is based on MNIST and FMNIST. The results on CIFAR-10 are not presented when $c=2$ since they can not even converge when there is no attack. 
 We set $m=20$ and use test error rates to measure the performance of attacks. The results are shown in Table \ref{iid} and Table \ref{noniids}. Overall, we know that Faker outperforms LA and MB in both IID and non-IID situations among all the datasets and defenses. Specifically, for MNIST,  when $c=10$, both LA and MB do not degrade the performances of FLTrust, FLAME, DiverseFL, and ShieldFL, and they only decrease a little when attacking  Krum and norm-clipping, but Faker can undermine all these defenses; for FMNIST and CIFAR-10, the impacts of Faker on these defenses are more significant. When $c=2$, both LA and MB degrade the accuracy of the defenses except FLAME, and Faker greatly undermines all the defenses. 
Compared with LA and MB, Faker is more powerful in attacking and more compatible with IID and non-IID data.

\begin{table}[H]
\caption{Error rates of different numbers of malicious clients when $c=5$ and $n=100$. The error rates of FA for the three datasets are 0.09, 0.17, and 0.52 when no attack. }
\centering
\arrayrulecolor{black}
\resizebox{\linewidth}{!}{
\begin{tabular}{ccccccccc}
\arrayrulecolor{black}\hline
\rowcolor[HTML]{FFCCC9} 
$m/n$                  & Dataset               & Attack & KM & NC & FT & FM & DF & SF \\ 
\hline
\hline
\multirow{2}{*}{0\%}  & MNIST  & N/A    &  0.48  & 0.11   &0.09    &  0.10  & 0.09   &  0.09 \\
                        & FMNIST                        & N/A   &  0.40  & 0.19   & 0.17   &  0.18  & 0.19   & 0.54  \\
                        & CIFAR-10                        & N/A   &  0.64  & 0.52   & 0.55   &0.58    & 0.52  & 0.56  \\
\arrayrulecolor{black}\cline{1-2}\arrayrulecolor{black}\cline{3-9}
\multirow{8}{*}{1\%}  & \multirow{4}{*}{MNIST}  
                                                 & LA    & 0.50   & 0.11   &  0.06  &  0.10  &  0.10  &  0.10 \\
                      &                         & MB      &  0.49  & 0.10 & 0.06   &   0.10  & 0.06 &   0.10 \\
                      &                         & Faker   & \textbf{0.54}   & \textbf{0.12}   & \textbf{0.10}   &  0.10  & \textbf{0.13}   &\textbf{0.90}   \\ 
\arrayrulecolor{black}\cline{2-2}\arrayrulecolor{black}\cline{3-9}
                      & \multirow{4}{*}{FMNIST}  
                      & LA    &  0.45  & 0.19   &  0.17  &  0.18  & 0.19  &  0.56 \\
                      &                         & MB      & 0.36   & 0.19  & 0.16   &0.18    & 0.18   &  0.55\\
                      &                         & Faker   &  \textbf{0.50}  & 0.19  & \textbf{0.19}   & \textbf{0.20}   & \textbf{0.23}   &   \textbf{0.90}  \\ 
\arrayrulecolor{black}\cline{2-2}\arrayrulecolor{black}\cline{3-9}
                      & \multirow{4}{*}{CIFAR-10}  
                      & LA    &  0.65  & 0.55   &  0.57  & 0.58   & 0.55  & 0.90 \\
                      &                         & MB      &  0.62  & 0.55  & 0.55   & 0.59   & 0.55  &0.61  \\
                      &                         & Faker   &  \textbf{0.66}  &  \textbf{0.60} & \textbf{0.58}    &  \textbf{0.63}  & \textbf{0.61}   &   0.90 \\ 

\arrayrulecolor{black}\cline{1-2}\arrayrulecolor{black}\cline{3-9}

\multirow{8}{*}{5\%}  & \multirow{4}{*}{MNIST}  
                                                & LA    &  0.54  & 0.11   & 0.06   & 0.10   & 0.10   & 0.24  \\
                      &                         & MB      &  0.52  & 0.10   &  0.06  & 0.10   & 0.10   &  0.16 \\
                      &                         & Faker   & \textbf{0.60}   & \textbf{0.14}   & \textbf{0.41}   &  \textbf{0.11} &  \textbf{0.36}  &  \textbf{0.90} \\ 
\arrayrulecolor{black}\cline{2-2}\arrayrulecolor{black}\cline{3-9}
                      & \multirow{4}{*}{FMNIST}  
                      & LA    & 0.34   & 0.20   & 0.18   & 0.18   &  0.20  &  0.57   \\
                      &                         & MB      & 0.46   & 0.19   & 0.17   & 0.18   & 0.18   &  0.59  \\
                      &                         & Faker   &  \textbf{0.53}  & 0.20   & \textbf{0.21}   & \textbf{0.24}   &  \textbf{0.37}  &  \textbf{0.90} \\ 
\arrayrulecolor{black}\cline{2-2}\arrayrulecolor{black}\cline{3-9}
                      & \multirow{4}{*}{CIFAR-10}  
                      & LA    & 0.65   & 0.55   &  0.65  &  0.76  & 0.55  & 0.90 \\
                      &                         & MB      & 0.62   & 0.52  & 0.64   &   0.78 & 0.55  & 0.77  \\
                      &                         & Faker   &  \textbf{0.68}  & \textbf{0.80} & \textbf{0.67}   & \textbf{0.83}    & \textbf{0.79}   &   0.90  \\ 

\arrayrulecolor{black}\cline{1-2}\arrayrulecolor{black}\cline{3-9}

\multirow{8}{*}{10\%} & \multirow{4}{*}{MNIST} 
                                                 & LA    &  0.60  & 0.10   & 0.09   & 0.10   & 0.10   & 0.78  \\
                      &                         & MB      &  0.57  & 0.11   & 0.08   & 0.10   & 0.10   &  0.67   \\
                      &                         & Faker   &  \textbf{0.64}  & \textbf{0.16}   & \textbf{0.90}   &  \textbf{0.20}  &  \textbf{0.59}  &   \textbf{0.90} \\ 
\arrayrulecolor{black}\cline{2-2}\arrayrulecolor{black}\cline{3-9}
                      & \multirow{4}{*}{FMNIST} 
                      & LA    &  0.50  & 0.11   & 0.23   &  0.18  & 0.21   &  0.85 \\
                      &                         & MB      &  0.48  & 0.11   & 0.11   &  0.18  & 0.19   &  0.79   \\
                      &                         & Faker   &  \textbf{0.61}  & \textbf{0.24}   & \textbf{0.55}   & \textbf{0.30}  &   \textbf{0.55}  &    \textbf{0.90}  \\ 
\arrayrulecolor{black}\cline{2-2}\arrayrulecolor{black}\cline{3-9}
                      & \multirow{4}{*}{CIFAR-10}  
                      & LA    &  0.66  & 0.55   &  0.77  & 0.82   & 0.55  & 0.90 \\
                      &                         & MB      &  0.63  &  0.53 &  0.76  &  0.80  &  0.55 & 0.90  \\
                      &                         & Faker   &  \textbf{0.70}  & \textbf{0.90}  &  \textbf{0.78}  &  \textbf{0.90}  & \textbf{0.90}   &   0.90 \\ 

\arrayrulecolor{black}\cline{1-2}\arrayrulecolor{black}\cline{3-9}
\multirow{8}{*}{20\%} & \multirow{4}{*}{MNIST}  
                                                & LA    & 0.61   & 0.11   & 0.08    & 0.10   & 0.10   &  0.90 \\
                      &                         & MB      &  0.60  &  0.11  &  0.06 & 0.10   &  0.10  &  0.90 \\
                      &                         & Faker   & \textbf{0.66}   & \textbf{0.18}   & \textbf{0.90}   & \textbf{0.43}    & \textbf{0.66}   & 0.90   \\ 
\arrayrulecolor{black}\cline{2-2}\arrayrulecolor{black}\cline{3-9}
                      & \multirow{4}{*}{FMNIST} 
                      & LA    &  0.90  & 0.21   & 0.30   & 0.32   &   0.21 &  0.90   \\
                      &                         & MB      &  0.59  &  0.21  & 0.25   & 0.30   & 0.18   &  0.90   \\
                      &                         & Faker   & 0.90   & \textbf{0.34}   & \textbf{0.60}   & \textbf{0.90}   & \textbf{0.69}   & 0.90   \\ 
\arrayrulecolor{black}\cline{2-2}\arrayrulecolor{black}\cline{3-9}
                      & \multirow{4}{*}{CIFAR-10}  
                      & LA    &   0.68 & 0.55   & 0.83   &  0.90  & 0.55  & 0.90 \\
                      &                         & MB      & 0.64   &  0.54 & 0.82   & 0.90   &  0.55 & 0.90\\
                      &                         & Faker   & \textbf{0.76}   & \textbf{0.90}  & \textbf{0.85}   & 0.90   & \textbf{0.90}   &  0.90   \\ 
\arrayrulecolor{black}\cline{1-2}\arrayrulecolor{black}\cline{3-9}
\multirow{8}{*}{50\%} & \multirow{4}{*}{MNIST} 
                                                & LA    &  0.90  & 0.11   & 0.23   &  0.40  &  0.10  &   0.90   \\
                      &                         & MB      &  0.59  & 0.11   & 0.11   & 0.54   &  0.10  &  0.90  \\
                      &                         & Faker   & 0.90   & \textbf{0.34}   & \textbf{0.90}   &  \textbf{0.60}  & \textbf{0.81}    & 0.90   \\ 
\arrayrulecolor{black}\cline{2-2}\arrayrulecolor{black}\cline{3-9}
                      & \multirow{4}{*}{FMNIST} 
                                                & LA    &  0.90  &  0.21  & 0.65   &0.60    &  0.21  &  0.90   \\
                      &                         & MB      &  0.90  &  0.21  &  0.63  & 0.70   & 0.19   &  0.90  \\
                      &                         & Faker   &   0.90 &   \textbf{0.39}  &  \textbf{0.72} & \textbf{0.90}   & \textbf{0.90}   &  0.90    \\
\arrayrulecolor{black}\cline{2-2}\arrayrulecolor{black}\cline{3-9}
                      & \multirow{4}{*}{CIFAR-10}  
                      & LA    & 0.90   &  0.55  & 0.90   & 0.90   & 0.55  &0.90   \\
                      &                         & MB      &  0.90  &  0.57 & 0.90   &  0.90  & 0.55  & 0.90   \\
                      &                         & Faker   &  0.90  & \textbf{0.90}  & 0.90   &  0.90  & \textbf{0.90}   &   0.90 \\ 
\arrayrulecolor{black}\cline{1-2}\arrayrulecolor{black}\cline{3-9}
\end{tabular}}
\arrayrulecolor{black}
\label{m_faker}
\end{table}

\textbf{Impacts of Malicious Clients' Number on Faker.} 
Intuitively, with more malicious clients, the impacts of the model poisoning attacks will be more significant. To support this idea, we set $n=100$ and vary the number of malicious clients $m$; and $m/n$ represents the proportion of malicious clients to all the clients. To simulate the practical application scenario of FL, we set $c=5$ to get the non-IID data.
The experimental results are presented in Table \ref{m_faker}. From the results, Faker successfully achieves the reduction in model performance regardless of the variation in $m/n$, datasets, and defenses and outperforms both LA and MB. Even when there is only one malicious client, Faker can still undermine similarity-based defenses, while the other two model poisoning attacks have poor performance. It is worth noting that LA and MB have almost no effect on norm-clipping because LA and MB can only meet the requirements of norm-clipping by maintaining or reducing the magnitude of the local model. With the upper and lower bounds we have set for norm-clipping, it is more difficult for LA and MB to generate the required poisoned models to attack norm-clipping. Since Faker only needs to generate poisoned models that meet the requirement of the upper bound according to the formula when attacking norm-clipping, the attack is not affected by the lower bound. From the results of the experiments, we can argue that the more malicious clients we have, the more successful the Faker attack will be. Furthermore, Faker can effectively perform well with few malicious clients so as to attack industrial FL \cite{shejwalkar2022back}.

\textbf{Single Round Attack.}  
In the previous experiments, we assume that the attackers launch attacks in each round, but in reality, smart attackers often launch attacks when the global model is converging in order to hide themselves and reduce the model performance at the same time. In our experiments, we test Faker's performance in launching a single attack in one round when the local loss is small. The results in Table \ref{one_round} indicate that Faker outperforms the benchmark attacks and can still undermine the global model by attacking only once.

\begin{table}[H]
\centering
\caption{Error rates  when only attack once  with $c=5$, $n=100$, and $m=20$. The error rates of FA for the three datasets are 0.09, 0.17, and 0.52 when no attack.}
\arrayrulecolor{black}
\resizebox{\linewidth}{!}{
\begin{tabular}{cccccccc} 
\hline
\rowcolor[HTML]{FFCCC9} 
Dataset                 & Attack & KM & NC & FT & FM & DF & SF\\
\hline
\hline
\multirow{4}{*}{MNIST}   
& N/A    &  0.48  & 0.11   &0.09    &  0.10  & 0.09   &  0.09  \\
& LA    &  0.48 &  0.11  &  0.09 & 0.10   & 0.12   &   0.12 \\
& MB      & 0.48   &0.11    & 0.09  &  0.10 &  0.11  &  0.10 \\
& Faker   &  \textbf{0.52}  &  \textbf{0.12}  &  \textbf{0.10}  & \textbf{0.12}   & \textbf{0.14}  &  \textbf{0.78}  \\ 
\arrayrulecolor{black}\cline{1-1}\arrayrulecolor{black}\cline{2-8}
\multirow{4}{*}{FMNIST}  
& N/A     &  0.40  & 0.19   & 0.17   &  0.18  & 0.19   & 0.54    \\
& LA    & 0.40  & 0.21  &  0.17  &  0.20  & 0.19  & 0.90\\
& MB      & 0.40   & 0.19  &  0.17 & 0.18   &  0.19  & 0.90  \\
& Faker   & \textbf{0.54}   & \textbf{0.26}   & \textbf{0.20}   &  \textbf{0.34}  &  \textbf{0.30}  &   0.90  \\ 
\arrayrulecolor{black}\cline{1-1}\arrayrulecolor{black}\cline{2-8}
\multirow{4}{*}{CIFAR-10} & N/A    &  0.64  & 0.52   & 0.55   &0.58    & 0.52  & 0.56  \\
                         & LA    &  0.64 & 0.52   & 0.55  &  0.60  &  0.52  & 0.90  \\
                         & MB      &  0.64  & 0.52  &   0.55 & 0.59   &  0.52 & 0.90  \\
                         & Faker   &  \textbf{0.68} &  \textbf{0.64}  &  \textbf{0.57}  & \textbf{0.80}    & \textbf{0.56}   &  0.90 \\
\arrayrulecolor{black}\cline{1-1}\arrayrulecolor{black}\cline{2-8}
\end{tabular}}
\arrayrulecolor{black}
\label{one_round}
\end{table}

\textbf{Success Rates of Faker.} 
We first define success rates as the percentage of effective attack rounds to total training rounds. For each defense, the measurements of a successful attack will be different. Since FLTrust and FLAME use similarity evaluation results to filter out outliers, we define that the attacks toward them are successful if they accept all the poisoned submissions. As for Krum, the attack is successful if one of the poisoned local models is selected as the global model. In our experiments, norm-clipping and DiverseFL both have lower and upper bounds, so we define the attack as successful if no discarded poisoned submissions. As for ShieldFL, we define the attack as successful if the poisoned model is assigned a higher weight during aggregation. Our definition of a successful attack is strict in that if any poisoned model is rejected in any round, the attack is not considered successful in that round. We set $n=100$, $m=20$, and $c=5$, then train 100 rounds. From the experimental results presented in Table \ref{success}, we can see that Faker can always get a 100\% success rate with different datasets and defenses. However, LA and MB can not always get high success rates, especially when attacking Krum and DiverseFL, because they generate poisoned local models in an iterative method which can not guarantee all the poisoned submissions can meet the requirements of defenses. Moreover, in Faker, we use rigorous mathematical analysis to derive the optimal attack strategies, ensuring defenses can accept all the poisoned local models.

\begin{table}
\centering
\caption{Evaluation of success rate of attack with $n=100$, $m=20$, and $c=5$ by running 100 rounds.}
\resizebox{\linewidth}{!}{
\begin{tabular}{c|ccc|ccc|ccc} 
\hline
\rowcolor[HTML]{FFCCC9} 
              & \multicolumn{3}{c|}{MNIST} & \multicolumn{3}{c|}{FMNIST} & \multicolumn{3}{c}{CIFAR-10} \\ 
\hline
\rowcolor[HTML]{C0C0C0} 
Defenses          & LA & MB & Faker          & LA & MB & Faker  & LA & MB & Faker         \\ 
\hline
\hline
KM          & 0.04     &  0.02  &      \textbf{1.00}          &  0.00    & 0.00   & \textbf{1.00}    & 0.09     & 0.05   & \textbf{1.00}            \\
NC &     0.94 &   0.89 &         \textbf{1.00}       &   0.87   &  0.85  & \textbf{1.00}   &   0.73   & 0.66   & \textbf{1.00}             \\
FT      &     1.00 &  1.00  &          1.00      &  1.00    &  1.00  &   1.00     &  1.00    & 1.00   & 1.00        \\
FM         &  0.00    & 0.00   &   \textbf{1.00}             &    0.45 & 0.37   &    \textbf{1.00}   & 1.00     & 1.00   & 1.00          \\
DF     &  0.00    & 0.00   &   \textbf{1.00}             &   0.06   &  0.00  &    \textbf{1.00}   &   0.00   & 0.00   & \textbf{1.00}          \\
SF      &  1.00    & 1.00   &    1.00            & 1.00     & 1.00   & 1.00     &  1.00    & 1.00   & 1.00        \\
\hline
\end{tabular}}
\label{success}
\end{table}

\begin{table}
\centering
\caption{Evaluation of time consumption with $n=100$, $m=20$, and $c=5$. The time cost is measured in seconds.}
\resizebox{\linewidth}{!}{
\begin{tabular}{c|ccc|ccc|ccc} 
\hline
\rowcolor[HTML]{FFCCC9} 
              & \multicolumn{3}{c|}{MNIST} & \multicolumn{3}{c|}{FMNIST} & \multicolumn{3}{c}{CIFAR-10} \\ 
\hline
\rowcolor[HTML]{C0C0C0} 
Defenses         & LA & MB & Faker          & LA & MB & Faker     & LA & MB & Faker       \\ 
\hline
\hline
KM          & 5.812     &  5.064  &      \textbf{0.827}          &  40.632    & 39.115   & \textbf{6.234}   &  7.254   &  7.143  & \textbf{4.242}             \\
NC &     0.026 &   0.022 &        \textbf{0.015}       &   0.142   &  0.122  & \textbf{0.083}    &   0.682   &0.136   & \textbf{0.093}             \\
FT       &     0.089 &  0.070  &          \textbf{0.015}      &  0.129    &  0.124  &   \textbf{0.016}   &  0.069    &  0.067  & \textbf{0.030}          \\
FM         & 0.231     & 0.211   &    \textbf{0.183}            & 0.443     &  0.486  &      \textbf{0.336}   &  0.568    &  0.435  & \textbf{0.245}         \\
DF     &  0.258    &  0.374  &  \textbf{0.214}               & 0.474    &  0.264  &  \textbf{0.173}   &   1.322   & 0.353   & \textbf{0.225}             \\
SF      &   0.231   & 0.293   &  \textbf{0.152}              &  0.835    & 0.459  &        \textbf{0.235}  &  0.335    & 0.362   & \textbf{0.242}        \\
\hline
\end{tabular}}
\label{time}
\end{table}

\textbf{Time Consumption of Faker.} Then, we explore the time cost of launching Faker by a single attacker, which is measured in seconds. By setting $n=100$, $m=20$, and $c=5$, we can directly get the running time of launching different attacks. According to results in Table \ref{time}, Faker is much more time efficient in launching attacks against similarity-based defenses with three datasets. Since our tests are conducted on a device with a GPU, all experimental values are within one minute, but for mobile devices that do not have high computing power, it would be time-consuming and impractical to launch LA and MB attacks when Krum is applied. Faker takes less time than the other attacks, even if attacking Krum.

\begin{figure}
    \centering
    \subfigure[MNIST.]{\includegraphics[width=0.15\textwidth]{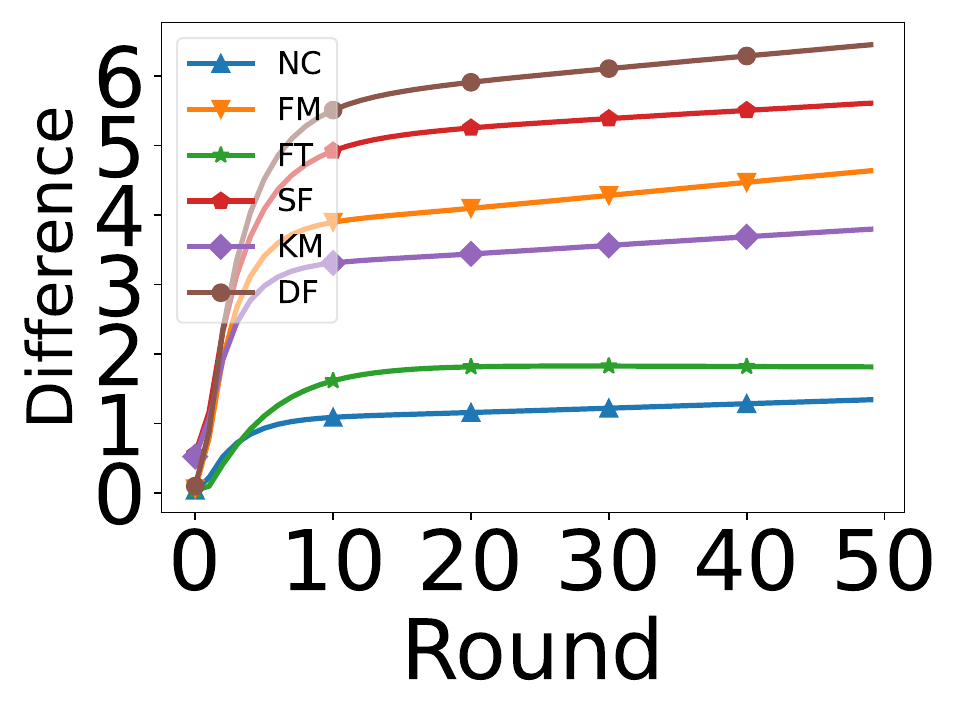}} 
    \subfigure[FMNIST.]{\includegraphics[width=0.15\textwidth]{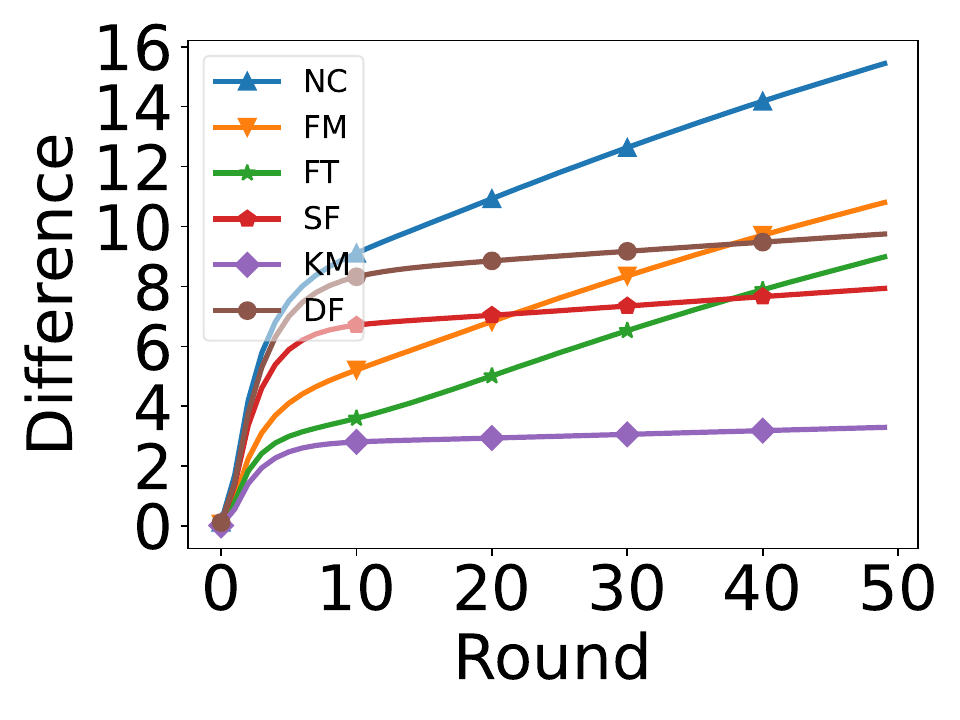}} 
    \subfigure[CIFAR-10.]{\includegraphics[width=0.15\textwidth]{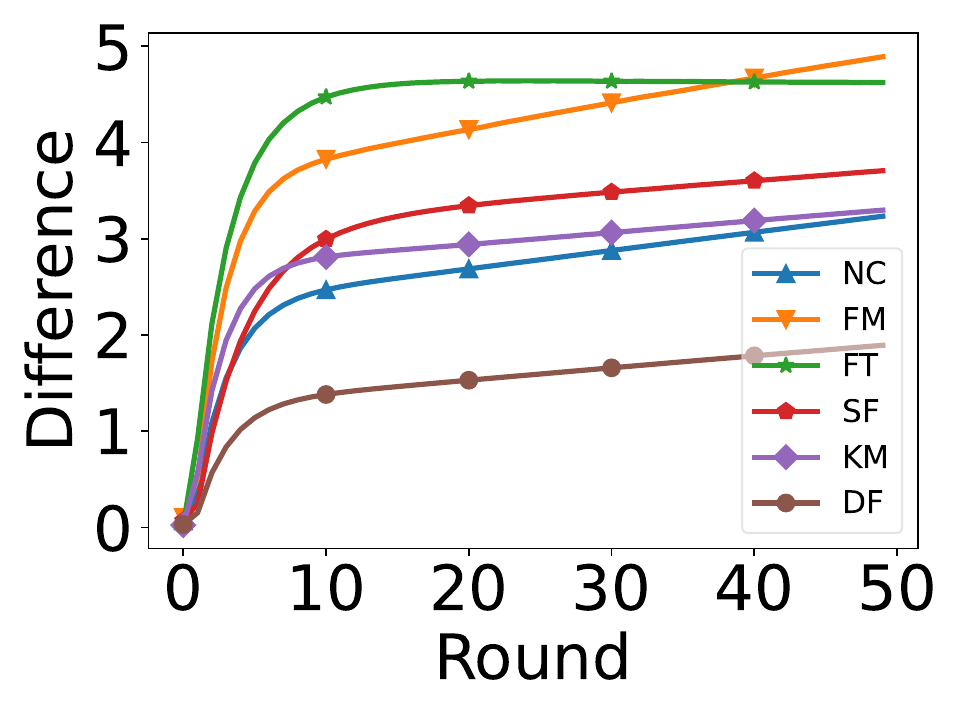}}
\caption{Evaluation of global model's difference with $n=100$, $m=20$, and $c=5$.}
\label{fig_xi}
\end{figure}

\textbf{Difference Between Benign and Poisoned Global Models.} 
We randomly select two parameters from both poisoned and unpoisoned global models in each round to measure the difference caused by Faker. We set $n=100$, $m=20$, and $c=5$, and the experimental results are presented in Fig. \ref{fig_xi}. From the results, it can be seen that the difference due to the poisoned model generated by Faker is greater than 1 after multiple rounds. It is important to note that the parameters of the FL model are usually numbers with absolute values much smaller than 1. Therefore, we can conclude that the poisoned model is sufficient to cause the performance degradation of the global model.

\textbf{Evaluation of Attacking Mode.} We also evaluate Faker's performance in different attack modes. The experimental results are shown in Table \ref{attacking_m}, which demonstrate that the cooperation mode outperforms the non-cooperation mode. However, the attack in non-cooperative mode is sufficient in terms of its effectiveness.
\begin{table}[H]
\centering
\caption{Error rates of Faker in different attacking modes with $c=5$, $n=100$, and $m=20$, where ``single'' and ``cooperation'' refer to non-cooperation and cooperation modes, respectively.}
\arrayrulecolor{black}
\resizebox{\linewidth}{!}{
\begin{tabular}{cccccccc} 
\hline
\rowcolor[HTML]{FFCCC9} 
Dataset                 & Mode & KM & NC & FT & FM & DF & SF\\
\hline
\hline
\multirow{2}{*}{MNIST}   
& cooperation      & \textbf{0.69}   & \textbf{0.24}    & 0.90  &  \textbf{0.26} &  \textbf{0.65}  &  0.90 \\
& single   &  0.64  &  0.16  &  0.90  &0.20  & 0.59  &  0.90  \\ 
\arrayrulecolor{black}\cline{1-1}\arrayrulecolor{black}\cline{2-8}
\multirow{2}{*}{FMNIST}  
& cooperation      & 0.90   & \textbf{0.37}  &  \textbf{0.64} & 0.90  &  \textbf{0.72}  & 0.90  \\
& single   & 0.90   & 0.34  & 0.60   &  0.90  &  0.69  &   0.90  \\ 
\arrayrulecolor{black}\cline{1-1}\arrayrulecolor{black}\cline{2-8}
\multirow{2}{*}{CIFAR-10} 
& cooperation      &  \textbf{0.79}  & 0.90  &   \textbf{0.89} & 0.90   &  0.90 & 0.90  \\
& single   &  0.76 &  0.90  &  0.85  & 0.90    & 0.90   &  0.90 \\
\arrayrulecolor{black}\cline{1-1}\arrayrulecolor{black}\cline{2-8}
\end{tabular}}
\arrayrulecolor{black}
\label{attacking_m}
\end{table}

\textbf{Extending Faker to Other Attacks.}
Even though we focus on the untargeted model poisoning attacks in this work, we still provide the evaluation of Faker-based targeted backdoor attacks and Sybil attacks, and their detailed designs are shown in Appendix \ref{faker_other_attacks}. From Table \ref{backdoor_faker}, we can see that the backdoor attack based on Faker can better maintain the main task accuracy (MA) and decrease the targeted task accuracy (TA) with less time consumption compared to the benchmark attack. The results in Table \ref{sybil_faker} show that the Faker-based Sybil attack can bypass the detection of FoolsGold \cite{fung2018mitigating} and has better performance in decreasing the accuracy, success rate, and time cost than the benchmark adopted in \cite{fung2018mitigating}. The above experimental results indicate that Faker can be well extended to other adversarial attacks, further illustrating the threats of similarity metrics' vulnerabilities.

\begin{table}[]
\centering
\caption{Evaluation of Faker-based backdoor attack. The performance is measured by main task accuracy (MA), targeted task accuracy (TA), and time cost (TC) of attack. NC w. Faker means using Faker to attack NC. ``-'' implies not applicable. We set $n=100$, $m=20$, and $c=5$.}
\resizebox{\linewidth}{!}{
\begin{tabular}{c|ccc|ccc|ccc}
\hline
\rowcolor[HTML]{FFCCC9} 
     & \multicolumn{3}{c|}{\cellcolor[HTML]{FFCCC9}MNIST} & \multicolumn{3}{c|}{\cellcolor[HTML]{FFCCC9}FMNIST} & \multicolumn{3}{c}{\cellcolor[HTML]{FFCCC9}CIFAR-10} \\ \hline
\rowcolor[HTML]{C0C0C0} 
Defenses & MA                       & TA        &TC              & MA                       & TA          &TC             & MA                        & TA     &TC                  \\ \hline
NC w. N/A  &  0.97                        &  0.99      &    -             &      0.82                    &   0.64          &      -      &      0.57                     &    0.45         &       -     \\
NC w. C\&S  &  0.86                        &    0.61        &   0.03         &     0.70                     &      0.45     &      0.09       &  0.43                        &               0.25     & 0.13 \\   
NC w. Faker  &  \textbf{0.93}                        &    \textbf{0.52}     &    \textbf{0.01}             &     \textbf{0.73}                     &      \textbf{0.32}      & \textbf{0.08}           &  \textbf{0.46}                         &               \textbf{0.23}     &   \textbf{0.09}   \\ \hline
FM w. N/A  &   0.97                       &    0.99         &     -     &     0.82                     &       0.67          &   -      &         0.57                 & 0.44         &         -       \\
FM w. C\&S   &    0.92                       &      0.50    &     0.03           &        0.72                   &        0.26       &    0.09        &          0. 49                 &     0.28            &    0.18      \\
FM w. Faker   &    \textbf{0.94}                       &      \textbf{0.48}           &   \textbf{0.02}      &        \textbf{0.74}                   &        \textbf{0.23}         &   \textbf{0.07}       &          \textbf{0. 51}                 &     \textbf{0.26}       &    \textbf{0.11}          \\\hline
DF w. N/A &   0.97                       &     0.99          &    -      &       0.82                   &     0.65          &      -     &     0.23                     &     0.21               & -     \\
DF w. C\&S   &    0.92                       &      0.63         &    0.25      &       0.68                    &       0.35          &     0.32    &   0.47                      &   0.23          &      0.43       \\
DF w. Faker   &    \textbf{0.95}                       &      \textbf{0.56}      &     \textbf{0.22}        &       \textbf{0.71}                    &       \textbf{0.32}          &     \textbf{0.24}      &   \textbf{0.49}                         &   \textbf{0.11}            &      \textbf{0.32}     \\ \hline
\end{tabular}}
\label{backdoor_faker}
\end{table}

\begin{table}[]
\centering
\caption{Evaluation of Faker-based Sybil attack. The performance is measured by error rate (ER), success rate (SR), and time cost (TC) of attack. FoolsGold w. Faker means using Faker to attack FoolsGold.  ``-'' implies not applicable. We set $n=100$, $m=20$, and $c=5$.}
\resizebox{\linewidth}{!}{
\begin{tabular}{c|ccc|ccc|ccc}
\hline
\rowcolor[HTML]{FFCCC9} 
     & \multicolumn{3}{c|}{\cellcolor[HTML]{FFCCC9}MNIST} & \multicolumn{3}{c|}{\cellcolor[HTML]{FFCCC9}FMNIST} & \multicolumn{3}{c}{\cellcolor[HTML]{FFCCC9}CIFAR-10} \\ \hline
\rowcolor[HTML]{C0C0C0} 
Defenses & ER                       & SR        &TC              & ER                       & SR           &TC            & ER                       & SR     &TC                  \\ \hline
FoolsGold w. N/A &  0.10                        &  -      &  -               &      0.18                    &  -          &    -        &      0.52                     &    -         &   -         \\
FoolsGold w. \cite{fung2018mitigating}  &  0.10                        &    0.00        &      0.12       &     0.18                     &      0.00     &     0.23        &  0.52                         &               0.00     & 0.56  \\   
FoolsGold w. Faker  & \textbf{0.63} &  \textbf{1.00}                        &    \textbf{0.01}             &     \textbf{0.49}                     &      \textbf{1.00}      &    \textbf{0.01}        &  \textbf{0.90}                         &               \textbf{1.00}     &   \textbf{0.03}   \\ \hline
\end{tabular}}
\label{sybil_faker}
\end{table}

\if()
\begin{table}[]
\centering
\caption{Evaluation of Faker-based Free-rider Attack.}
\resizebox{\linewidth}{!}{
\begin{tabular}{c|ccc|ccc|ccc}
\hline
\rowcolor[HTML]{FFCCC9} 
     & \multicolumn{3}{c|}{\cellcolor[HTML]{FFCCC9}MNIST} & \multicolumn{3}{c|}{\cellcolor[HTML]{FFCCC9}FMNIST} & \multicolumn{3}{c}{\cellcolor[HTML]{FFCCC9}CIFAR-10} \\ \hline
\rowcolor[HTML]{C0C0C0} 
Defenses & MA                       & TA        &TC              & MA                       & TA          &TC             & MA                        & TA     &TC                  \\ \hline
NC  &  0.10                        &  -     &  -               &      0.82                    &   0.64          &            &      0.57                     &    0.45         &            \\
NC w. Faker  &  \textbf{0.93}                        &    \textbf{0.52}        &             &     \textbf{0.73}                     &      \textbf{0.32}     &             &  \textbf{0.46}                         &               \textbf{0.23}     &   \\   
NC w. Faker  &  \textbf{0.93}                        &    \textbf{0.52}     &                &     \textbf{0.73}                     &      \textbf{0.32}      &            &  \textbf{0.46}                         &               \textbf{0.23}     &      \\ \hline
FM  &   0.97                       &    0.99         &           &     0.82                     &       0.67          &         &         0.57                 & 0.44         &                \\
FM w. Faker   &    \textbf{0.94}                       &      \textbf{0.48}    &                &        \textbf{0.74}                   &        \textbf{0.23}       &            &          \textbf{0. 51}                 &     \textbf{0.26}            &          \\
FM w. Faker   &    \textbf{0.94}                       &      \textbf{0.48}           &         &        \textbf{0.74}                   &        \textbf{0.23}         &          &          \textbf{0. 51}                 &     \textbf{0.26}       &              \\\hline
DF  &   0.97                       &     0.99          &          &       0.82                   &     0.65          &           &     0.23                     &     0.21               &      \\
DF w. Faker   &    \textbf{0.95}                       &      \textbf{0.56}         &          &       \textbf{0.71}                    &       \textbf{0.32}           &         &   \textbf{0.19}                         &   \textbf{0.11}           &             \\
DF w. Faker   &    \textbf{0.95}                       &      \textbf{0.56}      &             &       \textbf{0.71}                    &       \textbf{0.32}          &          &   \textbf{0.19}                         &   \textbf{0.11}            &            \\ \hline
\end{tabular}}
\end{table}
\fi
\textbf{Impacts of $T$ on Faker's Performance.} 
We explore the effect of $T$ on Faker's performance by varying its value. The results are presented in Table \ref{error_t} and Table \ref{time_cost_t}. We can see that increasing the value of $T$ will not greatly improve the overall attack performance, but it will significantly increase the time cost required to launch the attack. Therefore, we suggest that an attacker can choose a smaller $T$ in practice to achieve the expected attacking goal while reducing the time consumption.


\begin{table}
\centering
\caption{Error rates of different values of $T$ with $n=100$, $m=20$, and $c=5$.}
\resizebox{\linewidth}{!}{
\begin{tabular}{c|ccc|ccc|ccc} 
\hline
\rowcolor[HTML]{FFCCC9} 
& \multicolumn{3}{c|}{MNIST} & \multicolumn{3}{c|}{FMNIST} & \multicolumn{3}{c}{CIFAR-10} \\ \hline
\rowcolor[HTML]{C0C0C0} 
Defenses   & T=2 & T=J/2 & T=J   & T=2 & T=J/2 & T=J   & T=2 & T=J/2 & T=J  \\ \hline
\hline
KM     &   0.66  & 0.66   & 0.67    &  0.90   & 0.90   &  0.90  &  0.76  &  0.77  & 0.77  \\
NC     &   0.18  & 0.18   & 0.19    &  0.34   &  0.35  & 0.35   &  0.90  &  0.90  & 0.90  \\
FT     &   0.90  &  0.90  & 0.90    &  0.60   & 0.61   & 0.63   &  0.85  & 0.85   & 0.85  \\
FM     &   0.43  &  0.43  &  0.45   &  0.90   & 0.90   & 0.90   &  0.90  &  0.90  & 0.90  \\
DF     &   0.66  &  0.67  & 0.69    &  0.69   &  0.69  & 0.69   &  0.90  &  0.90  & 0.90  \\
SF     &   0.90  &  0.90  & 0.90    &  0.90   &  0.90  & 0.90   &  0.90  &  0.90  & 0.90  \\
\hline
\end{tabular}}
\label{error_t}
\end{table}

\begin{table}
\centering
\caption{Time cost  of different values of $T$ with $n=100$, $m=20$, and $c=5$. The time cost is measured in seconds.}
\resizebox{\linewidth}{!}{
\begin{tabular}{c|ccc|ccc|ccc} 
\hline
\rowcolor[HTML]{FFCCC9} 
              & \multicolumn{3}{c|}{MNIST} & \multicolumn{3}{c|}{FMNIST} & \multicolumn{3}{c}{CIFAR-10} \\ 
\hline
\rowcolor[HTML]{C0C0C0} 
Defenses          & T=2 & T=J/2 & T=J          & T=2 & T=J/2 & T=J   & T=2 & T=J/2 & T=J         \\ 
\hline
\hline
KM     &  0.827   &  2.365  & 6.236    & 6.234    &  14.782  & 45.247   & 4.242   & 12.982   & 53.652  \\
NC     &  0.015   &  0.056  & 0.324    & 0.083    &  0.153   & 0.832    & 0.093   &  0.193   & 1.832  \\
FT     &  0.015   &  0.082  & 0.532    & 0.016    &  0.752   & 7.523    & 0.030   &  0.182  &  8.672 \\
FM     &  0.183   &  0.621  & 3.213    & 0.336    &  1.236   & 10.672   & 0.245   &  1.232  & 10.982  \\
DF     &  0.214   &  1.062  & 10.873   & 0.173    &  1.082   & 9.624    & 0.225   &  1.922  & 14.924  \\
SF     &  0.152   &  0.872  & 6.342    & 0.235    &  1.762   & 12.732   & 0.242   &  1.821  &  20.124 \\
\hline
\end{tabular}}
\label{time_cost_t}
\end{table}



\textbf{Note:} We also present a large amount of extra experiments in Appendix \ref{ex_faker}. Specifically, we provide the experiments based on three classic models, i.e., LeNet-5 \cite{lecun1998gradient} for MNIST, adjusted LeNet for FMNIST, and AlexNet \cite{krizhevsky2012imagenet} for CIFAR-10. Besides, we conduct experiments based on other larger datasets, i.e., CIFAR-100 \cite{krizhevsky2009learning}, HAM10000 \cite{tschandl2018ham10000}, tiny ImageNet \cite{le2015tiny}, and Reuters\cite{Reuters21578}. We also adopt Dirichlet distribution to get the non-IID data \cite{zhao2018federated} for experiments.

\section{Similarity of Partial Parameters}\label{spp_fa}
Intuitively, to defend against Faker, we should address the vulnerabilities of similarity metrics. As discussed in Section \ref{sec_EvilTwin}, $L_2$ norm, Euclidean distance, and cosine similarity implemented in defending model poisoning attacks require that the input local model has multiple dimensions; thus, Faker can find multiple combinations of scalars to satisfy the similarity requirements. 
To resist Faker, 
we propose a method, \textit{similarity of partial parameters (SPP)}, to calculate the similarity of partial parameters of local models. 
The basic idea is to randomly select partial parameters for evaluation by the defender in each round. The specific number of selected parameters and parameter indexes can vary in rounds. The selection process is done after collecting all the submissions so that the attackers cannot know how their local models will be evaluated. 

\subsection{Security Analysis of SPP}\label{secure_spp}
Basically, SPP tries to improve the robustness of similarity metrics by selecting $J'$ parameters randomly among $J$ parameters during the local model evaluation, thus the selected $J'$ parameters are not known by the attackers. When launching a Faker attack, the attacker usually uses all the parameters of the local model to generate the poisoned model, that is, $\overline{w}_i=\alpha_i \otimes w_i$. Thus, all parameters in $w_i$ affect the generation of $\overline{w}_i$. Faker's core idea is to approximate $S(\overline{w}_i, w_r)$ by using $S(\overline{w}_i, w_i)$, and such an approximation will lead to a large bias in exposing poisoned local models once all parameters cannot be used in the evaluation. The attacker can also randomly select some parameters to generate $\overline{w}_i$, but the selected parameters are unlikely to be identical to those selected by the defender, and will still expose the poisoned local model. In general, Faker tries to manipulate the high-dimensional local model as a whole to forge a poisoned model to pass the evaluation of similarity metrics, and the evaluated result is not as expected if only partial parameters are evaluated. The above analysis proves the effectiveness of the SPP as a new evaluation method in defending against Faker.

\if()
We discuss three vectors $w_i$, $w_i$, and $\overline{w}_i$ with $J'$ selected parameters and a corresponding scalar vector $\alpha_i$ in this part. We need to prove that $S(w_i,w_i)$ significantly differs from $S(\overline{w}_i,w_i)$. 

\begin{proof}
When there is no attack, we can get $S(w_i,w_i)\to \theta$ since the parameters in both vectors have similar values. To generate effective poisoned local models, Faker has to enlarge the bias $\xi_j$. According to \textbf{Theorem} \ref{xi_j}, the calculation of $\xi_j$ is highly correlated to the value of $\alpha_{i,j}$. Since $\overline{w}_i$ is the same as $w_i$ when $\alpha_{i,j}=1$, the value of $\xi_j$ would be greater when the value of $\alpha_{i,j}$ is significantly larger or smaller than $1$.
Based on this observation, we can say that a successful Faker attack should have some scalars whose value should not equal 1. Besides, since Faker generates poisoned local models by considering all the parameters, the similarity requirement $\theta$ can not be satisfied once the 'integrity' of the parameters is broken. Take the $L_2$ norm as an example. Our goal is to allow $S(\overline{w}_i,w_i)=\frac{L(\overline{w}_i)}{L(w_i)}=\frac{\sum_{j=1}^{J'}\alpha_{i,j}^2 w_{i,j}^2}{\sum_{i=1}^{J'}w_{i,j}^2}\to 1$ and $\overline{w}_{i,j}=\alpha_{i,j}w_{i,j}$; when $\alpha_{i,j}=1$ for any $j$, then we get $S(\overline{w}_i,w_i)=S(w_i,w_i)\to 1$; while if $\alpha_{i,j}\neq 1$, then the above requirement can not always be met. Thus, the poisoned value will be detected. In general, Faker tries to manipulate the high-dimensional local model as a whole to forge a poisoned model to pass the evaluation of similarity metrics, and the evaluated result is not as expected if only partial parameters are evaluated. The above analysis proves the effectiveness of the SPP as a new evaluation method in defending against Faker.
\end{proof}
\fi
\subsection{Evaluation of SPP}
With the similar experimental settings in Section \ref{ev}, we conduct preliminary experiments to test the efficiency of SPP when defending against Faker with $n=100$ and varied $m$. Besides, we allow the defender to randomly choose about $\frac{J}{2}$ parameters to be evaluated during evaluation. Once the poisoned models are detected, the defender will discard them.
The results in Table \ref{spp} show that SPP has similar performance as the benchmark defense ERR \cite{fang2020local} and can defend against Faker even when there are 50\% attackers. Please note that ERR requires the defender to evaluate the local models with clean data and reject the local model with an abnormal error rate, it can detect malicious models effectively; 
while SPP does not need any other extra information in the evaluation, it can still have the same performance as ERR does.
In addition, the results in Table \ref{time_spp} indicate that SPP does not require intensive computational resources even when the number of selected parameters is large and SPP outperforms ERR in time cost. Please refer to Appendix \ref{ex_experiments} for more experimental evaluations of SPP on larger datasets and deep models.


\begin{table}
\centering
\caption{SPP Against Faker when $n=100$ and $c=5$. The performance is measured by error rates.}
\resizebox{\linewidth}{!}{
\begin{tabular}{c|ccc|ccc|ccc} 
\hline
\rowcolor[HTML]{FFCCC9} 
              & \multicolumn{3}{c|}{MNIST} & \multicolumn{3}{c|}{FMNIST} & \multicolumn{3}{c}{CIFAR-10} \\ 
\hline
\rowcolor[HTML]{C0C0C0} 
   Defenses    & $10\%$ & $20\%$ & $50\%$          & $10\%$ & $20\%$ & $50\%$      & $10\%$ & $20\%$ & $50\%$     \\ 
\hline
\hline
KM          & 0.64     &  0.66  &      0.90         &  0.61   & 0.90   & 0.90    &  0.70    & 0.76   & 0.90            \\
KM  w. ERR        & 0.48     &  0.50  &     0.70               &  0.40    & 0.40  & 0.42   &  0.64   & 0.64   & 0.67              \\

KM  w. SPP     & 0.48     &  0.50  &     0.70               &  0.40    & 0.40  & 0.42   &  0.64   & 0.64   & 0.67             \\
\hline
NC &     0.16 &   0.18 &     0.34      &   0.22   &  0.24  & 0.39   &    0.90  & 0.90  & 0.90               \\
NC w. ERR   & 0.11    &  0.11  &     0.12         &  0.18    & 0.20   & 0.32   & 0.52   & 0.52   & 0.53            \\

NC  w. SPP        & 0.11    &  0.11  &     0.12         &  0.18    & 0.20   & 0.32   & 0.52   & 0.52   & 0.53            \\
\hline
FT      &     0.90 &  0.90  & 0.90    &  0.55   &  0.60  &   0.72  &    0.78  &  0.85  &   0.90          \\
FT w. ERR       &     0.10 &  0.10  &   0.12   &  0.18   & 0.18  &  0.19  &   0.55   &  0.55  & 0.56      \\
FT w. SPP      &     0.10 &  0.10  &   0.12   &  0.18   & 0.18  &  0.19  &   0.55   &  0.55  & 0.56        \\
\hline
FM         &    0.20  & 0.43   &  0.60             &  0.30    &  0.90  &   0.90    &  0.90    & 0.90   & 0.90           \\
FM w. ERR         &0.10      & 0.10    &   0.13           &  0.19    & 0.19   & 0.22   &  0.52   & 0.52   & 0.54        \\
FM w. SPP        &0.10      & 0.10    &   0.13           &  0.19    & 0.19   & 0.22   &  0.52   & 0.52   & 0.54                \\
\hline
DF     &  0.59    &  0.66  & 0.81               & 0.55     & 0.69   & 0.90     &   0.90   & 0.90   & 0.90             \\
DF w. ERR      &  0.09    &  0.09 &   0.10             &   0.19   &  0.19  &    0.21    & 0.52     &  0.52  &   0.53        \\
DF w. SPP    &  0.09    &  0.09 &   0.10             &   0.19   &  0.19  &    0.21    & 0.52     &  0.52  &   0.53        \\
\hline
SF      &   0.90   & 0.90   & 0.90               &   0.90   & 0.90   &   0.90    &  0.90    & 0.90  & 0.90           \\
SF w. ERR     & 0.09     &0.09   &     0.10           &   0.54   & 0.54   & 0.57    &  0.56    &  0.56  &     0.55   \\
SF w. SPP     & 0.09     &0.09   &     0.10           &   0.54   & 0.54   & 0.57    &  0.56    &  0.56  &     0.55        \\
\hline
\end{tabular}}
\label{spp}
\end{table}

\if()
\begin{table}
\centering
\caption{SPP against Faker when $n=100$ and $c=5$. The performance of SPP is measured by error rates.}
\resizebox{\linewidth}{!}{
\begin{tabular}{c|ccc|ccc|ccc} 
\hline
\rowcolor[HTML]{FFCCC9} 
              & \multicolumn{3}{c|}{MNIST} & \multicolumn{3}{c|}{FMNIST} & \multicolumn{3}{c}{CIFAR-10} \\ 
\hline
\rowcolor[HTML]{C0C0C0} 
\tiny \diagbox{ AGM}{m/n}          & $10\%$ & $20\%$ & $50\%$          & $10\%$ & $20\%$ & $50\%$      & $10\%$ & $20\%$ & $50\%$     \\ 
\hline
\hline
KM          & 0.64     &  0.66  &      0.90         &  0.61   & 0.90   & 0.90    &  0.70    & 0.76   & 0.90            \\
KM  w. ERR       & \textbf{0.48}     &  \textbf{0.50}  &     \textbf{0.70}          &  \textbf{0.40}    & \textbf{0.40}   & \textbf{0.42}   &  \textbf{0.64}   & \textbf{0.64}   & \textbf{0.67}              \\

KM  w. SPP        & \textbf{0.48}     &  \textbf{0.50}  &     \textbf{0.70}          &  \textbf{0.40}    & \textbf{0.40}   & \textbf{0.42}   &  \textbf{0.64}   & \textbf{0.64}   & \textbf{0.67}              \\

KM  w. SPP        & \textbf{0.48}     &  \textbf{0.50}  &     \textbf{0.70}          &  \textbf{0.40}    & \textbf{0.40}   & \textbf{0.42}   &  \textbf{0.64}   & \textbf{0.64}   & \textbf{0.67}              \\
\hline
NC &     0.16 &   0.18 &     0.34      &   0.22   &  0.24  & 0.39   &    0.90  & 0.90  & 0.90               \\
NC w. SPP &     \textbf{0.11} &   \textbf{0.11} &   \textbf{0.12}     &   \textbf{0.18}   &  \textbf{0.20}  & \textbf{0.32}   &  \textbf{0.52}   & \textbf{0.52}   &  \textbf{0.53}              \\

NC  w. SPP        & \textbf{0.48}     &  \textbf{0.50}  &     \textbf{0.70}          &  \textbf{0.40}    & \textbf{0.40}   & \textbf{0.42}   &  \textbf{0.64}   & \textbf{0.64}   & \textbf{0.67}              \\

NC w. SPP        & \textbf{0.48}     &  \textbf{0.50}  &     \textbf{0.70}          &  \textbf{0.40}    & \textbf{0.40}   & \textbf{0.42}   &  \textbf{0.64}   & \textbf{0.64}   & \textbf{0.67}              \\
\hline
FT      &     0.90 &  0.90  & 0.90    &  0.55   &  0.60  &   0.72  &    0.78  &  0.85  &   0.90          \\
FT w. SPP      &     \textbf{0.10} &  \textbf{0.10}  &    \textbf{0.12}    &  \textbf{0.18}   &  \textbf{0.18}  &  \textbf{0.19}   &   \textbf{0.55}   &  \textbf{0.55}  & \textbf{0.56}        \\
FT w. SPP      &     \textbf{0.10} &  \textbf{0.10}  &    \textbf{0.12}    &  \textbf{0.18}   &  \textbf{0.18}  &  \textbf{0.19}   &   \textbf{0.55}   &  \textbf{0.55}  & \textbf{0.56}        \\
FT w. SPP      &     \textbf{0.10} &  \textbf{0.10}  &    \textbf{0.12}    &  \textbf{0.18}   &  \textbf{0.18}  &  \textbf{0.19}   &   \textbf{0.55}   &  \textbf{0.55}  & \textbf{0.56}        \\
\hline
FM         &    0.20  & 0.43   &  0.60             &  0.30    &  0.90  &   0.90    &  0.90    & 0.90   & 0.90           \\
FM w. SPP        &\textbf{0.10}      & \textbf{0.10}    &    \textbf{0.13}            &  \textbf{0.19}    & \textbf{0.19}   & \textbf{0.22}   &  \textbf{0.52}   & \textbf{0.52}   & \textbf{0.54}                \\
FM w. SPP        &\textbf{0.10}      & \textbf{0.10}    &    \textbf{0.13}            &  \textbf{0.19}    & \textbf{0.19}   & \textbf{0.22}   &  \textbf{0.52}   & \textbf{0.52}   & \textbf{0.54}                \\
FM w. SPP        &\textbf{0.10}      & \textbf{0.10}    &    \textbf{0.13}            &  \textbf{0.19}    & \textbf{0.19}   & \textbf{0.22}   &  \textbf{0.52}   & \textbf{0.52}   & \textbf{0.54}                \\
\hline
DF     &  0.59    &  0.66  & 0.81               & 0.55     & 0.69   & 0.90     &   0.90   & 0.90   & 0.90             \\
DF w. SPP    &  \textbf{0.09}    &  \textbf{0.09} &   \textbf{0.10}             &   \textbf{0.19}   &  \textbf{0.19}  &    \textbf{0.21}    & \textbf{0.52}     &  \textbf{0.52}  &   \textbf{0.53}        \\
DF w. SPP    &  \textbf{0.09}    &  \textbf{0.09} &   \textbf{0.10}             &   \textbf{0.19}   &  \textbf{0.19}  &    \textbf{0.21}    & \textbf{0.52}     &  \textbf{0.52}  &   \textbf{0.53}        \\
DF w. SPP    &  \textbf{0.09}    &  \textbf{0.09} &   \textbf{0.10}             &   \textbf{0.19}   &  \textbf{0.19}  &    \textbf{0.21}    & \textbf{0.52}     &  \textbf{0.52}  &   \textbf{0.53}        \\
\hline
SF      &   0.90   & 0.90   & 0.90               &   0.90   & 0.90   &   0.90    &  0.90    & 0.90  & 0.90           \\
SF w. SPP     & \textbf{0.09}     & \textbf{0.09}   &     \textbf{0.10}           &   \textbf{0.54}   & \textbf{0.54}   & \textbf{0.57}     &   \textbf{0.56}    &  \textbf{0.56}   &     \textbf{0.55}        \\
SF w. SPP     & \textbf{0.09}     & \textbf{0.09}   &     \textbf{0.10}           &   \textbf{0.54}   & \textbf{0.54}   & \textbf{0.57}     &   \textbf{0.56}    &  \textbf{0.56}   &     \textbf{0.55}        \\
SF w. SPP     & \textbf{0.09}     & \textbf{0.09}   &     \textbf{0.10}           &   \textbf{0.54}   & \textbf{0.54}   & \textbf{0.57}     &   \textbf{0.56}    &  \textbf{0.56}   &     \textbf{0.55}        \\
\hline
\end{tabular}}
\label{spp}
\end{table}
\fi

\begin{table}
\centering
\caption{Time cost of SPP with $n=100$ and $c=5$. The time consumption is measured in seconds.}
\resizebox{\linewidth}{!}{
\begin{tabular}{c|ccc|ccc|ccc} 
\hline
\rowcolor[HTML]{FFCCC9} 
              & \multicolumn{3}{c|}{MNIST} & \multicolumn{3}{c|}{FMNIST} & \multicolumn{3}{c}{CIFAR-10} \\ 
\hline
\rowcolor[HTML]{C0C0C0} 
Defenses          & $J/10$ & $J/4$ & $9J/10$        & $J/10$ & $J/4$ & $9J/10$     & $J/10$ & $J/4$ & $9J/10$      \\ 
\hline
\hline
 ERR         & \multicolumn{3}{c|}{1.118} & \multicolumn{3}{c|}{2.013} & \multicolumn{3}{c}{2.76}           \\
 \hline
SPP     &  0.012    & 0.015  &     0.018          & 0.021   &  0.024 &   0.028    & 0.028  & 0.031&   0.036     \\
\hline
\end{tabular}}
\label{time_spp}
\end{table}

\if()
\section{Limitations and Future Research Directions}\label{d_f}
In this section, we first discuss the limitations of this paper, and then we share our opinions toward future research directions.

\subsection{Limitations of Our Work}
We think our paper has the following limitations, but our theories and findings are still held despite these limitations.

\subsubsection{EvilTwin can be more efficient by carefully setting scalars} The essence of EvilTwin is to find a suitable scalar for each parameter. Although we provide an interval for selecting the value of the scalars, especially for $L_2$ norm and Euclidean distance, this interval is derived from the overall perspective, which means that this interval is sometimes conservative and does not allow the variance of the selected scalars to be too significant. In this way, the generated fake vector still retains some characteristics of the reference vector, which does not guarantee that the fake vector can maximize its negative impacts.

\subsubsection{The evaluated local models have a fixed form} In this paper, we assume that the local models submitted by the clients will be evaluated in the form of vectors, but in practice, they may be evaluated directly in tensor format. In this way, the methods of forging and evaluating local models need to be adjusted accordingly.

\subsubsection{The attackers are allowed to collude with each other} We assume that attackers can communicate with each other and adopt the same attack strategy when attacking Krum, but in practice, an attacker may not be aware of the existence of another attacker. Although this assumption is allowed in most existing model poisoning attacks, we wish Faker could empower each attacker to launch the attack independently.

\subsection{Future Research Directions}
In addition to addressing the above limitations, the following research directions are promising.
\subsubsection{The way that EvilTwin divides elements can be improved} We do not design the partitioning method of EvilTwin in detail, but the partitioning method affects the efficiency of EvilTwin and the quality of the generated data. We can consider maximizing the variance between the generated and original data to divide the data. Based on this, we can find the optimal division method. 
\subsubsection{Faker can be designed to attack non-similarity-based AGMs} Faker is for attacking similarity-based AGMs, while the AGMs based on other evaluation metrics can also be attacked by Faker. The critical step is to adjust EvilTwin with other metrics. We first have to analyze the deficiencies of other evaluation metrics, which generally need to satisfy EvilTwin's basic idea that there exists a vector $\overline{X}\neq X$ such that $S(X, Y)=S(\overline{X}, Y)=\theta$. Then we should follow the \textbf{Algorithm} \ref{al_ev} to get the poisoned local model. EvilTwin obtains different scalar equations for different metrics. After that, we can follow \textbf{Algorithm} 2 to launch the attack on a single device.

\subsubsection{Faker can be extended to backdoor attacks} In this work, Faker is designed to launch model poisoning attacks to degrade the overall performance of FL. Backdoor attacks are different from model poisoning attacks in that they aim to attack the performance of the global model on a certain labeled data, without compromising the overall performance. Model poisoning attacks are untargeted, while backdoor attacks are targeted. Furthermore, the methods for implementing backdoor attacks are different from those for model poisoning attacks. We introduce how to extend Faker to backdoor attacks, which will be our future work. Recall Faker's two foundations, i.e. I\&B and EvilTwin, we need to adapt these two mechanisms to achieve the purpose of launching backdoor attacks. First, the goals of I\&B should be adjusted to maximize the importance of poisoned local models during aggregation and minimize the bias of parameters. Then, we have to generate poisoned local models with EvilTwin according to different similarity metrics. Next, we should formulate backdoor attacks with Faker following \textbf{Algorithm} \ref{al_faker}. In the end, by solving the formulated problem, the attackers can get the optimal backdoor attack strategies.

\subsubsection{Lightweight Defenses are required} \label{light_defense} We evaluate the time consumption of different similarity-based AGMs with an NVIDIA A100 40GB GPU  when there are 10000 clients and the results are presented in Table \ref{time_agms}. We can see that even on a device equipped with a powerful GPU, the time consumed to complete the aggregation of models for 10,000 devices is huge and some AGMs can exhaust the computing power of the device so that fail to complete aggregation. 
In addition, we are only testing some basic deep learning models, and the aggregation time required for a large model, such as a large language model, could be enormous. Currently, in the field of FL security, the proposed secure AGMs are becoming more and more complicated, and although they are effective in defending against attacks, they may not be suitable for deployment in large-scale FL systems. Thus, lightweight AGMs are required in the future.

\begin{table}[ht]
\centering
\caption{Time consumption of different AGMs when $n=10000$. '-' means that the test cannot be continued because the device has run out of memory after a long period of time. The time cost is measured in seconds.}
\arrayrulecolor{black}
\resizebox{\linewidth}{!}{
\begin{tabular}{c|c|c|c|c|c|c|c} 
\hline
\rowcolor[HTML]{FFCCC9} 
             Dataset    & FA & KM & NC & FT & FM & DF & SF \\
\hline
\hline
MNIST   & 132.46   & -   & 439.27   & 1295.86  &  -  & 371.01 &  348.66 \\ \hline
FMNIST  &   423.84  &  - & -  &-    &  - &  -  &  - \\\hline
CIFAR-10  &  212.92  &  -  &  473.53  & 1542.37    &  - & 1071.23 & 419.81\\
\arrayrulecolor{black}\cline{1-1}\arrayrulecolor{black}\cline{2-8}
\end{tabular}}
\arrayrulecolor{black}
\label{time_agms}
\end{table}

\subsubsection{Robust Local Model Evaluation} 
We argue that there are several principles for robust local model evaluation. First of all, a robust evaluation metric is required. In this work, we have proved that the widely adopted similarity metrics, e.g., Euclidean distance, $L_2$ norm, and cosine similarity, are insufficient in securing the robustness of FL. Thus, we have to develop a new and robust metric to secure FL. Such a metric should not only be simple and robust but also have good statistical properties, i.e. it needs to be able to distinguish whether the local model is benign or malicious in different training situations. Next, which parameters will be used for model evaluation should not be known to the clients. Even though the detection mechanism and aggregation method can be known by the attackers, they cannot attack the global model if they do not know which parameters will be evaluated. This ensures that the attackers cannot forge data to fool the evaluation metric. Thus, we have to design schemes to assist the aggregator in hiding the details of model evaluation. Then, the designed mechanism should be flexible, which means that it can be applied to the most of existing FL systems, including robust and non-robust frameworks regardless of the data distribution, the type of attacks, and the convexity of the applied loss function. Last but not least, the time and space complexities of the designed detection mechanism should not be ignored. As we have mentioned in Section \ref{light_defense}, even though some methods are efficient in defending against attacks, they are computing resources intensive. 
\fi

\section{Conclusion}\label{conclusion}

In this paper, we first reveal the vulnerabilities of widely used similarity metrics, i.e., $L_2$ norm, Euclidean distance, and cosine similarity. Then, we design a novel and effective model poisoning attack named Faker to undermine FL by leveraging these vulnerabilities. We also extend Faker to other adversarial attacks such as backdoor and Sybil attacks. The extensive experimental results demonstrate that Faker outperforms the benchmark attacks. In addition, a novel model evaluation strategy, SPP, is proposed to defend against Faker. To the best of our knowledge, this work is the first step in studying the deficiencies of similarity metrics.


\bibliographystyle{plain}
\bibliography{reference.bib}

\begin{appendices}

\section*{Appendix} \label{ap}

\if()
\begin{table}[t]
\centering
\caption{The Table of Key Abbreviations.}
\begin{tabular}{c|c}
\hline
\rowcolor[HTML]{FFCCC9} 
Abbreviations & Meanings \\ \hline 
\hline
SPP     & Similarity of Partial Parameters         \\ \hline 
KM     & Krum         \\ \hline 
NC     & Norm-clipping         \\ \hline 
FT     & FLTrust         \\ \hline 
FM     & FLAME         \\ \hline 
DF     & DiverseFL         \\ \hline 
SF     & ShieldFL         \\ \hline 
FA     & FedAvg         \\ \hline 
\end{tabular}
\label{abbreviation}
\end{table}
\fi

\if()

\section{Dive into EvilTwin}\label{dev}

\subsection{Vulnerabilities of Similarity Metrics}\label{vul_sim}
The below analysis is based on the mathematical notations in Section \ref{ge_ev}. We employ mathematical induction to analyze the deficiencies of the similarity metrics. Proving that these three similarity metrics are unsafe is essentially proving that there exists more than one scalar combination that can satisfy $S(\overline{X},Y)=S(X,Y)=\theta$. We then analyze the security of each similarity metric as below.

\subsubsection{Vulnerabilities of $L_2$ Norm}
When $J=1$, we have $L(\overline{X})=([\sigma_1 x_1]^2)^\frac{1}{2}=\theta$, and $\sigma_1=[\frac{\theta_1^2}{x_1^2}]^\frac{1}{2}$. Thus, the value of $\sigma_1$ is not unique when the numerator  is not zero.

When $J=J'$, assume $\sigma_j$ for any $j<J'$ is known, we have $L(\overline{X})=([\sum_{-j=1}^{J'-1}\sigma_{-j}^2x_{-j}^2+\sigma_{J'}x_{J'}^2)^\frac{1}{2}=\theta$ where $\sigma_{-j}$ and $x_{-j}$ are any scalar besides $\sigma_{J'}$ and any element besides $x_{J'}$, respectively . We can get $\sigma_{J'}=(\frac{\theta^2-\sum_{-j=1}^{J'-1}\sigma_{-j}^2x_{-j}^2}{x_{J'}^2})^\frac{1}{2}$. Thus, the value of $\sigma_{J'}$ is not unique when the numerator  is not zero.

When $J=J'+1$, assume assume $\sigma_j$ for any $j<J'+1$ is known, we have $L(\overline{X})=([\sum_{-j=1}^{J'}\sigma_{-j}^2x_{-j}^2+\sigma_{J'+1}x_{J'+1}^2)^\frac{1}{2}=\theta$ where $\sigma_{-j}$ and $x_{-j}$ are any scalar besides $\sigma_{J'+1}$ and any element besides $x_{J'+1}$, respectively . We can get $\sigma_{J'+1}=(\frac{\theta^2-\sum_{-j=1}^{J'}\sigma_{-j}^2x_{-j}^2}{x_{J'}^2})^\frac{1}{2}$. Thus, the value of $\sigma_{J'+1}$ is not unique when the numerator  is not zero. 

Given $\sigma_{J'+1}=(\frac{\theta^2-\sum_{-j=1}^{J'}\sigma_{-j}^2x_{-j}^2}{x_{J'}^2})^\frac{1}{2}$, we can also get $L(\overline{X})=([\sum_{-j=1}^{J'}\sigma_{-j}^2x_{-j}^2+\sigma_{J'+1}x_{J'+1}^2)^\frac{1}{2}=\theta$ by substituting the value of $\sigma_{J'}$.

Thus, we can say that there could be multiple combinations of scalars that can allow $L(\overline{X})=\theta$.

\subsubsection{Vulnerabilities of Euclidean Distance}
When $J=1$, we have $E(\overline{X},Y)&=([\sigma_1 x_1-x_1]^2)^\frac{1}{2}=(\sigma_1^2x_1^2-2\sigma_1 x_1 + x_1^2)^\frac{1}{2}=\theta$. We can get $\sigma_1=\frac{1}{x_{1}}\pm(\frac{1}{x_1^2}-1)^\frac{1}{2}$. Thus, the value of $\sigma_{1}$ is not unique. 

When $J=J'$, assume $\sigma_j$ for any $j<J'$ is known, we have $E(\overline{X},Y)=(({\sigma_{J'}}^2-2\sigma_{J'}+1) x_{J'}+(\sum_{-j=1}^{J'-1} ({\sigma_{-j}}^2-2\sigma_{-j}+1) x_{-j}))^\frac{1}{2}=\theta$ where $\sigma_{-j}$ and $x_{-j}$ are any scalar besides $\sigma_{J'}$ and any element besides $x_{J'}$, respectively . We can get $\sigma_{J'}=\frac{1}{x_{J'}}(x_{J'}\pm [x_{J'}[\sum_{-j=1}^{J'-1}[(2\sigma_{-j}\notag-\sigma_{-j}^2-1)x_{-j}]+{\theta}^2]]^\frac{1}{2})$. Thus, the value of $\sigma_{J'}$ is not unique. 

When $J=J'+1$, assume assume $\sigma_j$ for any $j<J'+1$ is known, we have $E(\overline{X},Y)=(({\sigma_{J'+1}}^2-2\sigma_{J'+1}+1) x_{J'+1}+(\sum_{-j=1}^{J'} ({\sigma_{-j}}^2-2\sigma_{-j}+1) x_{-j}))^\frac{1}{2}=\theta$ where $\sigma_{-j}$ and $x_{-j}$ are any scalar besides $\sigma_{J'+1}$ and any element besides $x_{J'+1}$, respectively. We can get $\sigma_{J'+1}=\frac{1}{x_{J'+1}}(x_{J'+1}\pm [x_{J'+1}[\sum_{-j=1}^{J'}[(2\sigma_{-j}\notag-\sigma_{-j}^2-1)x_{-j}]+{\theta}^2]]^\frac{1}{2})$. Thus, the value of $\sigma_{J'+1}$ is not unique. 

Given $\sigma_{J'+1}=\frac{1}{x_{J'+1}}(x_{J'+1}\pm [x_{J'+1}[\sum_{-j=1}^{J'}[(2\sigma_{-j}\notag-\sigma_{-j}^2-1)x_{-j}]+{\theta}^2]]^\frac{1}{2})$, we can also get $E(\overline{X},Y)=(({\sigma_{J'+1}}^2-2\sigma_{J'+1}+1) x_{J'+1}+(\sum_{-j=1}^{J'} ({\sigma_{-j}}^2-2\sigma_{-j}+1) x_{-j}))^\frac{1}{2}=\theta$ by substituting the value of $\sigma_{J'}$. 

So, we can say that there are multiple combinations of scalars that can allow $E(\overline{X}, Y)=\theta$.

\subsubsection{Vulnerabilities of Cosine Similarity}
When $J=2$, we have $C(\overline{X},Y)=\frac{\sigma_1 x_1^2+\sigma_2 x_2^2}{(x_1^2+x_2^2)^\frac{1}{2}(\sigma_1^2*x_1^2 + \sigma_2^2*x_2^2)^\frac{1}{2}}=\theta$. Assume $\sigma_1$ is known, we can get $\sigma_2=\pm \sigma_{1}(\frac{x_{1}}{x_{2}})^\frac{1}{2}$. Thus, the value of $\sigma_{2}$ is not unique. 

When $J=J'$, we have $C(\overline{X},Y)=\frac{\sum_{j=1}^{J'}\sigma_{j} x_{j}}{(\sum_{j=1}^{j} x_{j})^\frac{1}{2} (\sum_{j=1}^{j} ({\sigma_{j}}^2 x_{j})^\frac{1}{2}}$. Assume $\sigma_{-j}$ and $x_{-j}$ are any scalar besides $\sigma_{J'+1}$ and any element besides $x_{J'+1}$, respectively, we can get Equation (\ref{e_cs}) by setting $T=J'$. Similarly, we can also get Equation (\ref{e_cs}) by allowing $T=J'+1$ when $J=J'+1$. Given Equation (\ref{e_cs}), $C(\overline{X},Y)=\frac{\sum_{j=1}^{J'+1}\sigma_{j} x_{j}}{(\sum_{j=1}^{j} x_{j})^\frac{1}{2} (\sum_{j=1}^{J'+1} ({\sigma_{j}}^2 x_{j})^\frac{1}{2}}$ can also be obtained. Thus, the vector of scalars $\Sigma$ is not unique.

Based on the above analysis, we can conclude that the similarity metrics are not robust. 
\fi

\if()
\subsection{Evaluation of EvilTwin}
We design experiments to verify the impacts of different grouping methods on the computational cost. We calculate the time consumption of EvilTwin when processing vectors of different lengths. The experimental results are shown in Table \ref{evil_twin}. We can see that the time spent on processing vectors is short when divided into two groups, while it takes more time when the number of groups is large. Therefore, the grouping method does affect the computational complexity of EvilTwin significantly.
\begin{table}[ht]
\centering
\caption{Time consumption of different grouping methods of EvilTwin. '-' means that the test cannot be continued because the device is running out of memory after a long period of time.}
\resizebox{\linewidth}{!}{
\begin{tabular}{c|cc|cc}
\hline

\rowcolor[HTML]{FFCCC9} 
                   & \multicolumn{2}{c|}{$J=10^4$}       & \multicolumn{2}{c|}{$J=10^{5}$}      \\ \hline
                   \hline
\rowcolor[HTML]{C0C0C0} 
Methods            & \multicolumn{1}{c|}{$T=2$} & $T=10^4$ & \multicolumn{1}{c|}{$T=2$} & $T=10^5$ \\ \hline
$L_2$ Norm         & \multicolumn{1}{c|}{0.004 }     &  0.252    & \multicolumn{1}{c|}{0.010}      &    5.907    \\ \hline
Euclidean Distance & \multicolumn{1}{c|}{0.006}      &    0.253     & \multicolumn{1}{c|}{0.012}      &    6.880    \\ \hline
Cosine Similarity  & \multicolumn{1}{c|}{0.009}      &   10.95     & \multicolumn{1}{c|}{0.014}      &     -  \\ \hline
\end{tabular}}
\label{evil_twin}
\end{table}

\begin{figure*}[ht]
\subfigure[$L_2$ norm with $\theta=1$.]{
\includegraphics[width=0.235\textwidth]{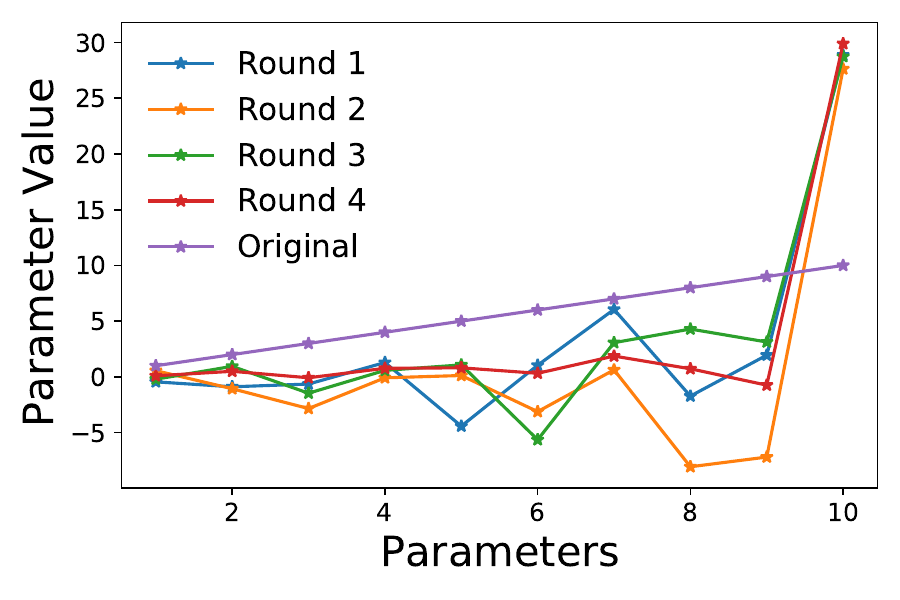}}
\subfigure[$L_2$ norm with $\theta=5$.]{
\includegraphics[width=0.235\textwidth]{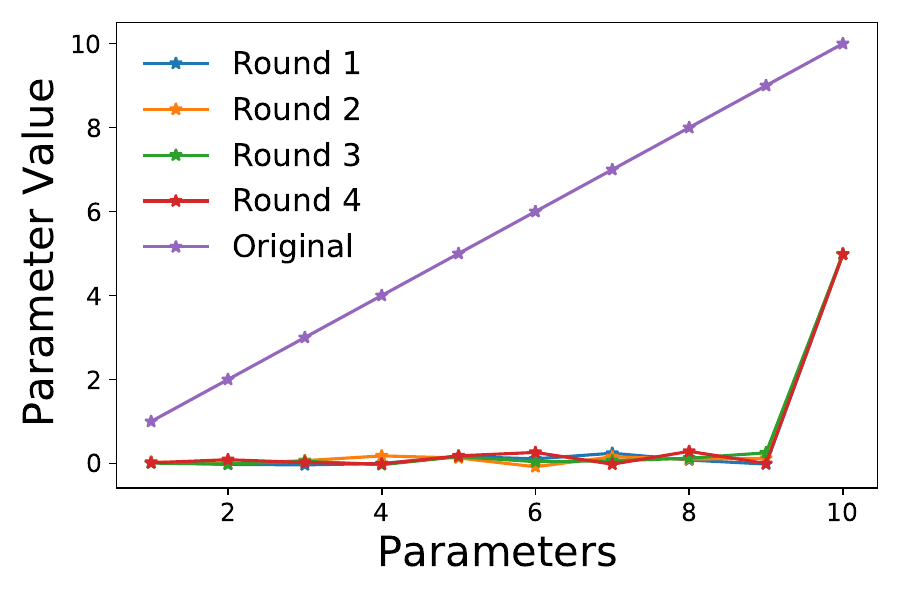}}
\subfigure[$L_2$ norm with $\theta=15$.]{
\includegraphics[width=0.235\textwidth]{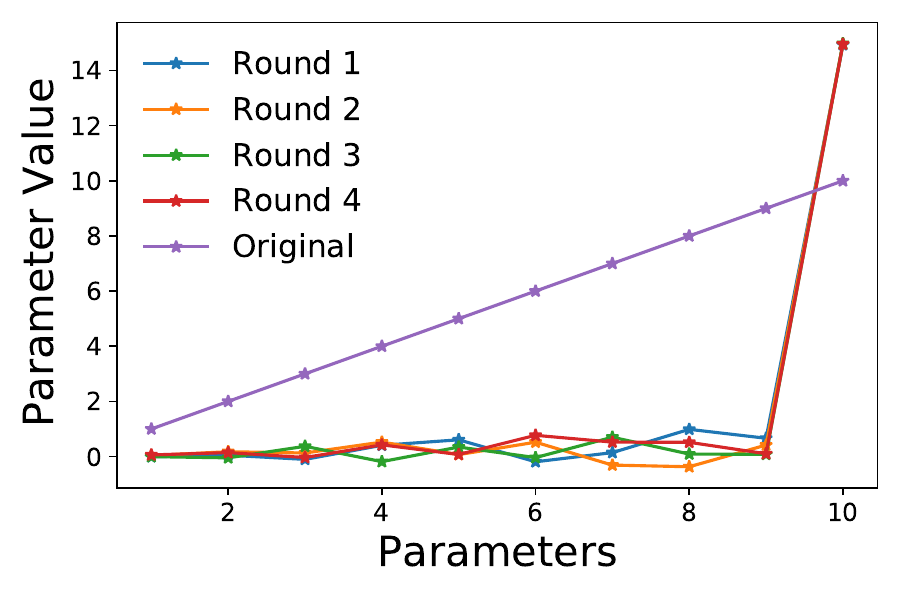}}
\subfigure[$L_2$ norm with $\theta=20$.]{
\includegraphics[width=0.235\textwidth]{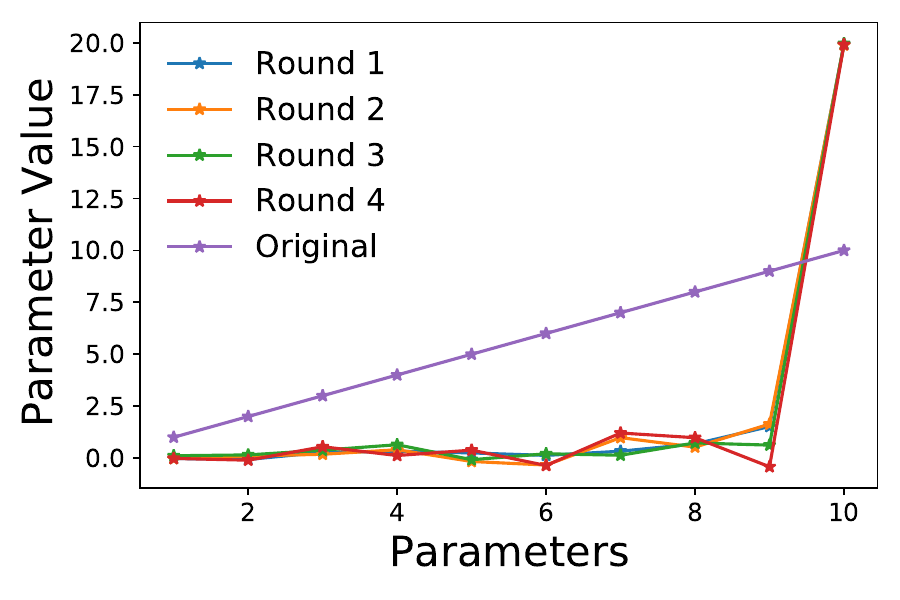}}

\subfigure[Euclidean distance with $\theta=1$.]{
\includegraphics[width=0.235\textwidth]{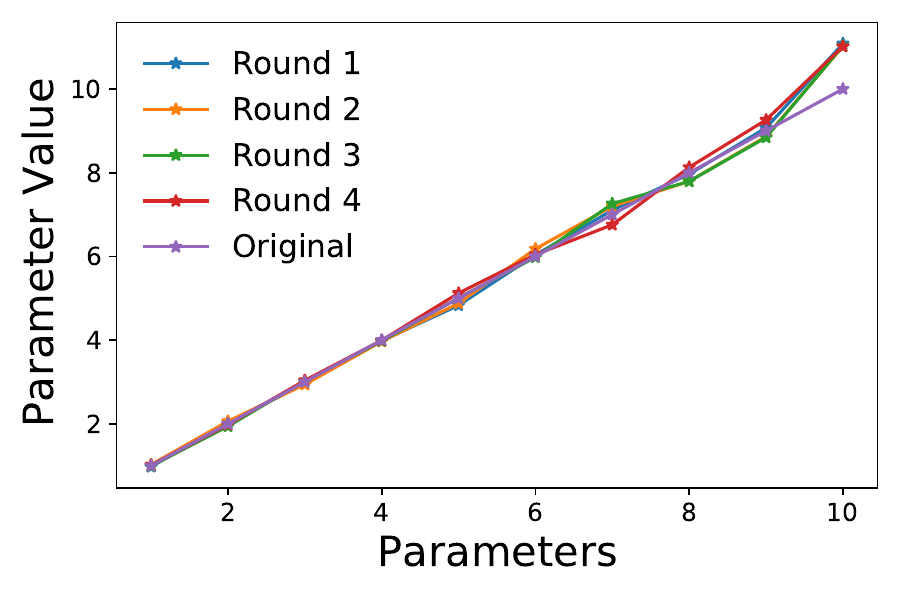}}
\subfigure[Euclidean distance with $\theta=5$.]{
\includegraphics[width=0.235\textwidth]{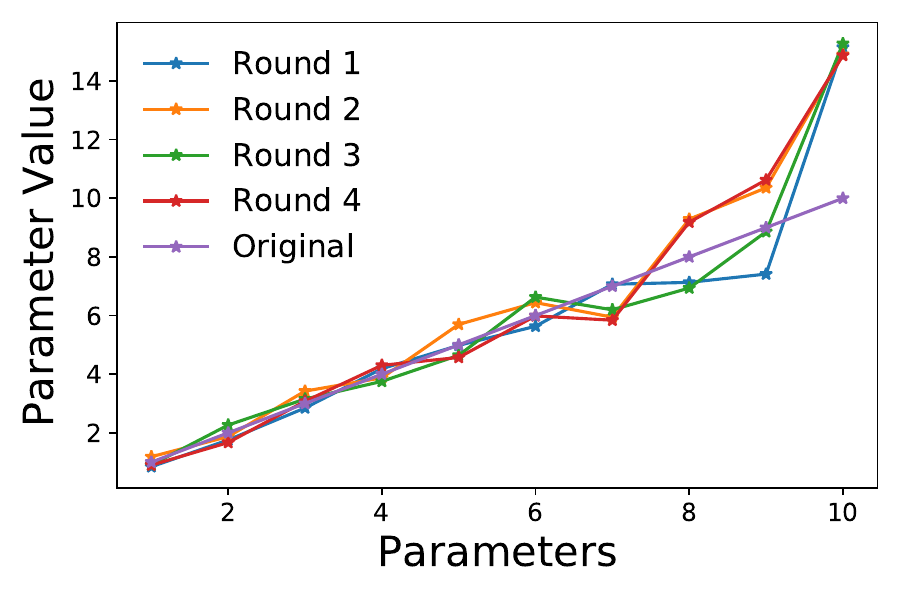}}
\subfigure[Euclidean distance with $\theta=15$.]{
\includegraphics[width=0.235\textwidth]{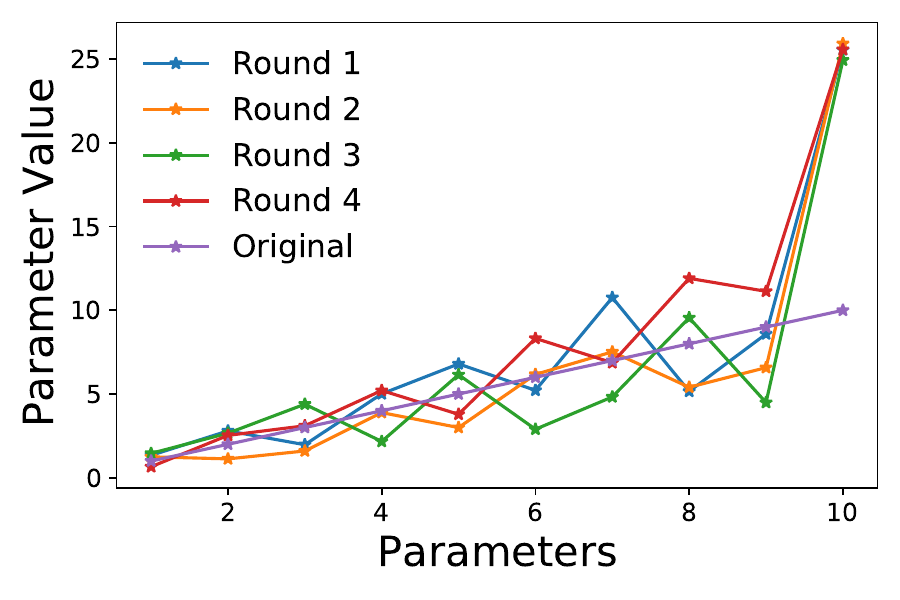}}
\subfigure[Euclidean Distance with $\theta=20$.]{
\includegraphics[width=0.235\textwidth]{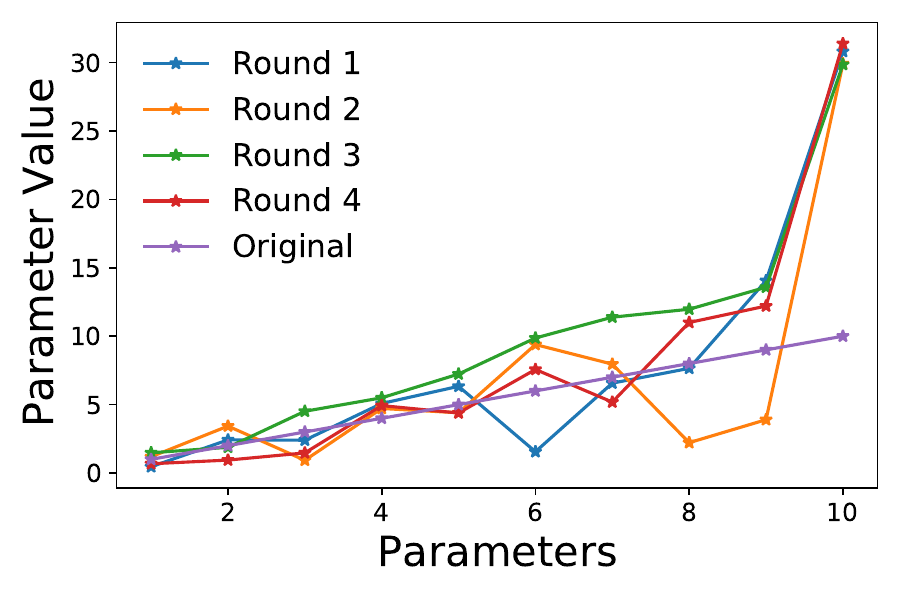}}

\subfigure[Cosine similarity with $\theta=-0.5$.]{
\includegraphics[width=0.235\textwidth]{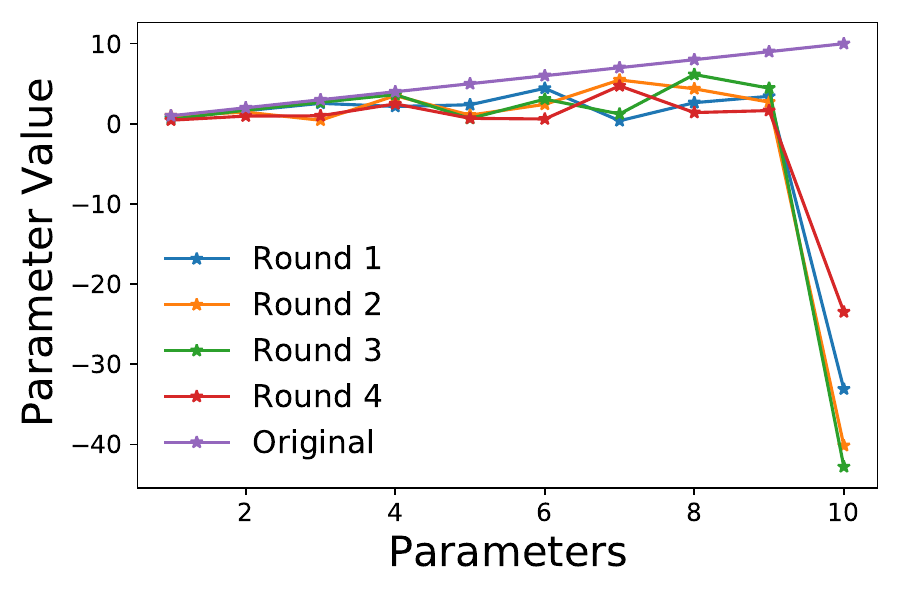}}
\subfigure[Cosine similarity with $\theta=0.1$.]{
\includegraphics[width=0.235\textwidth]{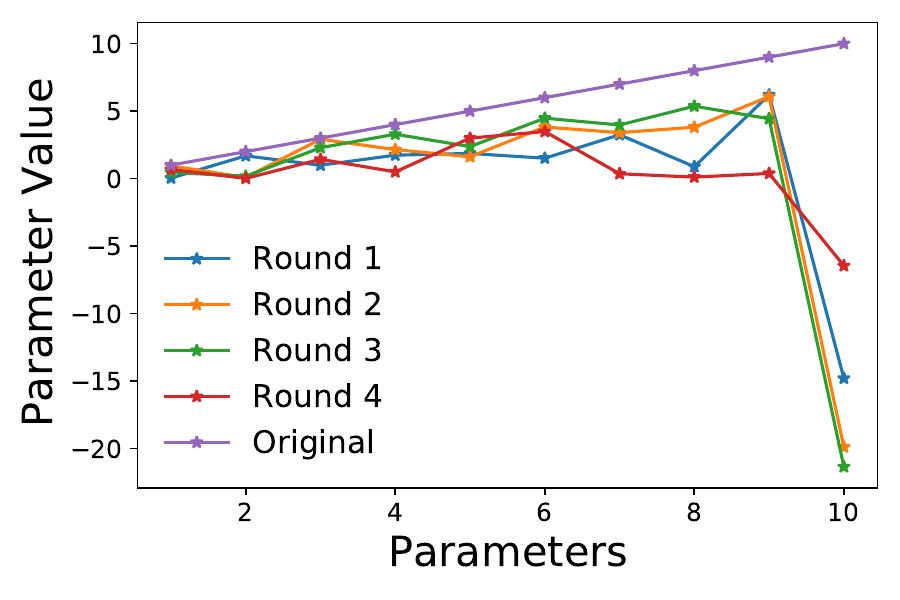}}
\subfigure[Cosine similarity with $\theta=0.5$.]{
\includegraphics[width=0.235\textwidth]{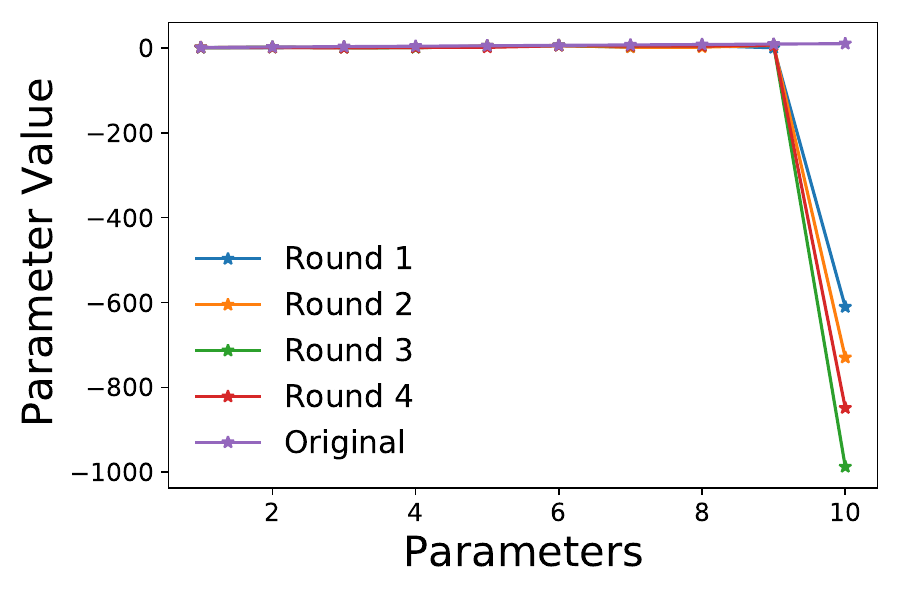}}
\subfigure[Cosine similarity with $\theta=0.9$.]{
\includegraphics[width=0.235\textwidth]{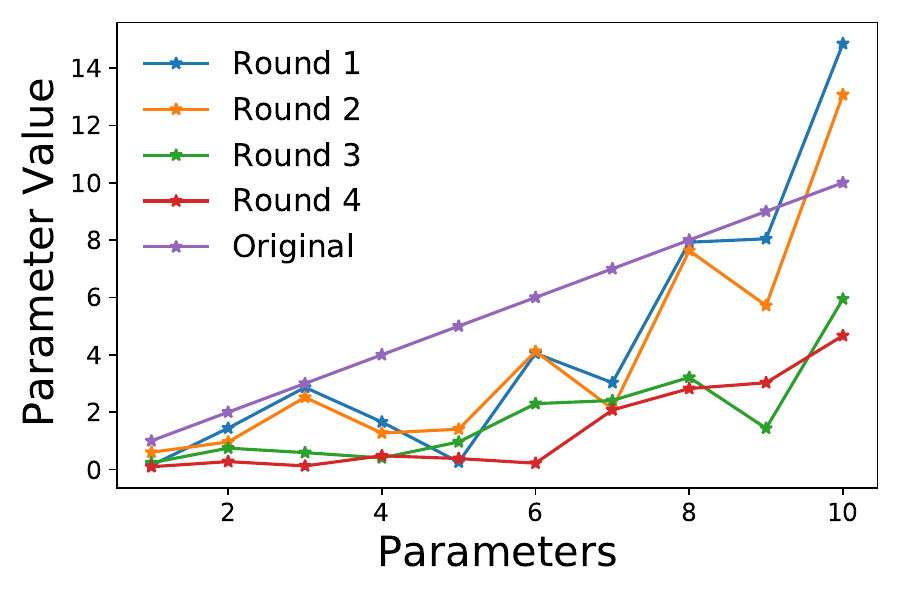}}
\caption{Evaluation of EvilTwin with $J=10$ and $T=10$.}
\label{fig_eviltwin}
\end{figure*}

Next, we present the forged vector generated by EvilTwin. We set $J=10$ and $T=10$ and use different values of $\theta$. The experimental results are shown in Fig. \ref{fig_eviltwin}, and we can see that the distribution of the forged vector is influenced by $\theta$. Specifically, in the $L_2$ norm, if $\theta$ is larger, all parameters except the last parameter will gradually close to 0. In Euclidean distance, a larger $\theta$ will increase the uneven distribution of the data. As for cosine similarity, the distributions of generated vectors are related to the positive and negative of $\theta$, but the experimental results do not show a general distribution pattern.
In summary, the fake vector generated by EvilTwin differs a lot in different rounds; so, for mechanisms sensitive to data values, EvilTwin is a significant threat.

\subsection{Use Case of EvilTwin}\label{case_evil}
In this part, we provide a simple use case to help researchers and developers understand the workflow of EvilTwin. Let $X=(1,2,3,4,5,6,7,8,9,10)$ and $Y=(11,12,13,14,15,16,17,18,19,20)$, and we can get $\frac{L(X)}{L(Y)}\approx 0.3936$; our goal is to obtain a poisoned vector $\overline{X}$ such that $\frac{L(\overline{X})}{L(Y)}= 0.3936$, and the similarity requirement $\theta=0.3936*L(Y)\approx 19.62$ (\textbf{Line} 1). According to \textbf{Algorithm} \ref{al_ev}, we first partition $Y$ into $T=2$ parts, e.g., $Y_1=(11,12,13,14,15)$ and $Y_2=(16,17,18,19,20)$ with the sizes $J_1=J_2=5$ (\textbf{Line} 2). Then, we get $v_1=11^2+12^2+13^2+14^2+15^2=885$ and $v_2=16^2+17^2+18^2+19^2+20^2=1630$ (\textbf{Line} 4), and $\overline{v}_1=\alpha_1v_1=855\alpha_1$ and $\overline{v}_2=\alpha_2 v_2 = 1630\alpha_2$ (\textbf{Line} 5). Thus, we get $V=(855,1630)$ and $\overline{V}=(855\alpha_1, 1630\alpha_2)$ (\textbf{Line} 6). Next, we rewrite $S(\overline{X},Y)=19.62$ as $S(\overline{V},V)=19.62$ with $\alpha_1, \alpha_2, v_1$, and $v_2$, i.e., $[\sum_{t=1}^2 ({\alpha_t}-1)^2 v_t]^\frac{1}{2}=19.62$. Following \textbf{Theorem} \ref{th_l}, we let $\alpha_1$ to be chosen from the domain $|\alpha_{-t}|\leq [\frac{\theta^2}{(T-1)\max(v_{-t})}]^\frac{1}{2}$, i.e., $[-0.6709,0.6709]$. We let $\alpha_1=0.0017$ and we can get $\alpha_2$ by following the equation $\alpha_{t}= \bigg(\frac{1}{v_{t}}\bigg[\sum_{-t=1}^{T-1} (-\alpha^2_{-t} v_{-t})+{\theta}^2\bigg]\bigg)^\frac{1}{2}$, so $\alpha_2\approx0.0263$ (\textbf{Line} 8). Last, the poisoned vector $\overline{X}=(0.6709*11,0.6709*12,0.6709*13,0.6709*14,0.6709*15, 0.0263*16,0.0263*17,0.0263*18,0.0263*19,0.0263*20)=(7.37,8.04,8.71,9.38,10.05,0.42,0.45,0.47,0.50,0.53)$, which is totally different from vector $X$ (\textbf{Lines} 9-10).
\fi
\if()
\section{Dive into I\&B}\label{dib}
\subsection{Novelty Declaration}
\fi


\section{Proofs of Theorems}\label{proofs}

\subsection{Proof of Theorem 1}\label{proof_alpha}
\begin{proof}

The equation $\overline{s}_i=S(\overline{w}_i,w_i)$ can be expressed as $S(\alpha_i\otimes w_i,w_i)-\overline{s}_i=0$. The most straightforward solution of $\alpha_i$ is to allow all the scalars to have the value 1. In this situation, $\overline{w}_i$ is exactly the same as $w_i$. During the attack, such a poisoned local model $\overline{w}_i$ is not effective, thus we should ensure that not all the values in $\alpha_i$ have the value 1. Besides, in a $J$-dimensional space, since the functions of $L_2$ norm, Euclidean distance, and cosine similarity are quadratic, there are multiple combinations of scalars in $\alpha_i$ as the solutions of them, and the attacker only needs one combination that satisfies all the scalars are positive and at least one scalar is not 1.
\label{the_1}
\end{proof}

\if()
\begin{proof}
Step 1: We first prove the calculation of $\xi_j$ for weight-based AGM.
The weight-based AGM can be expressed as $W=\sum_{i=1}^n g_i w_i$ when there is no attack, while if local clients launch the MPA, it is written as $\overline{W}=\sum_{i=1}^m \overline{g}_i \overline{w}_i +\sum_{i=1}^{n-m} g_i w_i$. Since $W_j=\sum_{i=1}^n g_i w_{i,j}$, we can approximately express it as $W_j \approx n g_i w_{i,j}$\footnote{In the rest of the proof, we have used the approximate expression several times, because we believe that it can simplify the computation significantly and can help us to obtain more concise expressions.}. Similarly, We have $\overline{W}_j=\sum_{i=1}^m \overline{g}_i \overline{w}_{i,j} +\sum_{i=1}^{n-m} g_i w_{i,j}$, we can approximately express it as 
\begin{align}
    \overline{W}_j\approx m\overline{g}_i \overline{w}_{i,j} + (n-m)g_i w_{i,j}.\notag
\end{align}

Since $\xi_j=|W_j-\overline{W}_j|$, we can have 
\begin{align}
    \xi_j &\approx |n g_i w_{i,j}-(m\overline{g}_i \overline{w}_{i,j} + (n-m)g_i w_{i,j})|\notag\\
    &=|n g_i w_{i,j}-m\overline{g}_i \overline{w}_{i,j}-n g_i w_{i,j}+mg_i w_{i,j}|\notag \\
    &=|m(g_i w_{i,j}-\overline{g}_i \overline{w}_{i,j})|.\notag
\end{align}

We have $\overline{w}_{i,j}=\alpha_{i,j} w_{i,j}$, so we can get:
\begin{align}
    \xi_j \approx |m(g_i w_{i,j}-\overline{g}_i \alpha_{i,j} w_{i,j})|=|mw_{i,j}(g_i-\overline{g}_i \alpha_{i,j})|.\notag
\end{align}

Because $m w_{i,j}$ is a constant, we have $\xi_j \approx |g_i-\overline{g}_i \alpha_{i,j}|$.

Step 2: Then we prove the calculation of $\xi_j$ for statistics-based AGM as below.
Assume $w_i$ will be selected as the global model.
Since the statistics-based AGM selects one local model as the global model, we have $W=w_i$ if there are no MPAs; otherwise, $\overline{W}=\overline{w}_i$. We have $\xi_j=|W_j-\overline{W}_j|$ and $\overline{w}_{i,j}=\alpha_{i,j}w_{i,j}$, we can get
\begin{align}
    \xi_j\approx|w_{i,j}-\overline{w}_{i,j}|=|w_{i,j}-\alpha_{i,j}w_{i,j}|.\notag
\end{align}

Since $w_{i,j}$ is a constant, we can have $\xi_j \approx |1-\alpha_{i,j}|$. 


\end{proof}

\subsection{Proof of Theorem 2}

\begin{proof}

By following the idea of EvilTwin, we first have to partition the vector $Y$ into $T$ groups, and the square sum group $t$ is $v_t=\sum_{y_{t,j}\in Y_t}y_{t,j}^2$. Since $\overline{v}=\alpha_tv_t$, we can have 
\begin{align}
    L(\overline{X})=\bigg[\sum_{j=1}^{J} \overline{x}_{j}^2\bigg]^\frac{1}{2}=\bigg[\sum_{t=1}^{T} \alpha^2_{t} v_{t}\bigg]^\frac{1}{2}=\theta
\end{align}

 We can rewrite the above equation as 
 \begin{align}
     L(\overline{X})=\alpha^2_{t} v_t + \sum_{t=1}^{T-1} \alpha^2_{-t} v_{-t}=\theta^2.\notag
 \end{align}
Solving it yields 
\begin{align}
    \alpha_t=\pm \bigg(\frac{1}{v_{t}}\bigg[\sum_{t=1}^{T-1} (-\alpha^2_{-t} v_{-t})+{\theta}^2\bigg]\bigg)^\frac{1}{2}.\notag
\end{align}
But we have to ensure that there will be a solution, i.e., $\sum_{t=1}^{T-1} (-\alpha^2_{-t} v_{-t})+{\theta}^2\geq 0$. We have ${\theta}^2\geq\sum_{t=1}^{T-1} (\alpha^2_{-t} v_{-t})$. Since
\begin{align}
    \sum_{t=1}^{T-1} \alpha^2_{-t} v_{-t}\leq (T-1)\alpha^2_{-t}\max(v_{-t}),\notag
\end{align}
we only need to ensure that
\begin{align}
    \theta^2\geq(T-1)\alpha^2_{-t}\max(v_{-t}).\notag
\end{align}
Then, we can get 
 \begin{align}
     |\alpha_{-t}|\leq \bigg[\frac{\theta^2}{(T-1)\max(v_{-t})}\bigg]^\frac{1}{2}.\notag
 \end{align}
 
\end{proof}

\subsection{Proof of Theorem 3}

\begin{proof}

We have 
\begin{align}
    E(\overline{X},Y)&=\bigg[\sum_{j=1}^J(\overline{x}_j-x_j)^2\bigg]^\frac{1}{2}\notag\\
    &=\bigg[\sum_{t=1}^T (\alpha^2_{t}-2\alpha_t+1) v_t\bigg]^\frac{1}{2}=\theta,\notag
\end{align}
and it can be rewritten as 
\begin{align}
    \theta=\bigg[(\alpha^2_{t}-2\alpha_t+1) v_t+(\sum_{-t=1}^{T-1} (\alpha^2_{-t}-2\alpha_{-t}+1) v_{-t})\bigg]^\frac{1}{2}.\notag
\end{align}
Solving it yields
\begin{align}
    \alpha_{t}&= \frac{1}{v_{t}}\bigg[v_{t}\pm \big[v_{t}(\sum_{-t=1}^{T-1}((2\alpha_{-t}\notag-\alpha^2_{-t}-1)v_{-t})+{\theta}^2)\big]^\frac{1}{2}\bigg].\notag
\end{align}
We let 
\begin{align}
    \sum_{-t=1}^{T-1}((2\alpha_{-t}\notag-\alpha^2_{-t}-1)v_{-t})+{\theta}^2\geq 0,
\end{align}
and we have 
\begin{align}
    (T-1)(\alpha_{-t}-\alpha^2_{-t}-1)\max (v_{-t})+\theta^2\geq 0.\notag
\end{align}
Then, we can get 
\begin{align}
    1-\frac{\theta}{[(T-1)\max(v_{-t})]^\frac{1}{2}}\leq\alpha_{-t}\leq 1+\frac{\theta}{[(T-1)\max(v_{-t})]^\frac{1}{2}},\notag
\end{align}
\end{proof}

\subsection{Proof of Theorem 4}
\begin{proof}

Based on EvilTwin, we have 
\begin{align}
    C(\overline{X},Y)&=\frac{\sum_{j=1}^J \overline{x}_j x_j}{(\sum_{j=1}^J \overline{x}_j^2)^\frac{1}{2}(\sum_{j=1}^J x_j^2)^\frac{1}{2}}\notag\\
    &=\frac{\sum_{t=1}^T\alpha_t v_t}{(\sum_{t=1}^T v_t)^\frac{1}{2} (\sum_{t=1}^T \alpha^2_{t} v_t)^\frac{1}{2}}=\theta.\notag
\end{align}
We can rewrite it as
\begin{align}
    \theta=\frac{\sum_{-t=1}^{T-1}\alpha_{-t} v_{-t}+\alpha_t v_t}{(\sum_{-t=1}^T v_{-t}+v_t)^\frac{1}{2} (\sum_{-t=1}^T \alpha^2_{-t} v_{-t}+\alpha^2_{t} v_t)^\frac{1}{2}},\notag
\end{align}
and we can get Equation $\ref{e_cs}$ directly by solving it. 

\label{the_4}
\end{proof}
\fi

\subsection{Proof of Theorem 2}
\begin{proof}

By analyzing $\textbf{P1}$, we know it is a non-linear programming problem with inequality constraints. We propose an approximation method to solve it. Specifically, we need to guarantee that $C(\overline{w}_i,w_i)>0$ so that the poisoned local model will not be discarded by FLTrust. In this way, we have
\begin{align}
    \overline{s}_i=\frac{\sum_{j=1}^J\alpha_{i,j} w^2_{i,j}}{(\sum_{j=1}^J w^2_{i,j})^\frac{1}{2} (\sum_{j=1}^J \alpha_{i,j}^2 w^2_{i,j})^\frac{1}{2}}>0,\notag
\end{align}
and it equivalents to allow $\sum_{j=1}^J\alpha_{i,j} w_{i,j}>0$. Solving it yields $\alpha_{i,j}>-\frac{\sum_{-j=1}^{J-1}\alpha_{i,-j} w^2_{i,-j}}{w^2_{i,j}}$, which is easy to be satisfied since we require the scalars to be positive. We can get the approximate optimal solution by analyzing the objective function. Then, we have to maximize$f(\alpha_i)= \frac{\sum_{j=1}^J w^2_{i,j} \alpha_{i,j} \sum_{j=1}^J\alpha_{i,j}}{\sum_{j=1}^J w^2_{i,j}\alpha_{i,j}^2}$. For simplicity, we can rewrite it as
\begin{align}
    f(\alpha_i)=\frac{(\lambda+w^2_{i,j}\alpha_{i,j})(\gamma +\alpha_{i,j}) }{\beta + w^2_{i,j}\alpha_{i,j}^2},\notag
\end{align}
where $\lambda=\sum_{-j=1}^{J-1} w^2_{i,-j} \alpha_{i,-j}$, $\gamma=\sum_{-j=1}^{J-1}\alpha_{i,-j}$, and $\beta=\sum_{-j=1}^{J-1} w^2_{i,-j} \alpha_{i,-j}^2$.  The first-order derivative of $f(\alpha_i)$ is 
\begin{align}
    \frac{\partial f(\alpha_i)}{\partial \alpha_{i,j}}&=\frac{\lambda(\beta-\alpha_{i,j}w^2_{i,j}(\alpha_{i,j}+2\gamma))}{(\alpha_{i,j}^2w^2_{i,j}+\beta)^2}\notag\\
    &+\frac{\lambda(w^2_{i,j}(2\alpha_{i,j}\beta-\alpha_{i,j}\alpha_{i,j}^2+\beta\gamma))}{(\alpha_{i,j}^2w^2_{i,j}+\beta)^2},\notag
\end{align}
and the second-order derivative of $f(\alpha_i)$ is
\begin{align}
    \frac{\partial ^2 f(\alpha_i) }{\partial ^2 \alpha_{i,j}}&=\frac{2w^2_{i,j}(\alpha_{i,j}^3w^2_{i,j}(\lambda+\gamma w^2_{i,j})+\beta(\beta-\lambda\gamma))}{(\alpha_{i,j}^2w^2_{i,j}+\beta)^3}\notag\\
    &-\frac{6w^2_{i,j}(\alpha_{i,j}^2w^2_{i,j}(\beta-\lambda\gamma)+\alpha_{i,j}\beta(\lambda+\gamma w^2_{i,j}))}{(\alpha_{i,j}^2 w^2_{i,t}+\beta)^3}.\notag
\end{align}

We get $\frac{\partial ^2 f(\alpha_i) }{\partial ^2 \alpha_{i,j}}<0$, thus $f(\alpha_i)$ is concave. Let $\frac{\partial f(\alpha_i)}{\partial \alpha_{i,j}}=0$, and solving it yields the equation in \textbf{Theorem 1}, which is the optimal value of $\alpha_{i,j}$.

\label{the_2}
\end{proof}

\subsection{Proof of Theorem 3}
\begin{proof}

We have 
\begin{align}
    E(\overline{w}_i,w_i)&=[\sum_{j=1}^J(\overline{w}_{i,j}-w_{i,j})^2]^\frac{1}{2}\notag  \\ 
    &=[\sum_{j=1}^J (\alpha^2_{i,j}-2\alpha^2_{i,j}+1) w^2_{i,j}]^\frac{1}{2}<E(w_g,w_i),\notag
    \nonumber
\end{align}
and it can be rewritten as 
\begin{align}
    E(w_g,w_i)&=[(\alpha^2_{i,j}-2\alpha_{i,j}+1) w^2_{i,j}\notag\\
    &+(\sum_{-j=1}^{J-1} (\alpha^2_{i,-j}-2\alpha_{i,-j}+1) w^2_{i,-j})]^\frac{1}{2}.\notag
\end{align}
Solving it yields
\begin{align}
   0< \alpha_{i,j}&< \frac{1}{w^2_{i,j}}[w_{i,j}+ [w^2_{i,j}(\sum_{-j=1}^{J-1}((2\alpha_{i,-j}-\alpha^2_{i,-j}-1)w^2_{i,-j})\notag\\
    &+{E(w_g,w_i)}^2)]^\frac{1}{2}].\notag
\end{align}

To ensure that $\alpha_{i,j}$ is positive, we let 
\begin{align}
    \sum_{-j=1}^{J-1}((2\alpha_{i,-j}-\alpha^2_{i,-j}-1)w^2_{i,-j})+{E(w_g,w_i)}^2>0,\notag
\end{align}
and we have 
\begin{align}
    (J-1)(\alpha_{i,-j}-\alpha^2_{i,-j}-1)\max (w^2_{i,-j})+E(w_g,w_i)^2> 0.\notag
\end{align}
Then, we can get 
\begin{align}
    0<\alpha_{i,-j}< 1+\frac{E(w_g,w_i)}{[(J-1)\max(w_{i,-j})]^\frac{1}{2}}.\notag
\end{align}
\label{the_3}
\end{proof}

\subsection{Proof of Theorem 4}
\begin{proof}
$f(\alpha_i)$ is a monotonically increasing function when $\alpha_{i,j}> 0$, we can get its lower and upper bounds by solving $C1$.\begin{align}
    L(\overline{w}_i)=[\sum_{j=1}^{J} \overline{w}_{i,j}^2]^\frac{1}{2}=[\sum_{j=1}^{J} \alpha^2_{i,j} w^2_{i,j}]^\frac{1}{2}\leq L(w_i).\notag
\end{align}

 We can rewrite the above equation as 
 \begin{align}
     L(\overline{w}_i)=\alpha^2_{i,j} \overline{w}_{i,j}^2 + \sum_{j=1}^{J-1} \alpha^2_{i,-j} \overline{w}_{i,-j}^2=L(w_i)^2.\notag
 \end{align}
 
Solving it yields 
\begin{align}
       \alpha_{i,j}^*= (\frac{1}{w^2_{i,j}}[\sum_{-j=1}^{J-1} (-\alpha_{i,-j}^2 w^2_{i,-j})+L(w_i)^2])^\frac{1}{2}.\notag
\end{align}

But we have to ensure that there will be a solution, i.e., $\sum_{-j=1}^{J-1} (-\alpha_{i,-j}^2 w^2_{i,-j})+L(w_i)^2\geq 0$. We have $L(w_i)^2\geq\sum_{j=1}^{J-1} (\alpha^2_{i,-j} w^2_{i,-j})$. Since
\begin{align}
    \sum_{j=1}^{J-1} \alpha^2_{i,-j} w^2_{-j}\leq (J-1)\alpha^2_{i,-j}\max(w_{i,-j}),\notag
\end{align}
we only need to ensure that
\begin{align}
    L(w_i)^2\geq(J-1)\alpha^2_{i,-j}\max(w^2_{i,-j}).\notag
\end{align}
Then, we can get 
 \begin{align}
     0<\alpha_{i,-j}\leq [\frac{L(w_i)^2}{(J-1)\max(w^2_{i,-j})}]^\frac{1}{2}.\notag
 \end{align}
 
\label{the_4}
\end{proof}

\if()
\subsection*{Proof of Theorem 1}
\begin{proof}
By following the idea of EvilTwin, we have 
\begin{align}
    L(X)=\bigg(\sum_{t=1}^{T} \sigma_{t}^2 v_{t}\bigg)^\frac{1}{2}=\theta.\notag
\end{align}
 We can rewrite the above equation as 
 \begin{align}
     L(X)=\sigma_t^2 v_t + \sum_{t=1}^{T-1} \sigma_{-t}^2 v_{-t}=\theta.\notag
 \end{align}
Solving it yields 
\begin{align}
    \sigma_t=\pm \frac{1}{v_{t}}\bigg[\sum_{t=1}^{T-1} (-\sigma_{-t}^2 v_{-t})+{\theta}^2\bigg]^\frac{1}{2}.\notag
\end{align}
But we have to ensure that there will be a solution, i.e., $\sum_{t=1}^{T-1} (-\sigma_{-t}^2 v_{-t})+{\theta}^2\geq 0$. We have 
\begin{align}
    \sum_{t=1}^{T-1} -\sigma_{-t}^2 v_{-t}\leq (T-1)\sigma_{-t}^2\max(v_{-t}),\notag
\end{align}
so we can have
\begin{align}
    \theta^2\geq(T-1)\sigma_{-t}^2\max(v_{-t}).\notag
\end{align}
Then, we can get 
 \begin{align}
     |\sigma_{-t}|\leq \bigg[\frac{\theta^2}{(T-1)\max(v_{-t})}\bigg]^\frac{1}{2}.\notag
 \end{align}


\end{proof}

\subsection*{Proof of Theorem 2}
\begin{proof}

We have 
\begin{align}
    E(X)=\bigg[\sum_{t=1}^T (\sigma_t^2-2\sigma_t+1) v_t\bigg]^\frac{1}{2}=\theta,\notag
\end{align}
and it can be rewritten as 
\begin{align}
    \theta=\bigg[(\sigma_t^2-2\sigma_t+1) v_t+(\sum_{t=1}^{T-1} (\sigma_{-t}^2-2\sigma_{-t}+1) v_{-t})\bigg]^\frac{1}{2}.\notag
\end{align}
Solving it yields
\begin{align}
    \sigma_{t}&= \frac{1}{v_{t}}\bigg[v_{t}\pm \big[v_{t}(\sum_{t=1}^{T-1}((2\sigma_{-t}\notag-\sigma_{-t}^2-1)v_{-t})+{\theta}^2)\big]^\frac{1}{2}\bigg].\notag
\end{align}
We let 
\begin{align}
    \sum_{t=1}^{T-1}((2\sigma_{-t}\notag-\sigma_{-t}^2-1)v_{-t})+{\theta}^2\geq 0,
\end{align}
and we have 
\begin{align}
    (T-1)(\sigma_{-t}-\sigma_{-t}^2-1)\max (v_{-t})+\theta^2\geq 0.\notag
\end{align}
Then, we can get 
\begin{align}
    1-\frac{\theta}{[(T-1)\max(v_{-t})]^\frac{1}{2}}\leq\sigma_{-t}\leq 1+\frac{\theta}{[(T-1)\max(v_{-t})]^\frac{1}{2}},\notag
\end{align}
\end{proof}

\subsection*{Proof of Theorem 3}
\begin{proof}

Based on EvilTwin, we have 
\begin{align}
    C(X)=\frac{\sum_{t=1}^T\sigma_t v_t}{(\sum_{t=1}^T v_t)^\frac{1}{2} (\sum_{t=1}^T \sigma_t^2 v_t)^\frac{1}{2}}=\theta.\notag
\end{align}
We can rewrite it as
\begin{align}
    \theta=\frac{\sum_{t=1}^{T-1}\sigma_{-t} v_{-t}+\sigma_t v_t}{(\sum_{t=1}^T v_{-t}+v_t)^\frac{1}{2} (\sum_{t=1}^T \sigma_{-t}^2 v_{-t}+\sigma_t^2 v_t)^\frac{1}{2}},\notag
\end{align}
and we can get equation $\ref{e_cs}$ directly by solving it. 


\end{proof}

\subsection*{Proof of Theorem 4}
\begin{proof}
We first prove the calculation of $\xi_j$ for weight-based AGM.

The weight-based AGM can be expressed as $\bm{\omega}=\sum_{i=1}^n g_i w_i$ when there is no attack, while if local clients launch the MPA, it is written as $\overline{\bm{\omega}}=\sum_{i=1}^m \overline{g}_i \overline{w}_i +\sum_{i=1}^{n-m} g_i w_i$. Since $\bm{\omega}_j=\sum_{i=1}^n g_i w_{i,j}$, we can approximately express it as $\bm{\omega}_j \approx n g_i w_{i,j}$. Similarly, We have $\overline{\bm{\omega}}_j=\sum_{i=1}^m \overline{g}_i \overline{w}_{i,j} +\sum_{i=1}^{n-m} g_i w_{i,j}$, we can approximately express it as 
\begin{align}
    \overline{\bm{\omega}}_j\approx m\overline{g}_i \overline{w}_{i,j} + (n-m)g_i w_{i,j}.\notag
\end{align}

Since $\xi_j=|\bm{\omega}_j-\overline{\bm{\omega}}_j|$, we can have 
\begin{align}
    \xi_j &\approx |n g_i w_{i,j}-(m\overline{g}_i \overline{w}_{i,j} + (n-m)g_i w_{i,j})|\notag\\
    &=|n g_i w_{i,j}-m\overline{g}_i \overline{w}_{i,j}-n g_i w_{i,j}+mg_i w_{i,j}|\notag \\
    &=|m(g_i w_{i,j}-\overline{g}_i \overline{w}_{i,j})|.\notag
\end{align}

We have $\overline{w}_{i,j}=\alpha_{i,j} w_{i,j}$, so we can get:
\begin{align}
    \xi_j \approx |m(g_i w_{i,j}-\overline{g}_i \alpha_{i,j} w_{i,j})|=|mw_{i,j}(g_i-\overline{g}_i \alpha_{i,j})|.\notag
\end{align}

Because $m w_{i,j}$ is a constant, we have $\xi_j \approx |g_i-\overline{g}_i \alpha_{i,j}|$.

Then we prove the calculation of $\xi_j$ for statistics-based AGM as below.

Assume $w_i$ will be selected as the global model.
Since the statistics-based AGM selects one local model as the global model, we have $\bm{\omega}=w_i$ if there is no MPA attack; otherwise, $\overline{\bm{\omega}}=\overline{w}_i$. We have $\xi_j=|\bm{\omega}_j-\overline{\bm{\omega}}_j|$ and $\overline{w}_{i,j}=\alpha_{i,j}w_{i,j}$, we can get
\begin{align}
    \xi_j\approx|w_{i,j}-\overline{w}_{i,j}|=|w_{i,j}-\alpha_{i,j}w_{i,j}|.\notag
\end{align}

Since $w_{i,j}$ is a constant, we can have $\xi_j \approx |1-\alpha_{i,j}|$. 


\end{proof}

\subsection*{Proof of Theorem 5}
\begin{proof}

By analyzing $\textbf{P1}$, we can see it is a non-linear programming problem with inequality constraints. We can try to consider using the Lagrange method to solve it, however, this is inefficient. Thus, we propose an approximation method. Specifically, we only need to guarantee that $C(\overline{w}_i,w_i)>0$ so that the poisoned local model will not be discarded by FLTrust; based on EvilTwin, we know that we can control the value of similarity by adjusting the value of $\alpha_{i,t}$. In this way, we have
\begin{align}
    \theta=\frac{\sum_{t=1}^T\alpha_t \psi_t}{(\sum_{t=1}^T \psi_t)^\frac{1}{2} (\sum_{t=1}^T (\alpha_t^2 \psi_t)^\frac{1}{2}}>0,\notag
\end{align}
and we only need to ensure $\sum_{t=1}^T\alpha_{i,t} \psi_{i,t}>0$. Solving it yields $\alpha_{i,t}>-\frac{\sum_{t=1}^{T-1}\alpha_{i,-t} \psi_{i,-t}}{\psi_{i,t}}$. Thus, we can get the approximate optimal solution by analyzing the objective function. Maximizing $\xi$ equivalents to approximately maximizing $ \bigg|-\frac{\sum_{t=1}^T \psi_{i,t} \alpha_{i,t} \sum_{t=1}^T\alpha_{i,t}}{\sum_{t=1}^T \psi_{i,t}\alpha_{i,t}^2}\bigg|$ since $g_i$ can be treated as a constant. This is a function with absolute value, so it is non-negative, and we can just analyze $\frac{\sum_{t=1}^T \psi_{i,t} \alpha_{i,t} \sum_{t=1}^T\alpha_{i,t}}{\sum_{t=1}^T \psi_{i,t}\alpha_{i,t}^2}$ first. For simplicity, we can rewrite it as
\begin{align}
    \xi'=\frac{(\beta+\psi_{i,t}\alpha_{i,t})(\gamma +\alpha_{i,t}) }{\beta + \psi_{i,t}\alpha_{i,t}^2},\notag
\end{align}
where $\beta=\sum_{t=1}^{T-1} \psi_{i,-t} \alpha_{i,-t}$, $\gamma=\sum_{t=1}^{T-1}\alpha_{i,-t}$, and $\beta=\sum_{t=1}^{T-1} \psi_{i,-t} \alpha_{i,-t}^2$.  The first-order derivative of $\xi'$ is 
\begin{align}
    \frac{\partial \xi'}{\partial \alpha_{i,t}}&=\frac{\beta(\beta-\alpha_{i,t}\psi_{i,t}(\alpha_{i,t}+2\gamma))}{(\alpha_{i,t}^2\psi+\beta)^2}\notag\\
    &+\frac{\beta(\psi_{i,t}(2\alpha_{i,t}\beta-\alpha_{i,t}\alpha_{i,t}^2+\beta\gamma))}{(\alpha_{i,t}^2\psi+\beta)^2},\notag
\end{align}
and the second-order derivative of $\xi'$ is

\begin{align}
    \frac{\partial ^2 \xi' }{\partial ^2 \alpha_{i,t}}&=\frac{2\psi_{i,t}(\alpha_{i,t}^3\psi_{i,t}(\beta+\gamma\psi_{i,t})+\beta(\beta-\beta\gamma))}{(\alpha_{i,t}^2\psi_{i,t}+\beta)^3}\notag\\
    &+\frac{-6\psi_{i,t}((\alpha_{i,t}^2\psi_{i,t}(\beta-\beta\gamma)+\alpha_{i,t}\beta(\beta+\gamma\psi_{i,t})))}{(\alpha_{i,t}^2\psi_{i,t}+\beta)^3}.\notag
\end{align}

It is hard to distinguish whether $\frac{\partial ^2 \xi' }{\partial ^2 \alpha_{i,t}}$ is positive or negative. But since we only need to consider the absolute value of $\xi'$, we can know that $\xi'$ has at least one extreme point. Let $\frac{\partial \xi'}{\partial \alpha_{i,t}}=0$, solving it yields equation (\ref{faker_fltrust}).


\end{proof}
\subsection*{Proof of Theorem 6}
\begin{proof}
Since the objective function  of $\textbf{P2}$ decreases monotonically when $\alpha_{i,t}<1$ and increases monotonically when $\alpha_{i,t}>1$, we only need to calculate the lower and upper bounds of it. We can directly get the bounds by solving $C1$, and the result is shown as (\ref{faker_krum}). 


\end{proof}

\subsection*{Proof of Theorem 7}
\begin{proof}
$\xi$ is a monotonically decreasing function when $\alpha_{i,t}<g_i$ and is a monotonically increasing function when $\alpha_{i,t}>g_i$, we can get its lower and upper bounds by solving $C1$, and we get
\begin{align}
   \alpha_{i,t}\geq -\frac{1}{\psi_{i,t}}[\sum_{t=1}^{T-1} (-\alpha_{i,-t}^2 \psi_{i,-t})+{\theta}^2]^\frac{1}{2},\notag
\end{align}
and 
\begin{align}
       \alpha_{i,t}\leq \frac{1}{\psi_{i,t}}[\sum_{t=1}^{T-1} (-\alpha_{i,-t}^2 \psi_{i,-t})+{\theta}^2]^\frac{1}{2}.\notag
\end{align}

There could be two optimal points on the lower bound and upper bound, respectively.

\end{proof}

\fi






\section{Dive into Faker}
\subsection{Faker Against Other Benchmark Defenses}\label{faker_other}
In this part, we introduce the designs of Faker against the other three benchmark defenses.

\textbf{Faker Against FLAME.}
FLAME adopts cosine similarity and $L_2$ norm as the evaluation metrics, which is similar to FLTrust. Thus, in Faker, we allow FLAME to share the same optimization objective function with FLTrust. The only difference between Faker in FLAME and FLTrust is the lower bound of $C(\overline{w}_i,w_i)$  since FLAME only accepts local models with higher cosine similarity. We can follow \textbf{Theorem 2} to set the scalars but scale them down to close to 1.

\textbf{Faker Against DiverseFL.} DiverseFL is similar to FLTrust, which applies both $L_2$ norm and cosine similarity to mitigate the negative influence of attacks. In this way, we can use the method of attacking FLTrust to attack DiverseFL.

\textbf{Faker Against ShieldFL.} ShieldFL employs cosine similarity to filter out malicious models, while according to the vulnerabilities of cosine similarity, we can use a larger scalar (usually larger than $n$) to scale up the local models to get the poisoned ones which will not change the measured similarity.

\subsection{Faker-based Other Attacks}\label{faker_other_attacks}
Below we provide the design of the Faker-based backdoor attack and Sybil attack.

\textbf{Faker-based Backdoor Attack.} Faker has to maximize $\overline{w}_i$ and minimize $\delta_i$ to launch the backdoor attack. We allow an attacker to get an initial poisoned model by means of other backdoor attacks, and then we let Faker maintain the parameters in several critical layers such as the output layer, and adjust the similarity according to the requirements of the defender by modifying the parameters of the rest of the layers.

\textbf{Faker-based Sybil Attack.} 
The method by which Faker launches a Sybil attack is as simple as generating $m$ poisoned local models based on $w_i$ or $w_g$ (if $w_i$ is not accessible) and distributing them to Sybil clients.

\if()
\subsection{Basic Faker Attack}\label{basic_faker}

When $T=2$, we call the attack the basic Faker Attack.
\begin{itemize}
    \item Basic Faker Attack Against FLTrust and DiverseFL:
    \begin{align}
        \alpha_{i,t}^*=\pm \alpha_{i,-t}\bigg[\frac{\psi_{i,-t}}{\psi_{i,t}}\bigg]^\frac{1}{2}.
        \notag
    \end{align}
    \item Basic Faker Attack Against Krum:

\begin{align}
    \frac{1}{\psi_{i,t}}(\psi_{i,t}-\Omega)<\alpha_{i,t}<\frac{1}{\psi_{i,t}}(\psi_{i,t}+\Omega)\notag,
\end{align}
where $\Omega=[\psi_{i,t}((2\alpha_{i,-t}\notag-\alpha_{i,-t}^2-1)\psi_{i,-t}+{\theta}^2)]^\frac{1}{2}$.
    
    \item Basic Faker Attack Against Norm-clipping and FLAME:
    \begin{align}
        \alpha_{i,t}^*=\pm \bigg[\frac{1}{\psi_{i,t}}(\theta^2-\alpha_{i,-t}^2\psi_{i,-t})\bigg]^\frac{1}{2}.\notag
    \end{align}
\end{itemize}

\fi

\begin{table*}[ht]
\centering
\caption{Evaluation of success rate of attack with $n=5$ and $m=1$ by running 100 rounds.}
\begin{tabular}{c|ccc|ccc|ccc|ccc} 
\hline
\rowcolor[HTML]{FFCCC9} 
              & \multicolumn{3}{c|}{CIFAR-100} & \multicolumn{3}{c}{HAM10000} & \multicolumn{3}{c|}{Tiny ImageNet}& \multicolumn{3}{c}{Reuters} \\ 
\hline
\rowcolor[HTML]{C0C0C0} 
AGM          & LA & MB & Faker          & LA & MB & Faker  & LA & MB & Faker          & LA & MB & Faker \\ 
\hline
\hline
KM          &   0.76 & 0.89   &      \textbf{1.00}          &  0.75    & 0.63  & \textbf{1.00}  &  1.00  & 1.00  & 1.00  & 0.79 &0.58 & \textbf{1.00}     \\
NC &    0.00   & 0.00   &        \textbf{1.00}       & 0.00    & 0.00   & \textbf{1.00}    &   0.54   & 0.34  & \textbf{1.00}  &0.00 & 0.00 & \textbf{1.00}     \\
FT      &    1.00  & 1.00   &    1.00            &  1.00   & 1.00   &  1.00    &  1.00    & 1.00  & 1.00   &0.32 & 0.25 & \textbf{1.00}      \\
FM         &    0.53  & 0.67 &   \textbf{1.00}             &   0.07  &  0.14  &    \textbf{1.00} &  0.65    & 0.56  & \textbf{1.00}  & 0.02& 0.00 & \textbf{1.00}      \\
DF     &  0.00    &  0.00  &   \textbf{1.00}             &   0.00  & 0.00   &    \textbf{1.00}  & 0.13    &0.09  & \textbf{1.00}   &  0.09& 0.05 & \textbf{1.00}      \\
SF      &   1.00   &  1.00  &      1.00          &    0.00  & 0.00   & \textbf{1.00}    &  1.00   & 1.00 &  1.00   & 1.00   & 1.00     & 1.00   \\
\hline
\end{tabular}
\label{success_ex}
\end{table*}

\begin{table*}[ht]
\centering
\caption{Evaluation of time cost with $n=5$ and $m=1$. The time consumption is measured in seconds.}
\begin{tabular}{c|ccc|ccc|ccc|ccc} 
\hline
\rowcolor[HTML]{FFCCC9} 
              & \multicolumn{3}{c|}{CIFAR-100} & \multicolumn{3}{c}{HAM10000} & \multicolumn{3}{c|}{Tiny ImageNet}& \multicolumn{3}{c}{Reuters}\\ 
\hline
\rowcolor[HTML]{C0C0C0} 
AGM          & LA & MB & Faker          & LA & MB & Faker  & LA & MB & Faker & LA & MB & Faker     \\ 
\hline
\hline
KM          &    75.002    & 74.220   &      \textbf{7.080}          &  78.482    &  73.237 & \textbf{6.897}           & 89.569  &80.345   & \textbf{10.264}    &6.391  & 4.293  & \textbf{1.272}  \\
NC &   0.806   &  0.161  &         \textbf{0.076}       & 0.759    &  0.457  & \textbf{0.087}     & 2.792  &  1.092 & \textbf{0.796}    &  0.495 &  0.583 & \textbf{0.146} \\
FT      &   0.267   & 0.373   &  \textbf{0.102}              & 0.453    &  0.642  & \textbf{0.096}    &  2.352 & 1.675  & \textbf{0.993}    & 0.767  & 0.645  & \textbf{0.212}  \\
FM         & 2.070     & 2.16 &   \textbf{0.424}             &  1.974   & 1.983   &    \textbf{0.394}   &  0.973 & 0.623  & \textbf{0.432}    & 0.482  & 0.428  & \textbf{0.327}\\
DF     &   1.863   & 1.142    &   \textbf{0.427}             & 1.542    &  1.082  &    \textbf{0.362}   & 1.394  & 1.072  & \textbf{0.513}    & 0.394  & 0.341  & \textbf{0.235}\\
SF      &  0.545    & 0.865 &        \textbf{0.223}        &  0.657    & 0.762   &   \textbf{0.132}  & 1.782  &  0.983 & \textbf{0.732}    & 0.282  & 0.327  & \textbf{0.186}\\
\hline
\end{tabular}
\label{time_ex}
\end{table*}

\begin{figure*}[ht]
\centering
\subfigure[CIFAR-100]{
\includegraphics[width=0.23\textwidth]{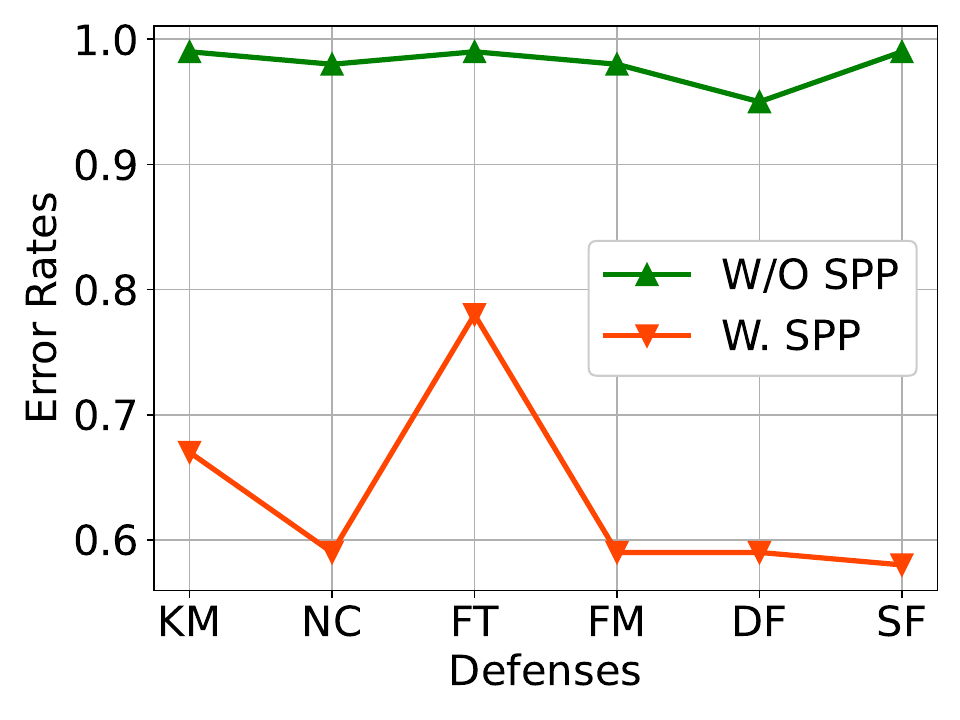}}
\subfigure[HAM10000.]{
\includegraphics[width=0.23\textwidth]{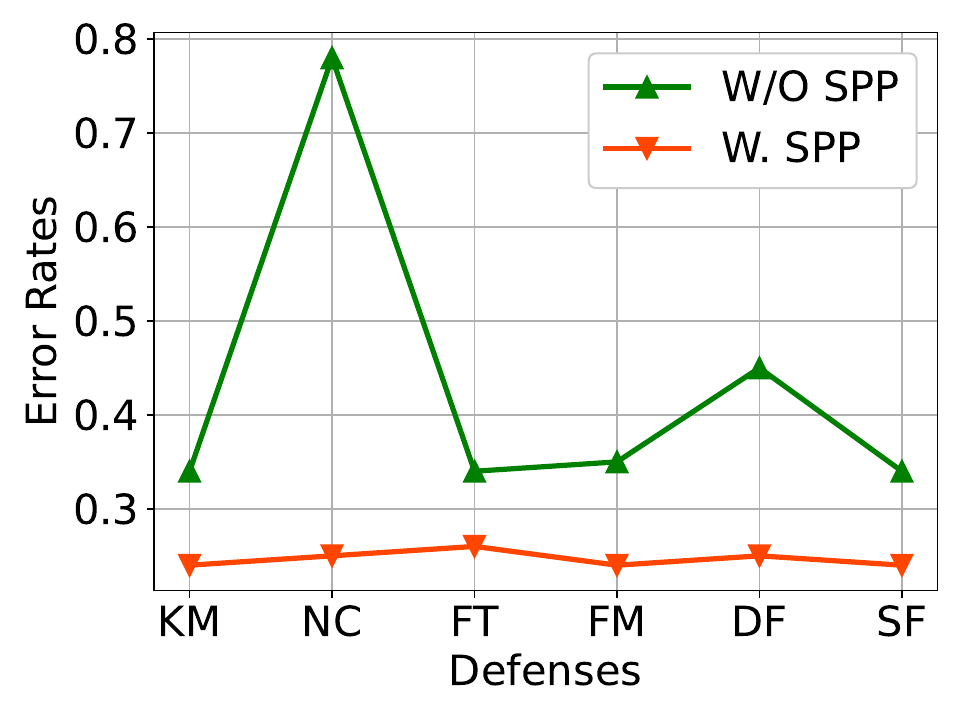}}
\subfigure[Tiny ImageNet.]{
\includegraphics[width=0.23\textwidth]{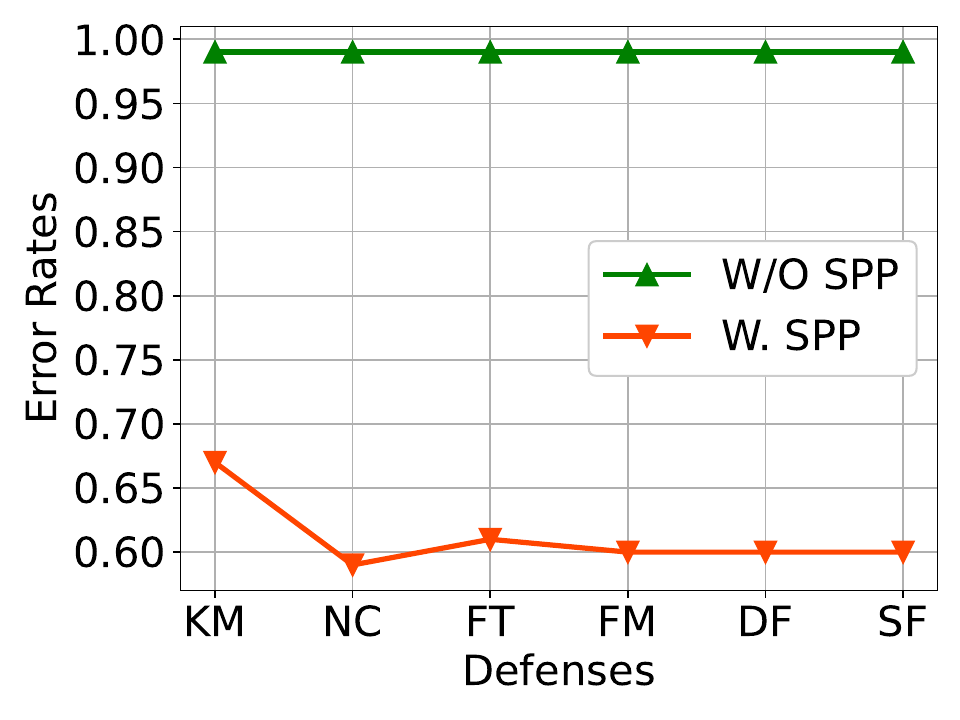}}
\subfigure[Reuters.]{
\includegraphics[width=0.23\textwidth]{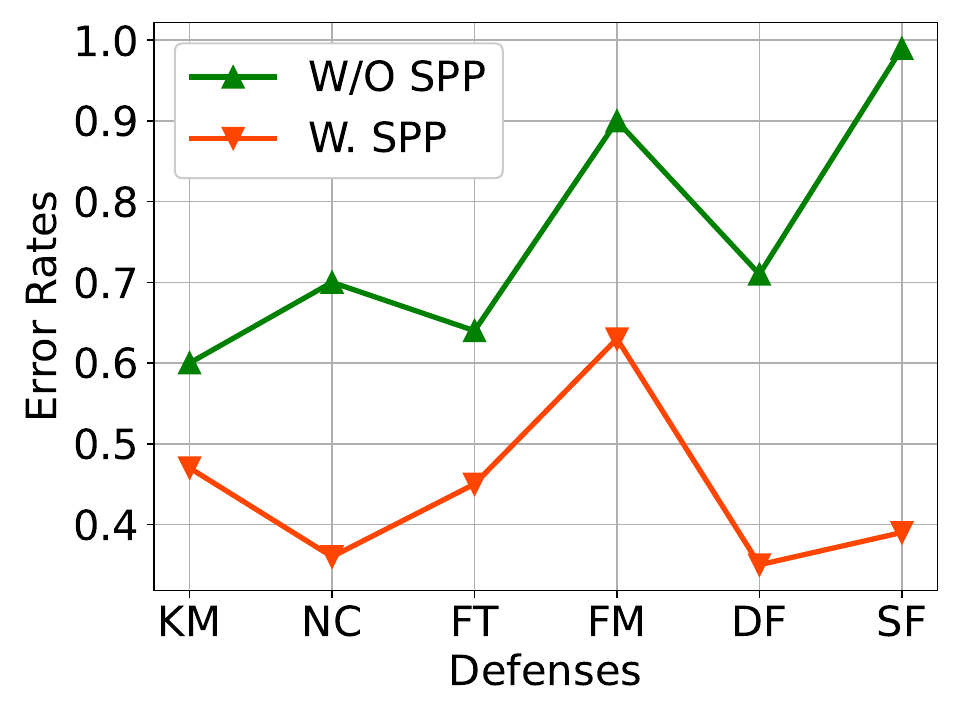}}
\caption{SPP against Faker when $n=5$ and $m=1$. W. SPP means that the defenses are protected by SPP, and W/O SPP means that the defenses are not protected by SPP.}
\label{fig_spp}
\end{figure*}

\subsection{Extra Experiments on Faker}\label{ex_faker}
\if()
\begin{table}[ht]
\centering
\caption{Error rates of IID data with $n=100$ and $m=20$. The error rates of FA for the two datasets are 0.59 and 0.21 when there is no attack. }
\arrayrulecolor{black}
\resizebox{\linewidth}{!}{
\begin{tabular}{cccccccc} 
\hline
\rowcolor[HTML]{FFCCC9}
Dataset                 & Attack & KM & NC & FT & FM & DF & SF \\
\hline
\hline
\multirow{4}{*}{MNIST}  & N/A    & 0.07   & 0.03   & 0.03  &0.03  & 0.03  &  0.03\\
                         & LA    &0.07   & 0.03  & 0.03  & 0.03   &  0.03  & 0.03   \\
                         & MB      & 0.07  & 0.03& 0.03  & 0.03  &  0.03 & 0.03 \\ 
                         & Faker   &\textbf{0.15}    & \textbf{0.18}  &\textbf{0.07}  & \textbf{0.25}  &  \textbf{0.11}  & \textbf{0.30}  \\ 
\arrayrulecolor{black}\cline{1-1}\arrayrulecolor{black}\cline{2-8}
\multirow{4}{*}{FMNIST}  & N/A    &    &    &   &  &   &  \\
                         & LA    &   &   &   &    &    &    \\
                         & MB      &   & &   &   &   &  \\ 
                         & Faker   &    &   &  &   &    &   \\ 
\arrayrulecolor{black}\cline{1-1}\arrayrulecolor{black}\cline{2-8}
\multirow{4}{*}{CIFAR-10}  & N/A    &    &    &   &  &   &  \\
                         & LA    &   &   &   &    &    &    \\
                         & MB      &   & &   &   &   &  \\ 
                         & Faker   &    &   &  &   &    &   \\ 
\arrayrulecolor{black}\cline{1-1}\arrayrulecolor{black}\cline{2-8}

\end{tabular}}
\arrayrulecolor{black}
\label{iid_large}
\end{table}
\fi

\subsubsection{Datasets and Models} 1) HAM10000 is a collection of 10,015 dermatoscopic images of common pigmented skin lesions with seven classes, containing 7000 training images and 3015 test images. This dataset is non-IID, which means that the amount of data in each category is obviously different. We use a 10-layer CNN neural network to categorize it. The details of the model are as follows. There are four Conv2D layers, two MaxPooling2D layers, one Flatten layer, and three Dense layers. 2) 
CIFAR-100 consists of 100 classes with 600 images per class, totaling 60,000 images. The dataset is designed to evaluate object recognition algorithms and covers a wide range of visual concepts, including animals, vehicles, household objects, and natural scenes. Each image is a 32x32 RGB image, making it a challenging dataset for machine learning models to classify objects accurately. An 8-layer CNN model is applied, including two Conv2D layers, two MaxPooling2D layers, one Flatten layer, and three Dense layers. 3) Tiny ImageNet is a downscaled version of the ImageNet dataset, which contains 200 classes, 100,000 training images, and 10,000 test images. All images in Tiny ImageNet are of 64x64 pixels. We apply ResNet-18 to process this dataset. 4) The Reuters dataset is a collection of short news articles that were published by Reuters in the 1980s.  The dataset consists of 10,788 news articles, divided into a training set of 8,982 articles and a test set of 1,806 articles. We use an LSTM model to process this dataset.

\subsubsection{Experimental Results} \label{ex_experiments}
We follow the basic settings as in Section \ref{exp_se} to evaluate Faker and SPP with more experiments. Below are the experimental results.

\textbf{Further Evaluation of Faker on MNIST, FMNIST, and CIFAR-10.}
We take Dirichlet distribution to obtain non-IID training data by setting the concentration parameter as 0.5, and we use LeNet-5 for MNIST, adjusted LeNet for FMNIST, and AlexNet for CIFAR-10. The experimental results are shown in Table \ref{noniid_ex_dirle}, which indicates that Faker can undermine similarity-based defenses with different deep learning models and data distributions and Faker outperforms the benchmark attacks.

\begin{table}[ht]
\centering
\caption{Error rates of non-IID data obtained by Dirichlet distribution with $n=100$ and $m=20$, and the concentration parameter is $0.5$. The error rates of FA for the three datasets are 0.03, 0.18, and 0.43 when there is no attack. }
\arrayrulecolor{black}
\resizebox{\linewidth}{!}{
\begin{tabular}{cccccccc} 
\hline
\rowcolor[HTML]{FFCCC9}
Dataset                 & Attack & KM & NC & FT & FM & DF & SF \\
\hline
\hline
\multirow{4}{*}{MNIST}  & N/A    & 0.07   & 0.03   & 0.03  &0.03  & 0.03  &  0.03\\
                         & LA    &0.07   & 0.03  & 0.03  & 0.03   &  0.03  & 0.03   \\
                         & MB      & 0.07  & 0.03& 0.03  & 0.03  &  0.03 & 0.03 \\ 
                         & Faker   &\textbf{0.15}    & \textbf{0.18}  &\textbf{0.07}  & \textbf{0.25}  &  \textbf{0.11}  & \textbf{0.30}  \\ 
\arrayrulecolor{black}\cline{1-1}\arrayrulecolor{black}\cline{2-8}
\multirow{4}{*}{FMNIST}  & N/A     & 0.38  & 0.18 &  0.17 & 0.18 & 0.18  & 0.17 \\
                         & LA    &  0.38 & 0.18 &  0.22 &  0.18  &   0.23 &  0.22  \\
                         & MB      & 0.38  & 0.18 & 0.21  & 0.18  &  0.18 & 0.19 \\ 
                         & Faker   &  \textbf{0.57}  & \textbf{0.64}  & \textbf{0.51} &\textbf{0.63}   &  \textbf{0.76}  &  \textbf{0.90} \\ 
\arrayrulecolor{black}\cline{1-1}\arrayrulecolor{black}\cline{2-8}
\multirow{4}{*}{CIFAR-10}  & N/A    & 0.86   & 0.43   & 0.44  & 0.43 &  0.42 & 0.77 \\
                         & LA    &0.90   &  0.43 &0.80   & 0.49   & 0.73   & 0.90   \\
                         & MB      & 0.90  & 0.43 & 0.68   & 0.47  &  0.64 &0.90  \\ 
                         & Faker   &  0.90  & \textbf{0.90}  & \textbf{0.90}  & \textbf{0.90}  &\textbf{0.90}    &0.90   \\ 
\arrayrulecolor{black}\cline{1-1}\arrayrulecolor{black}\cline{2-8}

\end{tabular}}
\arrayrulecolor{black}
\label{noniid_ex_dirle}
\end{table}

\begin{table}[]
\centering
\caption{Error rates of non-IID  data based on CIFAR-100, HAM10000, Tiny ImageNet, and Reuters, with $n=5$ and $m=1$. The error rates of FA for the four datasets are 0.59, 0.25, 0.58, and 0.36 when there is no attack.}
\arrayrulecolor{black}
\resizebox{\linewidth}{!}{
\begin{tabular}{cccccccc} 
\hline
\rowcolor[HTML]{FFCCC9}
Dataset                 & Attack & KM & NC & FT & FM & DF & SF \\
\hline
\hline
\multirow{4}{*}{CIFAR-100}   & N/A    &  0.67  & 0.59  &  0.78 &  0.59  & 0.59   & 0.58  \\
                         & LA    &  0.99  &  0.59 &  0.85 &  0.90  &  0.59  & 0.99 \\
                         & MB      & 0.99 &0.58  & 0.82    &   0.87 &  0.59 & 0.99 \\
                         & Faker   & 0.99  &  \textbf{0.98}  &  \textbf{0.99}  &  \textbf{0.98}  & \textbf{0.95}   & 0.99   \\
\arrayrulecolor{black}\cline{1-1}\arrayrulecolor{black}\cline{2-8}
\multirow{4}{*}{HAM10000}  & N/A    & 0.24   &  0.25  &   0.26 & 0.24  &   0.25 &  0.24 \\
                         & LA    & 0.28   &  0.25  &  0.33  &  0.34  & 0.25   & 0.24   \\
                         & MB      & 0.26   & 0.25  &  0.34  &  0.34  &  0.25  & 0.24  \\ 
                         & Faker   & \textbf{0.34}   & \textbf{0.78}  & 0.34   &  \textbf{0.35}  &  \textbf{0.45}  &  \textbf{0.34}  \\ 
\arrayrulecolor{black}\cline{1-1}\arrayrulecolor{black}\cline{2-8}
\multirow{4}{*}{Tiny ImageNet}  & N/A    & 0.68   &  0.58  & 0.61  & 0.60 &  0.60 & 0.60 \\
                         & LA    & 0.93  &  0.72 & 0.84  & 0.73   & 0.69 & 0.99 \\
                         & MB      & 0.82  & 0.65  & 0.79  &  0.71 & 0.63  & 0.99 \\ 
                         & Faker   &  \textbf{0.99}  & \textbf{0.99}  & \textbf{0.99} & \textbf{0.99}  & \textbf{0.99}   & 0.99  \\ 
\arrayrulecolor{black}\cline{1-1}\arrayrulecolor{black}\cline{2-8}
\multirow{4}{*}{Reuters}  & N/A    & 0.47   &  0.36  & 0.45  & 0.63 & 0.35 & 0.39  \\
                         & LA    & 0.57  & 0.36  &  0.64 &  0.64  &  0.38  & 0.99   \\
                         & MB      & 0.51  & 0.36& 0.64  &   0.63 &  0.37 &0.99  \\ 
                         & Faker   & \textbf{0.60}   & \textbf{0.70}  &  \textbf{0.64}&  \textbf{0.90} &  \textbf{0.71}  & 0.99  \\ 
\arrayrulecolor{black}\cline{1-1}\arrayrulecolor{black}\cline{2-8}
\end{tabular}}
\arrayrulecolor{black}
\label{noniid_ex}
\end{table}

\textbf{Evaluation of Faker on Other Datasets.} 
Since the performance of all four datasets trained in the FL system decreases significantly in terms of accuracy as the number of users increases, we choose to set a smaller number of users to simulate the cross-silo FL case. We set $n=5$ and $m=1$, the other settings are the same as the experiments mentioned in Section \ref{exp_se}. The results of the error rates are shown in Table \ref{noniid_ex}. We can see that Faker outperforms the benchmark attacks on these datasets in decreasing test accuracy. The results in Table \ref{success_ex} show that Faker can always pass the detection of the defender.
The results in Table \ref{time_ex} indicate that Faker is more time-efficient than the benchmark attacks.

\textbf{Evaluation of SPP on Other Datasets.}
We also test SPP on these four datasets, and the results are presented in Fig. \ref{fig_spp}, which show that SPP is effective in mitigating the negative impacts of Faker on the global model.

\section{Future Research Directions}\label{f_d}
Below are several potential research directions.

\textbf{In-depth Study of the Robustness of Similarity Metrics.} In this paper, the robustness of the similarity metrics is analyzed only in the FL system, while more theoretical and experimental evaluations are required for its application when employed in other systems.

\textbf{Adapting Faker to Undermine Other Similarity-based Mechanisms.} Faker is not limited to launching adversarial attacks, but can also be used to assist clients in obtaining overrides during fairness assessments.

\textbf{More Efficient Solutions for $\textbf{P0}$.} In this paper, we take an approximate solution for $\textbf{P0}$, although experiments have shown that its performance meets the requirements for launching an attack, more efficient and accurate solutions may be required in the future.

\textbf{Studying the Robustness of Other Evaluation Metrics.} This paper is the first attempt to analyze the robustness of evaluation metrics in FL systems, while the robustness of other evaluation metrics is not yet clear.

\if()
\subsection{Use Case of Faker}
In this part, we provide a use case to illustrate the workflow of Faker. We take Faker against norm-clipping as an example. For simplicity, we use 10 parameters of LeNet-5 when processing MNIST to represent the models of FL. We let $ w_i =(0.0409,0.02380,0.0068,0.0312,0.0006,0.0024,0.0208,\\0.0403,0.04274,0.0435)$ and $\theta=1$, which means that the server will discard the local models if $L(w_i)>1$. 
\fi
\end{appendices}

\end{document}